\documentclass[10pt]{article}

\usepackage[utf8]{inputenc} % allow utf-8 input
\usepackage[T1]{fontenc}    % use 8-bit T1 fonts
\usepackage{hyperref}       % hyperlinks
\usepackage{url}            % simple URL typesetting
\usepackage{booktabs}       % professional-quality tables
\usepackage{amsfonts}       % blackboard math symbols
\usepackage{nicefrac}       % compact symbols for 1/2, etc.
\usepackage{microtype}      % microtypography

\usepackage{booktabs,colortbl}

\usepackage{amssymb}
\usepackage{amsmath}
\usepackage{enumerate}
\usepackage{capt-of}
\usepackage{graphicx} % default *.eps
\usepackage{epstopdf}
\usepackage{wrapfig}
\usepackage{multirow}
\usepackage[ruled]{algorithm2e}
\usepackage{algorithmic}

\usepackage{subfig}
\usepackage{tabularx}
\usepackage{afterpage}
\usepackage{floatflt}
\usepackage{array}
\usepackage{arydshln}
\usepackage{makecell}
\usepackage{xcolor}
\usepackage{graphbox}

\usepackage[accepted]{icml2020}

\icmltitlerunning{Minimally distorted Adversarial Examples with a Fast Adaptive Boundary Attack}

\newtheorem{proposition}{Proposition}[section]

\newenvironment{proof}{\textit{Proof.}}{\hfill$\square$}

\newtheorem{innercustomprop}{Proposition}
\newenvironment{customprop}[1]{\renewcommand\theinnercustomprop{#1}\innercustomprop}{\endinnercustomprop}

\newcommand{\inner}[1]{\left\langle#1\right\rangle}

\def\R{\mathbb{R}}

\newcommand{\norm}[1]{\left\|#1\right\|}

\def\argmax{\mathop{\rm arg\,max}\limits}%    a math operator.
\def\argmin{\mathop{\rm arg\,min}\limits}%    a math operator.

\def\maxop{\mathop{\rm max}\limits}
\def\sign{\mathop{\rm sign}\limits}

\def\min{\mathop{\rm min}\nolimits}
\def\max{\mathop{\rm max}\nolimits}

\newcolumntype{C}[1]{>{\centering\arraybackslash}p{#1}}
\newcolumntype{L}[1]{>{\raggedright\arraybackslash}p{#1}}
\newcolumntype{R}[1]{>{\raggedleft\arraybackslash}p{#1}}
\newcommand{\commentout}[1]{}

\newif\ifpaper
\papertrue
\title{Minimally distorted Adversarial Examples with a Fast Adaptive Boundary Attack}

\begin{document}
	
\twocolumn[
\icmltitle{Minimally distorted Adversarial Examples with a Fast Adaptive Boundary Attack}
	\begin{icmlauthorlist}
		\icmlauthor{Francesco Croce}{aff}
		\icmlauthor{Matthias Hein}{aff}	
	%}
	\end{icmlauthorlist}
	
	\icmlaffiliation{aff}{University of T{\"u}bingen, Germany}
	%\icmlaffiliation{to}{Department of Computation, University of Torontoland, Torontoland, Canada}
	%\icmlaffiliation{goo}{Googol ShallowMind, New London, Michigan, USA}
	%\icmlaffiliation{ed}{School of Computation, University of Edenborrow, Edenborrow, United Kingdom}
	
	\icmlcorrespondingauthor{F. Croce}{francesco.croce@uni-tuebingen.de}
	%\icmlcorrespondingauthor{Cieua Vvvvv}{c.vvvvv@googol.com}
	%\icmlcorrespondingauthor{Eee Pppp}{ep@eden.co.uk}
	
	\icmlkeywords{Machine Learning, ICML}
	
	\vskip 0.3in
	]
	
	
	
	\printAffiliationsAndNotice{}  % leave blank if no need to mention equal contribution
	%\printAffiliationsAndNotice{\icmlEqualContribution} % otherwise use the standard text
	
	\begin{abstract}
	The evaluation of robustness against adversarial manipulation of neural networks-based classifiers is mainly tested with empirical attacks as methods for the exact computation, even when available, do not scale to large networks. We propose in this paper a new white-box adversarial attack wrt the $l_p$-norms for $p \in \{1,2,\infty\}$ aiming at finding the minimal perturbation necessary to change the class of a given input. It has an intuitive geometric meaning, yields quickly high quality results, minimizes the size of the perturbation (so that it returns the robust accuracy at every threshold with a single run). It performs better or similar to state-of-the-art attacks which are partially specialized to one $l_p$-norm, and is robust to the phenomenon of gradient masking.
	\end{abstract}
		
\section{Introduction}
The finding of the vulnerability of neural networks-based classifiers to adversarial examples, that is small perturbations of the input able to modify the decision of the models, started a fast development of a variety of attack algorithms. The high effectiveness of adversarial attacks reveals the fragility of these networks which questions their safe and reliable use in the real world, especially in safety critical applications. Many defenses have been proposed to fix this issue \citep{GuRig2015,ZheEtAl2016,PapEtAl2016a,HuaEtAl2016,BasEtAl2016,MadEtAl2018}, but with limited success, as new more powerful attacks showed \citep{CarWag2017, AthEtAl2018, MosEtAl18}. In order to trust the decision of a model, it is necessary to evaluate the exact adversarial robustness. Although this is possible for ReLU networks \citep{KatzEtAl2017,TjeTed2017}
%using mixed-integer programming,
these techniques do not scale to commonly used large networks. Thus, the robustness is evaluated approximating the solution of the minimal adversarial perturbation problem through adversarial attacks.\\
One can distinguish attacks into black-box \citep{NarKas16,BreRauBet18,SuVarKou19}, where one is only allowed to query the classifier, and white-box attacks, where one has full control over the network, according to the attack model used to create adversarial examples (typically some $l_p$-norm, but others have become popular as well, e.g. \cite{BroEtAl2017,EngEtAl2017, WonEtAl2019}), whether they aim at the minimal adversarial perturbation \citep{CarWag2016,CheEtAl2018, CroRauHei2019} or rather any perturbation below a threshold \citep{KurGooBen2016a,MadEtAl2018,ZheEtAl2019}, if they have lower \citep{MooFawFro2016,ModEtAl19} or higher \citep{CarWag2016,CroRauHei2019} computational cost. Moreover, it is clear that due to the non-convexity of the problem there exists no universally best attack (apart from the exact methods), since this depends on runtime constraints, networks architecture, dataset, etc. However, our goal is to have an attack which performs well under a broad spectrum of conditions with minimal amount of hyperparameter tuning.

In this paper we propose a new white-box attack scheme which performs comparably or better than established attacks and has the following features: first, it 
aims at adversarial samples with \textit{minimal distance} to the attacked point, measured wrt the $l_p$-norms with $p\in\{1,2, \infty\}$. Compared to the popular PGD (projected gradient descent)-attack of \cite{MadEtAl2018} this has the clear advantage that our method does not need to be restarted for every threshold $\epsilon$
%of an attack model of allowed perturbations
if one wants to evaluate the success rate of the attack with perturbations constrained to be in
$\{\delta \in \R^d \,|\, \norm{\delta}_p\leq \epsilon\}$. Thus it is particularly suitable to get a complete picture on the robustness of a classifier with low computational cost.
%for several thresholds $\epsilon$.
Second, it achieves \textit{fast} good quality in terms of average distortion or robust accuracy.
%compared to the PGD-attack and other fast attacks.
At the same time we show that increasing the number of restarts keeps improving the results and makes it competitive to or stronger than the strongest available attacks.
Third, although it comes with a few parameters, these generalize well across datasets, architectures and norms considered, so that we have an almost \textit{off-the-shelf method}.
%for which it is sufficient to specify the number of iterations and restarts.
Most importantly, unlike PGD and other methods, there is no step size parameter, which potentially has to be carefully adapted to every new network, and we show 
that it is scaling invariant. Both properties lead to the fact that it is robust to gradient masking which can be a problem for PGD \cite{TraBon2019}.

\section{FAB: a Fast Adaptive Boundary Attack}\label{sec:FAB}
We first introduce minimal adversarial perturbations, then we recall the definition and properties of the projection wrt the $l_p$-norms of a point on the intersection of a hyperplane and box constraints, as they are an essential part of our attack. Finally, we present our FAB-attack algorithm to generate minimally distorted adversarial examples.

\subsection{Minimal adversarial examples}\label{sec:minimal}
Let $f:\R^d \rightarrow \R^K$ be a classifier which assigns every input $x\in \R^d$ (with $d$ the dimension of the input space) to one of the $K$ classes according to $\argmax_{r = 1, \ldots, K} f_r(x)$. In many scenarios the input of $f$ has to satisfy a specific set of constraints $C$, e.g. images are represented as elements of $[0,1]^d$. Then, given a point $x\in\R^d$ with true class $c$, we define the \textit{minimal adversarial perturbation} for $x$ wrt the $l_p$-norm as \begin{equation} \begin{split}
\label{eq:advopt}
&\delta_{\textrm{min},p} = \argmin_{\delta \in \mathbb{R}^d} \; \norm{\delta}_p, \\ &\textrm{s.th.}  \quad    \maxop_{l\neq c} \; f_l(x+\delta) \geq f_c(x+\delta), \quad x+\delta \in C.\end{split}\end{equation}
The optimization problem \eqref{eq:advopt} is non-convex and NP-hard for non-trivial classifiers \cite{KatzEtAl2017} and, although for some classes of networks %the optimization problem
it can be formulated as a mixed-integer program \cite{TjeTed2017}, the computational cost of solving it is prohibitive for large, normally trained networks. Thus, $\delta_{\textrm{min},p}$ is usually approximated by an \textit{attack algorithm}, which can be seen as a heuristic to solve \eqref{eq:advopt}. We will see in the experiments that current attacks sometimes drastically overestimate $\norm{\delta_{\textrm{min},p}}_p$ and thus the robustness of the networks.

\subsection{Projection on a hyperplane with box constraints}
Let $w\in\R^d$ and $b\in\R$ be the normal vector and the offset defining the hyperplane $\pi:\inner{w,x} + b=0$. Let $x\in \R^d$, we denote by the \textit{box-constrained projection} wrt the $l_p$-norm of $x$ on $\pi$ (projection onto the intersection of the box $C=\{z \in \R^d: l_i \leq z_i \leq u_i \}$  and the hyperplane $\pi$) the following minimization problem:
\begin{align}\label{eq:proj_1}
&z^* =  \argmin_{z \in \R^d}\; \norm{z-x}_p \\ &\textrm{s.th.}\quad \inner{w,z} + b=0, \quad l_i \leq z_i \leq u_i, \; i=1,\ldots, d,\nonumber%\end{split}
\end{align} 
where $l_i, u_i\in \R$ are lower and upper bounds on each component of $z$. For $p\geq 1$ the optimization problem \eqref{eq:proj_1} is convex.
\cite{HeiAnd2017} proved that for $p\in \{1, 2, \infty \}$ the solution can be obtained in $\mathcal{O}(d \log d)$ time, that is the complexity of sorting a vector of $d$ elements, as well as determining that there exists no feasible point.

Since this projection is part of our iterative scheme, we need to handle specifically the case of \eqref{eq:proj_1} being infeasible. In this case, defining $\rho= \sign (\inner{w,x}+b)$, we instead compute $z' = \argmin_{l_i \leq z_i \leq u_i} \rho \cdot ( \inner{w,z}+b)$, whose solution is
\begin{align}\label{eq:proj_2}
\commentout{
z' \hspace{-0.6mm}&=
                     \hspace{-0.5mm}\argmin_{l_i \leq z_i \leq u_i} \rho ( \inner{w,z}+b)
                  \hspace{-0.5mm}= \hspace{-0.6mm}\begin{cases} l_i & \hspace{-1.4mm}\textrm{if } \rho w_i >0,\\ u_i & \hspace{-1.4mm}\textrm{if } \rho w_i <0,\\ 
									x_i & \hspace{-1.4mm}\textrm{if } w_i=0\end{cases} \,  i=1,\ldots,d.}
									%
%z' &= \argmin_{l_i \leq z_i \leq u_i} \rho \cdot ( \inner{w,z}+b)\\
z_i' & = \hspace{0mm}\begin{cases} l_i &  \textrm{if } \rho w_i >0,\\ u_i &  \textrm{if } \rho w_i <0,\\ 
x_i & \textrm{if } w_i=0\end{cases} \;\textrm{for}\;  i=1,\ldots,d.
\end{align} 
Assuming that the point $x$ satisfies the box constraints (as it holds in our algorithm), this is equivalent to identifying the corner of the $d$-dimensional box, defined by the componentwise constraints on $z$, closest to the hyperplane $\pi$. Note that if \eqref{eq:proj_1} is infeasible then the objective function of \eqref{eq:proj_2} stays positive and the points $x$ and $z$ are strictly contained in the same of the two halfspaces divided by $\pi$.
Finally, we define the projection operator \begin{equation} \textrm{proj}_p:(x,\pi,C)\longmapsto \left\lbrace\begin{array}{l l} z^* & \textrm{if }\eqref{eq:proj_1} \textrm{ is feasible} \\ z' & \textrm{else} \end{array}\right. \label{eq:op_proj_p}
\end{equation}
which yields the point as close as possible to $\pi$ without violating the box constraints.

\subsection{FAB-attack}
We introduce now our algorithm to produce minimally distorted adversarial examples, wrt any $l_p$-norm for $p\in \{1, 2, \infty \}$, for a given point $x_\textrm{orig}$ initially correctly classified by $f$ as class $c$. The high-level idea is that, first, we use the linearization of the classifier at the current iterate $x^{(i)}$ to compute the box-constrained projections of $x^{(i)}$ respectively $x_\textrm{orig}$ onto the approximated decision hyperplane and, second, we take a convex combinations of these projections depending on the distance of $x^{(i)}$ and $x_\textrm{orig}$ to the decision hyperplane. Finally, we perform an extrapolation step. We explain below the geometric motivation behind these steps. The attack closest in spirit is DeepFool \cite{MooFawFro2016} which is known to be very fast but suffers from low quality. DeepFool just
tries to find the decision boundary quickly but has no incentive to provide a solution close to $x_\textrm{orig}$. Our scheme resolves this main problem and, together
with the exact projection we use, leads to a principled way to track the decision boundary (the surface where the decision of $f$ changes) \textit{close} to $x_\textrm{orig}$.

If $f$ was a linear classifier then the closest point to $x^{(i)}$ on the decision hyperplane
%with a different classification
could be found in closed form. However neural networks are highly non-linear (although ReLU networks, i.e. neural networks which use ReLU as activation function, are piecewise affine functions and thus locally a linearization of the network is an exact description of the classifier). Let $l \neq c$, then the decision boundary between classes $l$ and $c$ can be locally approximated using a first order Taylor expansion at $x^{(i)}$ by the hyperplane %$\pi_l(z)$ defined as
\begin{equation} \begin{split} \pi_l(z): \; &f_l(x^{(i)}) - f_c(x^{(i)})\\ &+ \inner{\nabla f_l(x^{(i)}) - \nabla f_c(x^{(i)}), z - x^{(i)}} = 0.\end{split} \label{eq:lin_db} 
\end{equation}
Moreover the $l_p$-distance $d_p(x^{(i)},\pi_l)$ of $x^{(i)}$ to $\pi_l$ is given, assuming $\frac{1}{p}+\frac{1}{q}=1$, by \begin{equation}\label{eq:lin_dist_db} d_p(x^{(i)},\pi_l) = \frac{|f_l(x^{(i)}) - f_c(x^{(i)})|}{\norm{\nabla f_l(x^{(i)}) - \nabla f_c(x^{(i)})}_q}.
\end{equation}
Note that if $d_p(x^{(i)},\pi_l)=0$ then $x^{(i)}$ belongs to the true decision boundary. Moreover, if the local linear approximation of the network is correct then 
the class $s$ with the decision hyperplane closest to the point $x^{(i)}$ can be computed as 
\begin{equation}s = \argmin_{l \neq c} \frac{|f_l(x^{(i)}) - f_c(x^{(i)})|}{\norm{\nabla f_l(x^{(i)}) - \nabla f_c(x^{(i)})}_q}.\label{eq:min_dist} \end{equation}
Thus, given that the approximation holds in some large enough neighborhood, the projection $\textrm{proj}_p(x^{(i)},\pi_s,C)$ of $x^{(i)}$ onto $\pi_s$ lies on the decision boundary
%(without taking the box constraints into account).
(unless \eqref{eq:proj_1} is infeasible).

\begin{figure}[t]
	%\flushleft
	\centering
	\includegraphics[clip,trim=180mm 30mm 160mm 40mm,scale=0.2]{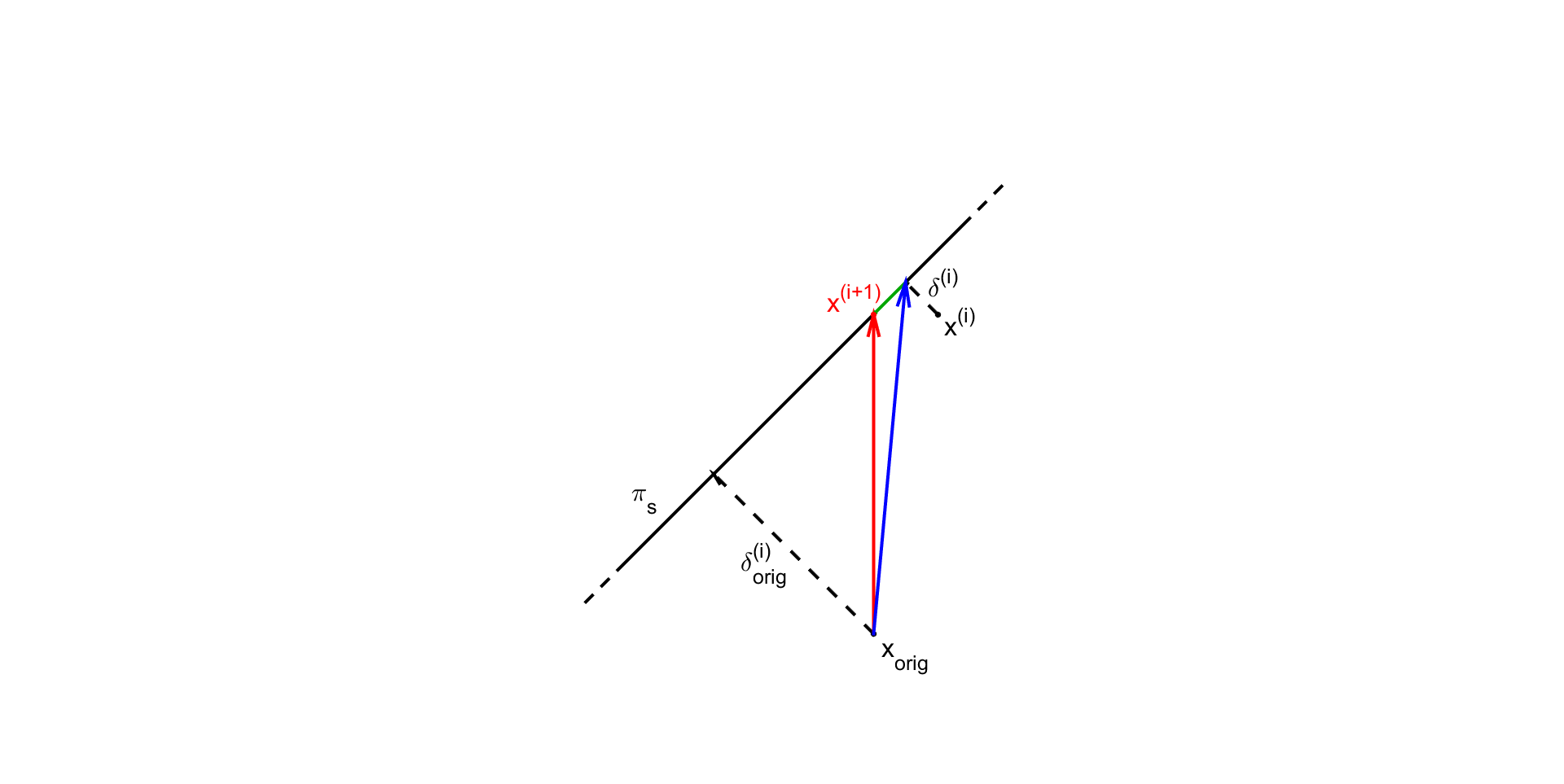}
	%\flushright
	\includegraphics[clip,trim=180mm 30mm 160mm 40mm,scale=0.2]{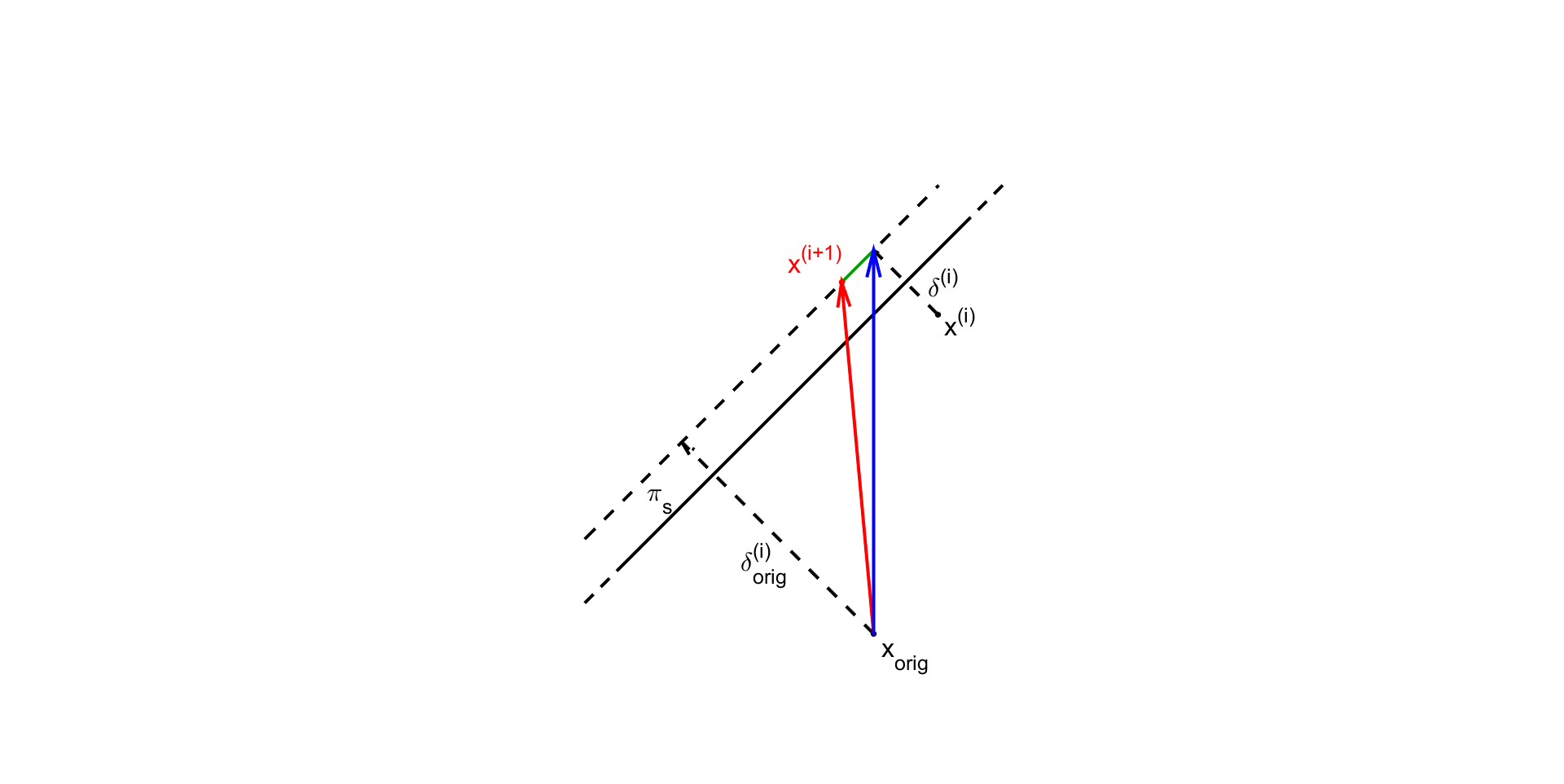}
	\caption{Visualization of FAB-attack scheme: Left, case $\eta=1$, right, $\eta>1$ (extrapolation). In blue we show $\textrm{proj}_p(x^{(i)},\pi_s,C)$, the iterate one would get without any bias towards $x_\textrm{orig}$, in green the effect of the bias we introduce and in red the actual iterate $x^{(i+1)}$ of FAB-attack in \eqref{eq:iter_step}. FAB-attack stays closer to $x_\textrm{orig}$ compared to the unbiased gradient step with $\textrm{proj}_p(x^{(i)},\pi_s,C)$.}
	\label{fig:vis_scheme}
\end{figure}

\paragraph{Biased gradient step:}
The iterative algorithm $x^{(i+1)}=\textrm{proj}_p(x^{(i)},\pi_s,C)$ would be similar to DeepFool except that our projection operator is exact whereas they project onto the hyperplane and then clip to $[0,1]^d$. This scheme is not biased towards the original target point $x_\textrm{orig}$,
thus it goes typically further than necessary to find a point on the decision boundary as basically the algorithm
does not aim at the minimal adversarial perturbation. Then we consider additionally $\textrm{proj}_p(x_\textrm{orig},\pi_s,C)$ and use instead the iterative step, with $x^{(0)}=x_{\textrm{orig}}$ and $\alpha \in [0,1]$, defined as
\begin{align} \label{eq:iter_step_noextra} 
 x^{(i+1)} = (1-\alpha)\, \textrm{proj}_p(x^{(i)},\pi_s,C) + \alpha\, \textrm{proj}_p(x_\textrm{orig},\pi_s,C),
\end{align}
which biases the step towards $x_\textrm{orig}$ (see Figure \ref{fig:vis_scheme}). Note that this is a convex combination of two points on $\pi_s$ and in $C$ and thus also $x^{(i+1)}$ lies on $\pi_s$ and is contained in $C$.

As we wish a scheme with minimal
amount of parameters, our goal is an automatic selection of $\alpha$ based on the available geometric quantities. Let 
\begin{gather*} \delta^{(i)} = \textrm{proj}_p(x^{(i)},\pi_s,C) - x^{(i)}, \\ %\quad \textrm{ and }
\delta_{\textrm{orig}}^{(i)} = \textrm{proj}_p(x_\textrm{orig},\pi_s,C) - x_\textrm{orig}.\end{gather*}
Note that $\norm{\delta^{(i)}}_p$ and $\norm{\delta^{(i)}_{\textrm{orig}}}_p$ are the distances of $x^{(i)}$ and $x_\textrm{orig}$ to $\pi_s$ (inside $C$). We propose to use for the parameter $\alpha$ the relative magnitude of these two distances, that is
\begin{equation} \alpha = \min\left\lbrace \frac{\norm{\delta^{(i)}}_p}{\norm{\delta^{(i)}}_p + \norm{\delta_{\textrm{orig}}^{(i)}}_p}, \alpha_\textrm{max} \right\rbrace \in [0,1]\label{eq:alpha_step}.
\end{equation}
The motivation for doing so is that if $x^{(i)}$ is close to the decision boundary then we should stay close to this point (note that $\pi_s$ is the approximation of $f$ computed at $x^{(i)}$ and thus it is valid in a small neighborhood of $x^{(i)}$, whereas $x_\textrm{orig}$ is farther away). On the other hand
we want to have the bias towards $x_\textrm{orig}$ in order not to go too far away from $x_\textrm{orig}$. This is why $\alpha$ depends on the distances of $x^{(i)}$ and $x_\textrm{orig}$ to $\pi_s$ but we limit it from above with $\alpha_{\max}$.
Finally, we use a small extrapolation step as we noted empirically, similarly to  \cite{MooFawFro2016}, that this helps to cross faster
the decision boundary and get an adversarial sample. This leads to the final scheme:
\begin{align} \label{eq:iter_step} 
x^{(i+1)} \hspace{-0.3mm}=  \hspace{-0.3mm}\textrm{proj}_C\Big(\hspace{-0.3mm}(1 - \alpha)\big(x^{(i)} +\eta\delta^{(i)}\big) +\alpha \big(x_\textrm{orig} + \eta \delta^{(i)}_\textrm{orig}\big) \hspace{-0.3mm}\Big),
\end{align}
where $\alpha$ is chosen as in $\eqref{eq:alpha_step}$, $\eta\geq 1$ and $\textrm{proj}_C$ is the projection onto the box which can be done by clipping. In Figure \ref{fig:vis_scheme} we visualize the scheme: in black one can see the hyperplane $\pi_s$ and the vectors $\delta_{\textrm{orig}}^{(i)}$ and $\delta^{(i)}$, in blue the step not biased towards $x_\textrm{orig}$, while in red the biased step of FAB-attack, see \eqref{eq:iter_step}. The green vector shows the bias towards the original point we introduce.
On the left of Figure \ref{fig:vis_scheme} we use $\eta=1$, while on the right we use extrapolation $\eta>1$.

\paragraph{Interpretation of $\textrm{proj}_p(x_\textrm{orig},\pi_s,C)$:} 
The projection of the target point $x_{\textrm{orig}}$ onto the intersection of $\pi_s$ and $C$ is %defined as
\begin{equation*}
\argmin_{z \in \R^d}\; \norm{z-x_{\textrm{orig}}}_p \; \textrm{s.th.}\; \inner{w,z} + b=0, \; l_i \leq z_i \leq u_i, 
\end{equation*}
Note that replacing $z$ by $x^{(i)}+\delta$ we can rewrite this as 
\begin{equation*}
\begin{split}
\argmin_{\delta \in \R^d}\; &\norm{x^{(i)}+\delta - x_{\textrm{orig}}}_p \\ \textrm{s.th.}\quad &\inner{w,x^{(i)}+\delta} + b=0, \quad l_i \leq x_i+\delta_i \leq u_i. \end{split}
\end{equation*}
This can be interpreted as the minimization of the distance of the next iterate $x^{(i)}+\delta$ to the target point $x_{\textrm{orig}}$ so that $x^{(i)}+\delta$
lies on the intersection of the (approximate) decision hyperplane and the box $C$. This point of view on the projection $\textrm{proj}_p(x_\textrm{orig},\pi_s,C)$ again justifies using a convex combination of the two projections in our scheme in \eqref{eq:iter_step}.

\paragraph{Backward step:} The described scheme finds in a few iterations adversarial perturbations. However, we are interested in minimizing their norms. Thus, once we have a new point $x^{(i+1)}$, we check whether it is assigned by $f$ to a class different from $c$. In this case, we apply \begin{equation} x^{(i+1)} = (1-\beta) x_\textrm{orig} + \beta x^{(i+1)}, \quad \beta\in(0,1), \label{eq:step_adv} \end{equation} 
that is we go back towards $x_\textrm{orig}$ on the segment $[x^{(i+1)},x_\textrm{orig}]$, effectively starting again the algorithm at a point which is close to the decision boundary. In this way, due to the bias of the method towards $x_\textrm{orig}$ we successively find adversarial perturbations of smaller norm, meaning that the algorithm \textit{tracks} the decision boundary while getting closer to $x_\textrm{orig}$. We fix $\beta=0.9$ in all experiments.

\paragraph{Final search:} Our scheme finds points close to the decision boundary but often they are slightly off as the linear approximation is not exact and we apply the extrapolation step with $\eta > 1$. 
Thus, after finishing $N_\textrm{iter}$ iterations of our algorithmic scheme, we perform a last, fast step to further improve the quality of the adversarial examples. Let $x_\textrm{out}$ be the closest point to $x_\textrm{orig}$ classified differently from $c$, say $s\neq c$, found with the iterative scheme. It holds that $f_s(x_\textrm{out}) - f_c(x_\textrm{out}) > 0$ and $f_s(x_\textrm{orig}) - f_c(x_\textrm{orig}) <0$. This means that, assuming $f$ continuous, there exists a point $x^*$ on the segment $[x_\textrm{out},x_\textrm{orig}]$ such that $f_s(x^*)-f_c(x^*)= 0$ and $\norm{x^* - x_\textrm{orig}}_p < \norm{x_\textrm{out} - x_\textrm{orig}}_p$. If $f$ is linear \begin{equation} x^* = x_\textrm{out} - \frac{\left(f_s(x_\textrm{out}) - f_c(x_\textrm{out})\right)(x_\textrm{out} - x_\textrm{orig})}{f_s(x_\textrm{out}) - f_c(x_\textrm{out}) + f_s(x_\textrm{orig}) - f_c(x_\textrm{orig})}.
\label{eq:linear_approx_search}
\end{equation} Since $f$ is non-linear, we compute iteratively for a few steps \begin{equation} x_\textrm{temp} = x_\textrm{out} - \frac{\left(f_s(x_\textrm{out}) - f_c(x_\textrm{out})\right) (x_\textrm{out} - x_\textrm{orig})}{f_s(x_\textrm{out}) - f_c(x_\textrm{out}) + f_s(x_\textrm{orig}) - f_c(x_\textrm{orig})},
\label{eq:linear_approx_search_2} \end{equation} each time replacing in \eqref{eq:linear_approx_search_2} $x_\textrm{out}$  with $x_\textrm{temp}$ if $f_s(x_\textrm{temp}) - f_c(x_\textrm{temp})> 0$ or $x_\textrm{orig}$  with $x_\textrm{temp}$ if instead $f_s(x_\textrm{temp}) - f_c(x_\textrm{temp})< 0$. With this kind of modified binary search one can find a
better adversarial sample with the cost of a few forward passes (which is fixed to $3$ in all experiments).

\paragraph{Random restarts:}
So far all the steps are deterministic. To improve the results, we introduce the option of random restarts, that is $x^{(0)}$ is randomly sampled in the proximity of $x_\textrm{orig}$ instead of being $x_\textrm{orig}$ itself. Most attacks benefit from random restarts, e.g. \cite{MadEtAl2018,ZheEtAl2019}, especially dealing with models trained for robustness
\cite{MosEtAl18}, as it allows a wider exploration of the input space. We choose to sample from the $l_p$-sphere centered in the original point with radius half the $l_p$-norm of the current best adversarial perturbation (or a given threshold if no adversarial example has been found yet).

\begin{algorithm}[t]
	\SetAlgoNoLine
	%\SetInd{0.0em}{0.0em}
	\caption{FAB-attack}
	\label{alg:algorithm-label}
	\SetKwInOut{Input}{Input}
	\SetKwInOut{Output}{Output}
	\Input{$x_\textrm{orig}$ original point, $c$ original class, $N_\textrm{restarts}, N_\textrm{iter},\alpha_\textrm{max},\beta,\eta, \epsilon, p$}
	\Output{$x_\textrm{out}$ adversarial example}
	$u \gets +\infty$\\
	\For{$j=1,\ldots,N_\textrm{restarts}$}{
		\lIf{$j = 1$}{$x^{(0)} \gets x_\textrm{orig}$} \lElse{$ x^{(0)} \gets  \textrm{randomly sampled s.th.}  \norm{x^{(0)} - x_\textrm{orig}}_p = \nicefrac{\min\{u,\epsilon \}}{2}$}
		\For{$i=0,\ldots,N_\textrm{iter} - 1$}{$s \gets \argmin_{l \neq c} \frac{|f_l(x^{(i)}) - f_c(x^{(i)})|}{\norm{\nabla f_l(x^{(i)}) - \nabla f_c(x^{(i)})}_q}$\\
	$\delta^{(i)}\gets \textrm{proj}_p(x^{(i)},\pi_s,C)$\\
		$\delta^{(i)}_\textrm{orig} \gets \textrm{proj}_p(x_\textrm{orig},\pi_s,C)$\\
		compute $\alpha$ as in Equation \eqref{eq:alpha_step}\\
		$\begin{aligned} x^{(i+1)} \gets \textrm{proj}_C  \Big(&(1-\alpha) \left(x^{(i)} +\eta \delta^{(i)} \right)\\
		& + \alpha (x_\textrm{orig} + \eta \delta^{(i)}_\textrm{orig}) \Big) \end{aligned}$\\
		\If{$x^{(i+1)}$ is not classified as $c$}
			{\If{$\norm{x^{(i+1)} - x_\textrm{orig}}_p < u$}{$x_\textrm{out} \gets x^{(i+1)}$\\
				$u \gets \norm{x^{(i+1)} - x_\textrm{orig}}_p$}
			$x^{(i+1)}\gets (1 - \beta)x_\textrm{orig} + \beta x^{(i+1)}$
		}
		}
	
	}
	perform 3 steps of final search on $x_\textrm{out}$ as in \eqref{eq:linear_approx_search_2}
\end{algorithm}

\paragraph{Computational cost:}
Our attack, in Algorithm \ref{alg:algorithm-label}, consists of two main operations: the computation of $f$ and its gradients and solving the projection \eqref{eq:proj_1}. We perform, for each iteration, a forward and a backward pass of the network in the gradient step and a forward pass in the backward step.
The projection can be efficiently implemented to run in batches on the GPU and its complexity depends only on the input dimension. Thus, except for shallow models, its cost is much smaller than the passes through the network.
We can approximate the computational cost of our algorithm by the total number of calls of the classifier 
$N_\textrm{iter} \times N_\textrm{restarts} \times (2 \times \textrm{forward passes} + 1 \times \textrm{backward pass})$
Per restart one has to add the three forward passes for the final search.
\begin{figure}[t]
	%\flushleft
	\centering
	\includegraphics[clip,trim=38mm 0mm 40mm 0mm,width=\columnwidth]{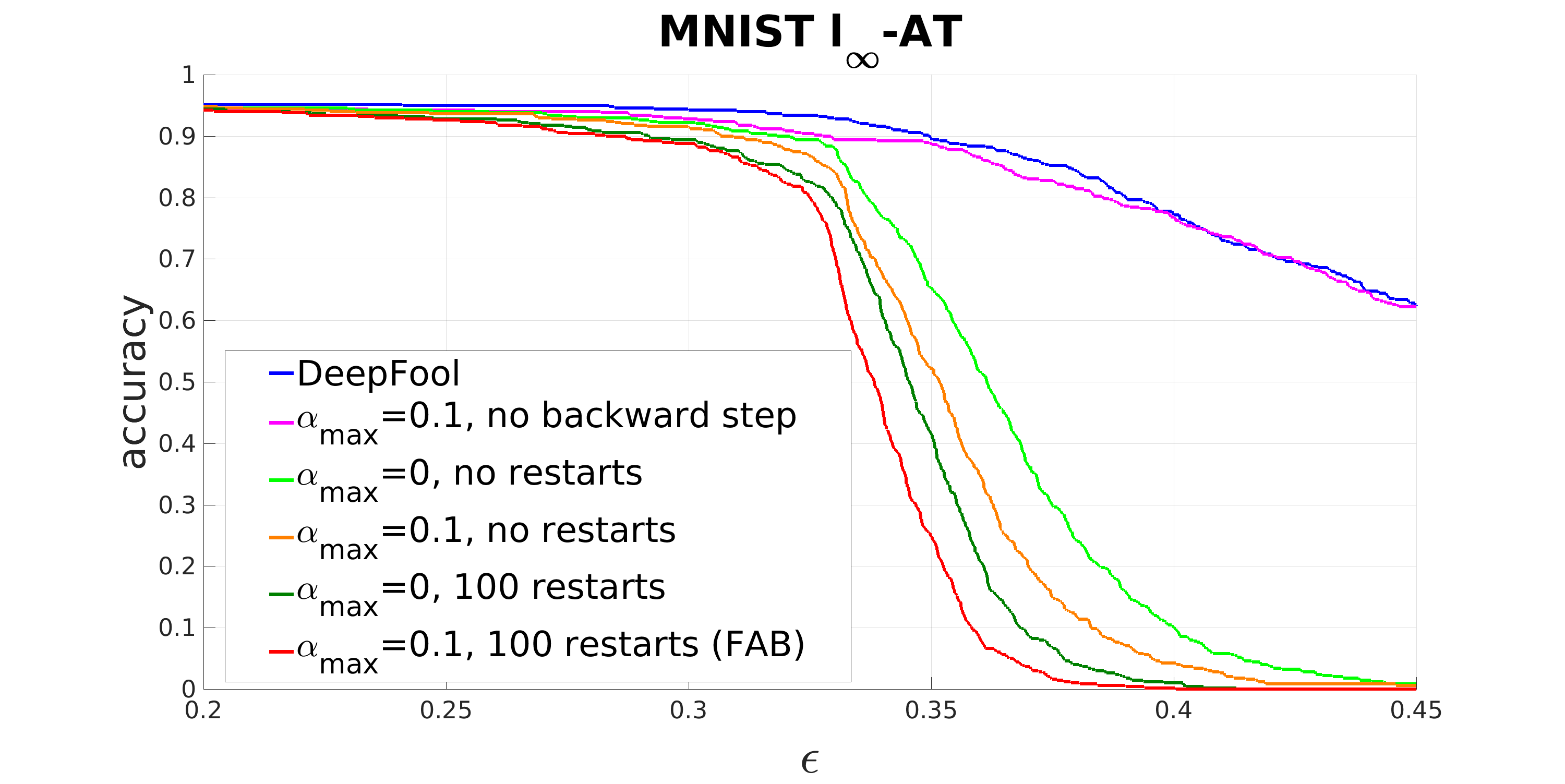}
	%\flushright
	%\includegraphics[clip,trim=40mm 0mm 39mm 0mm,width=0.48\columnwidth]{pl_df_details_cifar10_new}
	\caption{\textbf{Ablation study to DeepFool for $l_\infty$-attacks.}
		The curves show the robust accuracy as a function of the threshold $\epsilon$ under different attacks on the $l_\infty$-AT model on MNIST (lower values mean stronger attacks).
		The introduction of the convex combination ($\alpha_{\max}=0.1$, no backward step) already improves over DeepFool. If one uses the backward step, the case $\alpha_{\max}=0$ (which can be seen as an improved iterative DF) is worse than $\alpha_{\max}=0.1$ with the same number of restarts.
		%The curves show the robust accuracy as a function of the threshold $\epsilon$ under the different attacks on the $l_\infty$-AT model on MNIST.
		%(left)
		%and the \textit{plain} model on CIFAR-10 (right).
	}
	\label{fig:com_df}
\end{figure}

\subsection{Scale Invariance of FAB-attack} \label{sec:scale_inv}
For a given classifier $f$, the decisions and thus adversarial samples do not change if we rescale the classifier $g=\alpha f$ for $\alpha>0$ or shift its logits as $h = f + \beta$ for $\beta \in \R$.
The following proposition states that FAB-attack is invariant under both rescaling and shifting (proof in \ifpaper Section \ref{sec:app_scale_inv}\else supplement\fi).
\begin{proposition} \label{prop:scale_inv}
Let $f:\R^d \rightarrow \R^K$ be a classifier. Then for any $\alpha>0$ and $\beta \in \R$ the output $x_\textrm{out}$  of Algorithm \ref{alg:algorithm-label}
for the classifier $f$ is the same as of the classifiers $g=\alpha f$ and $h = f + \beta$.
\end{proposition}

We note that the cross-entropy loss $\textrm{CE}(x,y,f)=-\log(e ^{f_y(x)}/ \sum_{j=1}^K e^{f_j(x)})$ used as objective in the normal PGD attack and its gradient wrt $x$ 
\begin{align*}
\nabla_x \textrm{CE}(x,y,f) = -\nabla_x f_y(x) + \frac{\sum_{j=1}^K e^{f_j(x)} \nabla_x f_j(x)}{\sum_{j=1}^K e^{f_j(x)}}
\end{align*}
are not invariant under rescaling. Moreover,  we observe that the gradient vanishes for $\alpha f$ if $f_y(x)>f_j(x)$ for $j\neq y$ (correctly classified point)
as $\alpha \rightarrow \infty$. 
Due to finite precision the gradient becomes zero for finite $\alpha$ and it is obvious that in this case PGD gets stuck.
Due to the rescaling invariance FAB-attack is not 
affected by gradient masking due to this phenomenon
as it uses the gradient of the differences
of the logits and not the gradient of the cross-entropy loss. The latter one runs much earlier into numerical problems when one upscales the classifier due to the
exponential function. In the experiments (see below) we show that PGD can 
catastrophically fail due to a ``wrong'' scaling whereas FAB-attack is unaffected.

\subsection{Comparison to DeepFool}
The idea of exploiting the first order local approximation of the decision boundary is not novel but the basis of one of the first white-box adversarial attacks, DeepFool (DF) from \cite{MooFawFro2016}. While DF and our FAB-attack share the strategy of using a linear approximation of the classifier and projecting on the decision hyperplanes, we want to point out many key differences: first, DF does not solve the projection \eqref{eq:proj_1} but its simpler version without box constraints, clipping afterwards. Second, their gradient step does not have any bias towards the original point, that is equivalent to $\alpha=0$ in \eqref{eq:iter_step}. Third, DF does not have any backward step, final search or restart, as it stops as soon as a misclassified point is found (its goal is to provide quickly an adversarial perturbation of average quality).\\
We perform an 
ablation study of the differences to DF in Figure \ref{fig:com_df}, where we show
robust accuracy as a function of the threshold $\epsilon$ (lower is better).
We present the results of DeepFool (blue) and FAB-attack with the following variations: $\alpha_\textrm{max}=0.1$ and no backward step (magenta), $\alpha_\textrm{max}=0$ (that is no bias in the gradient step) and no restarts (light green), $\alpha_\textrm{max}=0.1$ and no restarts (orange), $\alpha_\textrm{max}=0$ and 100 restarts (dark green) and $\alpha_\textrm{max}=0.1$ and 100 restarts, that is FAB-attack, (red). We can see how every addition we make to the original scheme of DeepFool contributes to the significantly improved performance of FAB-attack when compared to the original DeepFool.

\section{Experiments}\label{sec:exps}
\paragraph{Models:}
We run experiments on MNIST, CIFAR-10 \cite{CIFAR10} and Restricted ImageNet \cite{TsiEtAl18}. For each dataset we consider a normally trained model (\textit{plain}) and two adversarially trained ones as in \cite{MadEtAl2018} wrt the $l_\infty$-norm ($l_\infty$-AT) and the $l_2$-norm ($l_2$-AT) (see \ifpaper\ref{sec:app_models} \else supplementary material \fi for details).

\begin{figure*}[h!]\centering \begin{tabular}{c}
		\textbf{MNIST - $l_\infty$-attack:}\quad
		\raisebox{-.5\height}{\includegraphics[width=0.5\columnwidth, clip, trim=10mm 0mm 15mm 0mm]{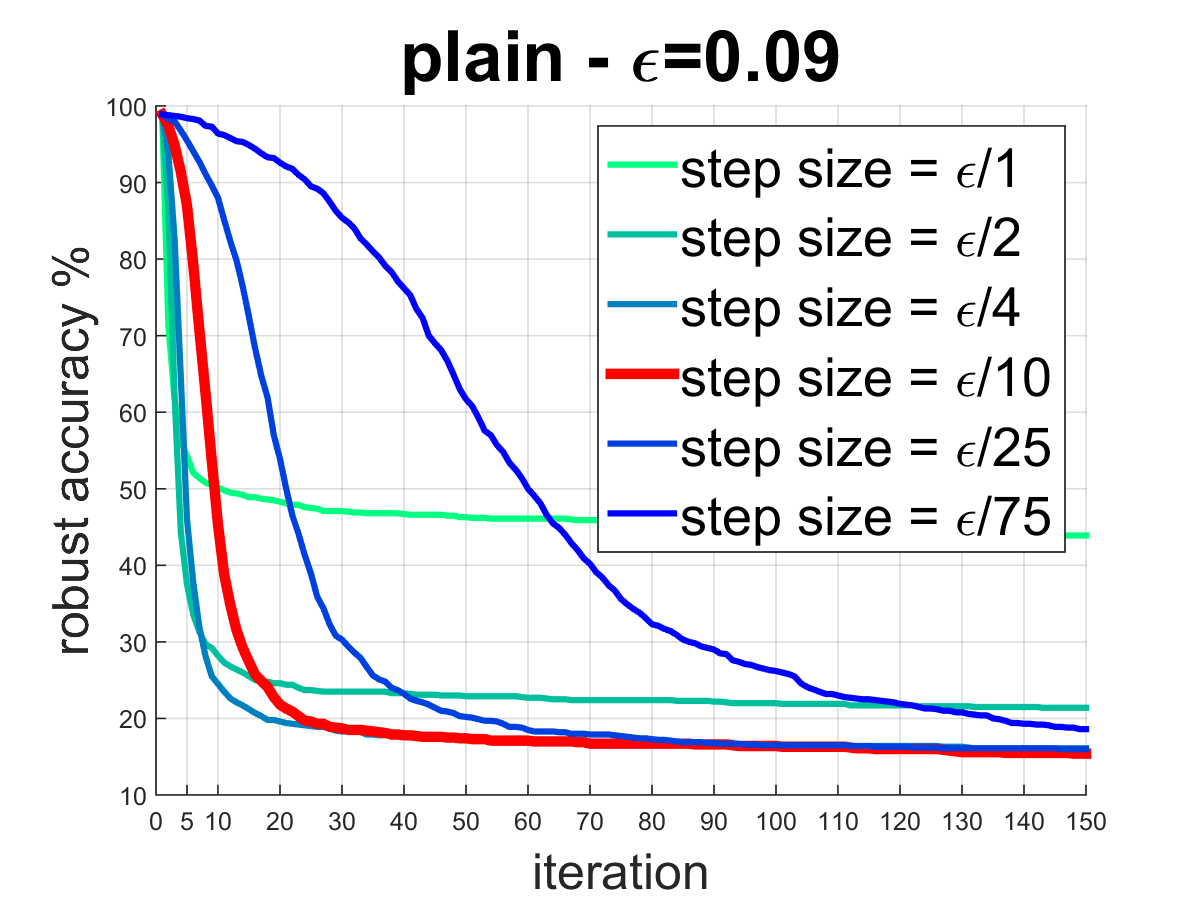}} 
		\raisebox{-.5\height}{\includegraphics[width=0.5\columnwidth, clip, trim=10mm 0mm 15mm 0mm]{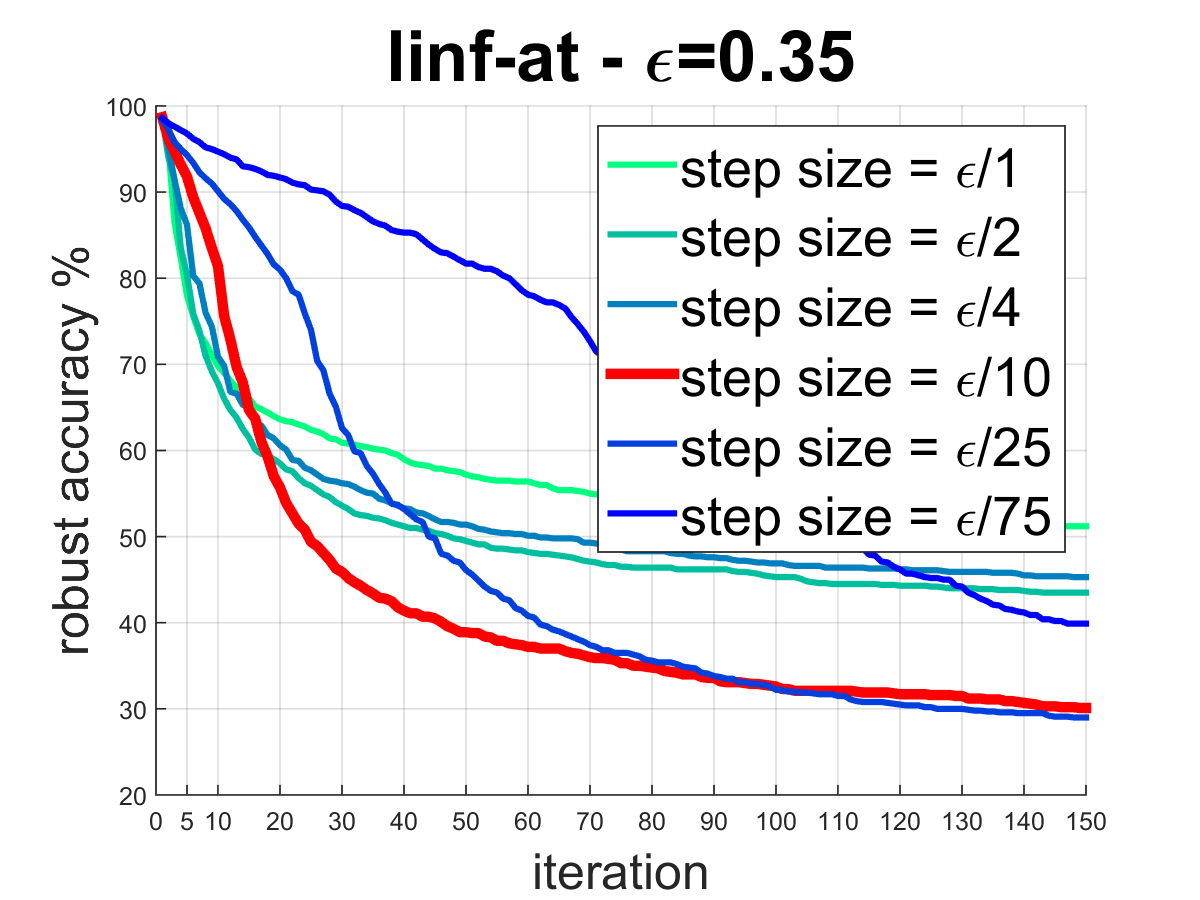}}
		\raisebox{-.5\height}{\includegraphics[width=0.5\columnwidth, clip, trim=10mm 0mm 15mm 0mm]{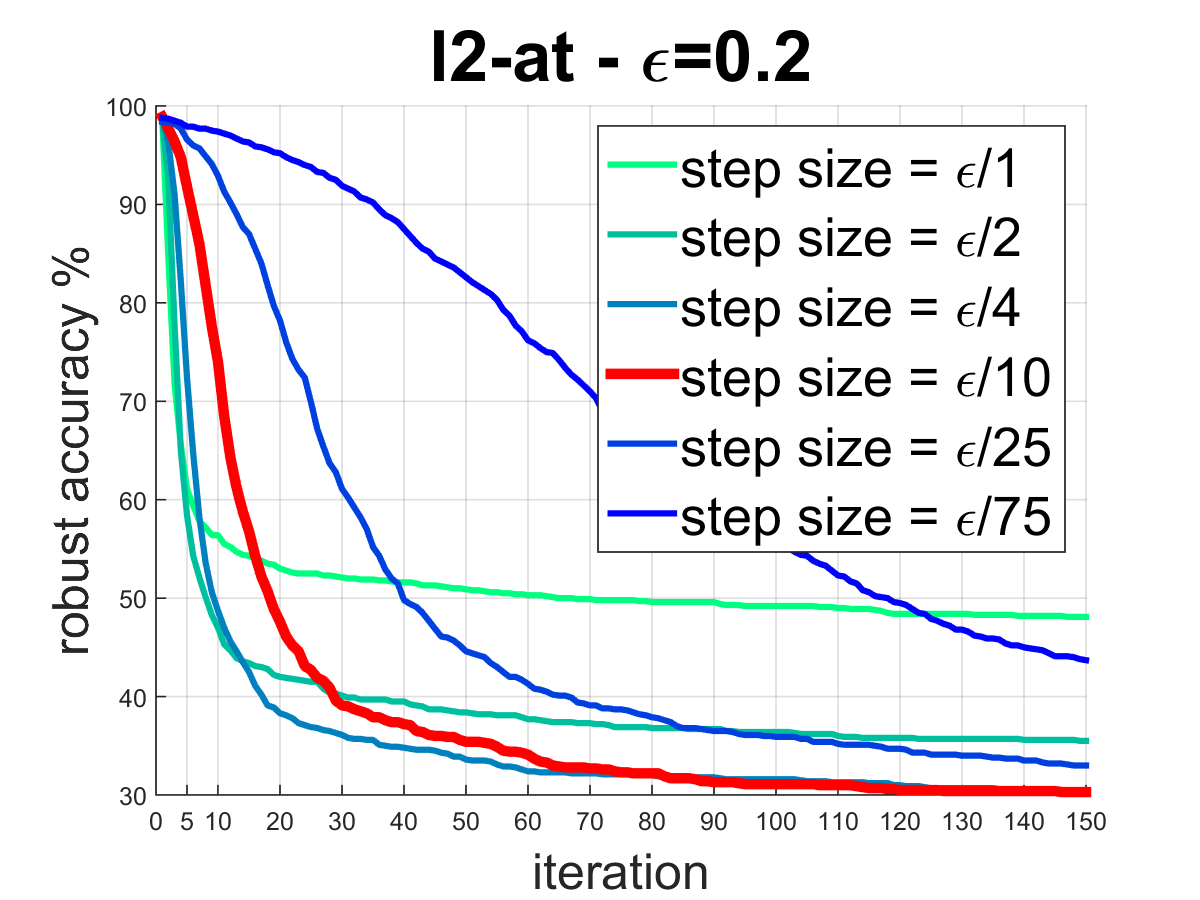}}\\
		%\hline
		\textbf{MNIST - $l_2$-attack:} \quad %\\[2mm]
		\raisebox{-.5\height}{\includegraphics[width=0.5\columnwidth, clip, trim=10mm 0mm 15mm 0mm]{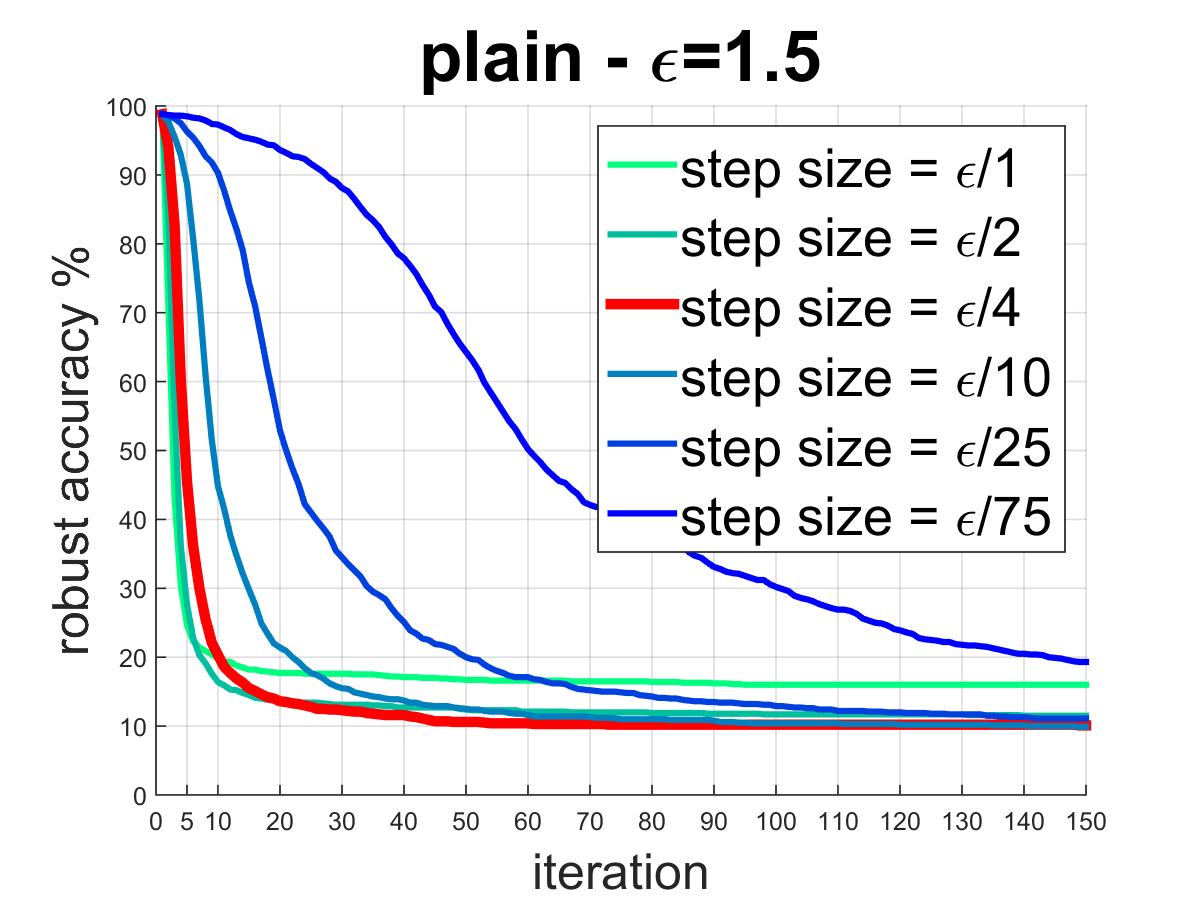}} 
		\raisebox{-.5\height}{\includegraphics[width=0.5\columnwidth, clip, trim=10mm 0mm 15mm 0mm]{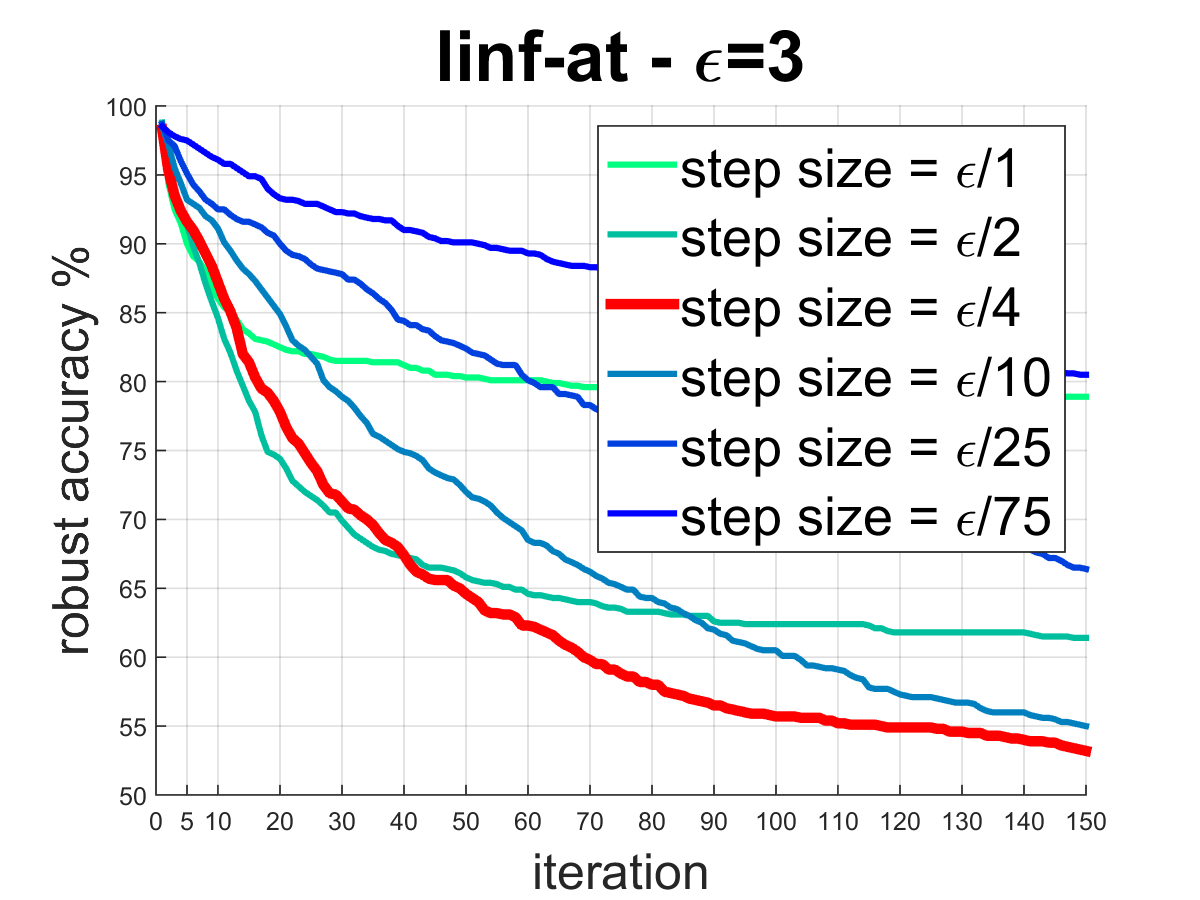}} 
		\raisebox{-.5\height}{\includegraphics[width=0.5\columnwidth, clip, trim=10mm 0mm 15mm 0mm]{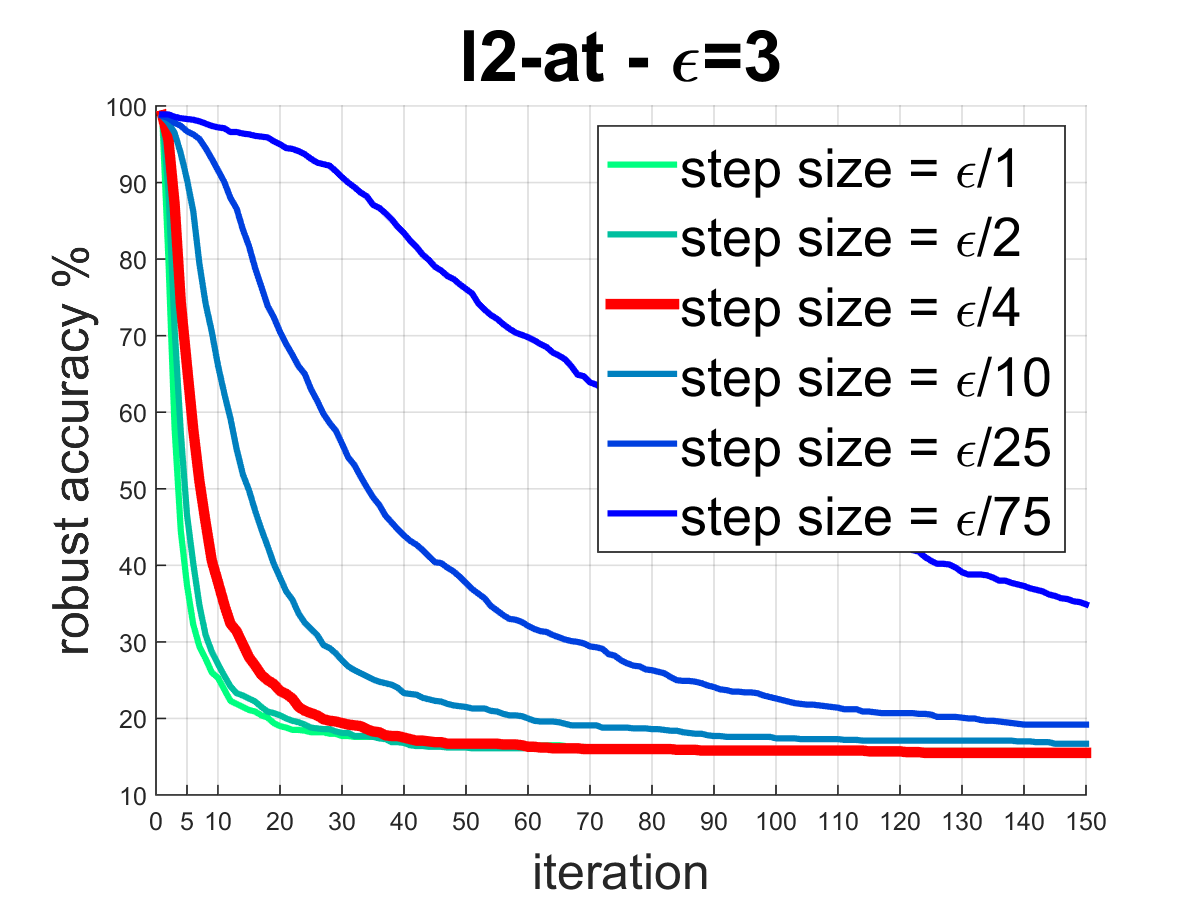}}\\
     \textbf{MNIST - $l_1$-attack:} \quad %\\[2mm]
		\raisebox{-.5\height}{\includegraphics[width=0.5\columnwidth, clip, trim=10mm 0mm 15mm 0mm]{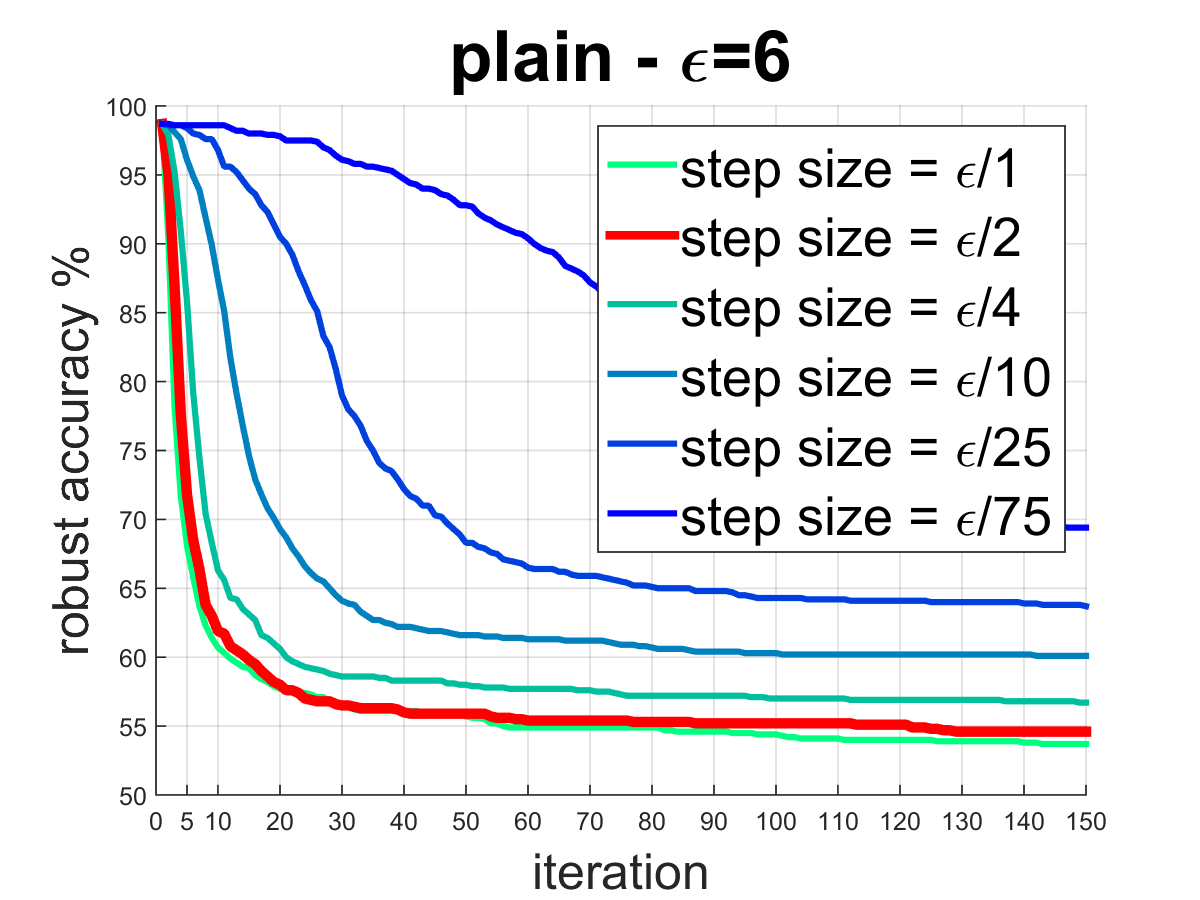}} 
		\raisebox{-.5\height}{\includegraphics[width=0.5\columnwidth, clip, trim=10mm 0mm 15mm 0mm]{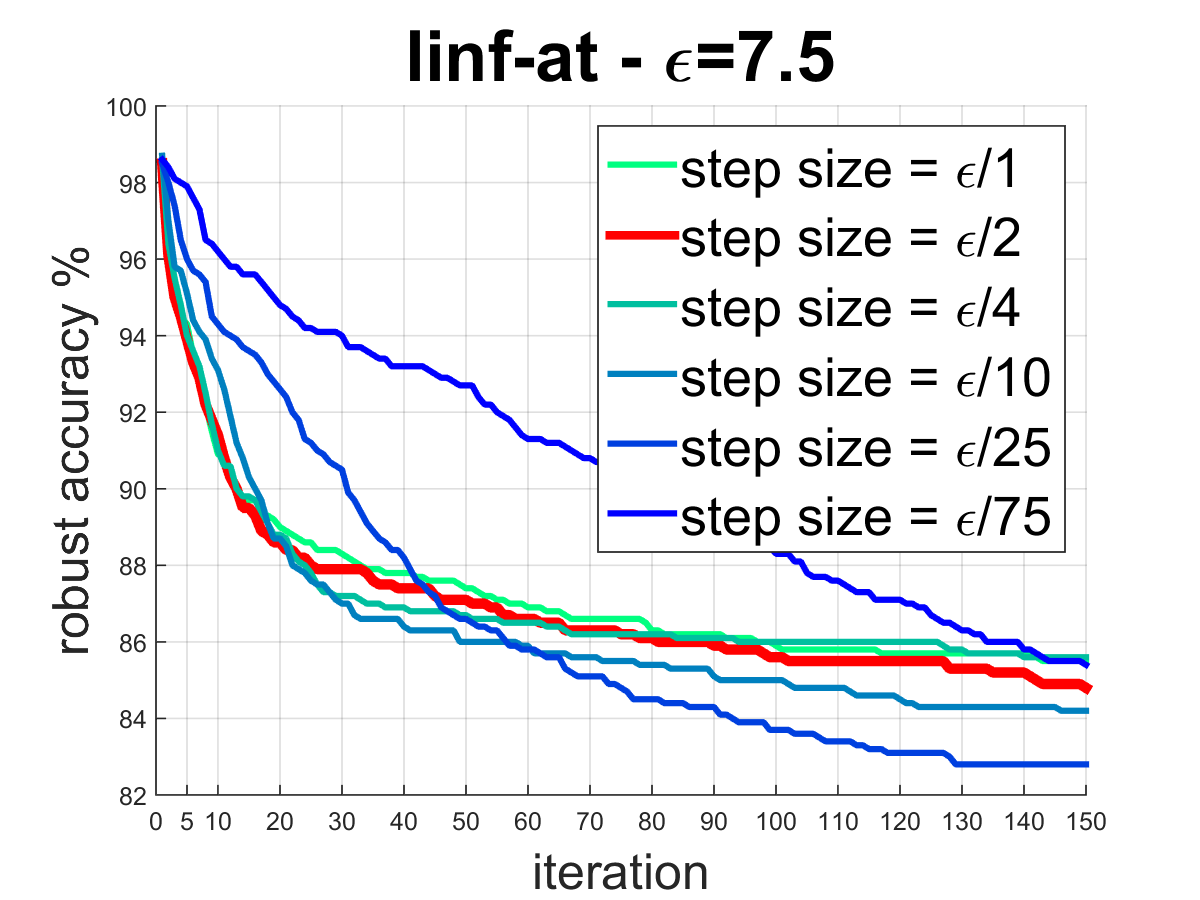}}
		\raisebox{-.5\height}{\includegraphics[width=0.5\columnwidth, clip, trim=10mm 0mm 15mm 0mm]{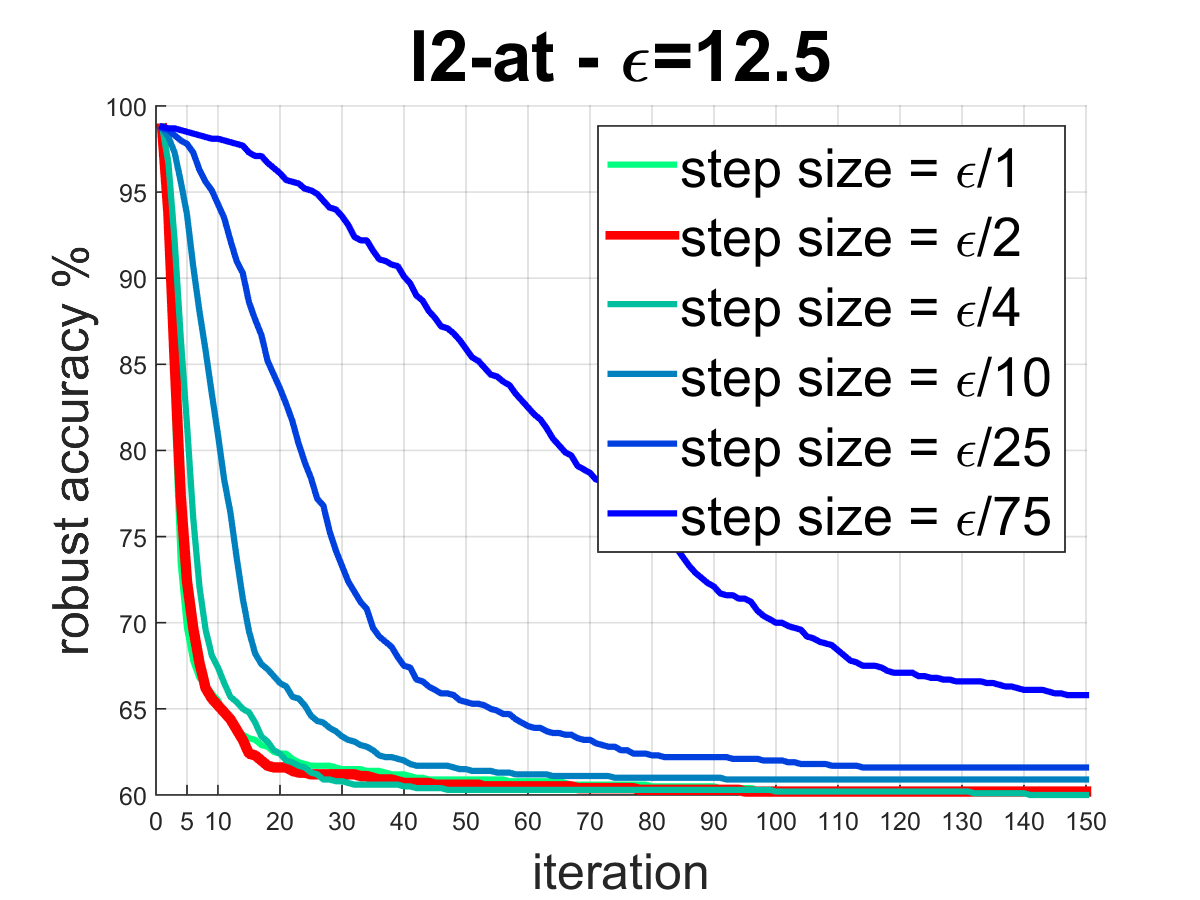}}\\
		%\hline
		\end{tabular}
		\caption{Evolution of robust accuracy across iterations for different step sizes for PGD wrt $l_1,l_2,l_\infty$ for three different models on MNIST. In red  we highlight the step size we used for each norm in the experiments. Notice that it performs on average the best. Here we evaluate on MNIST - the supplementary material contains also CIFAR-10 and other thresholds. \label{fig:PGDStepsizeShort}}
\end{figure*}

\begin{table*}[t]
	\centering
	%\vspace{-5mm}
	\extrarowheight=0.3mm
	{%\scriptsize
		\small
		\caption{\label{tab:perf_2}Performance summary of all attacks on MNIST and CIFAR-10 (aggregated).
			We report, for each norm, "avg. rob. acc.",  the mean of robust accuracies across all models and datasets (lower is better), "\# best", number of times the attack is the best one, "avg. diff. to best" and "max diff. to best", the mean and maximum difference of the robust accuracy of the attack to the robust
			accuracy of the best attack for each model/threshold (on the first 1000 points for $l_\infty$ and $l_1$, 500 for $l_2$, of the test sets). The numbers after the name of the attacks indicate the number of restarts.
			In total we have 5 thresholds $\times$ 6 models = 30 cases for each of the 3 norms.
			*Note that for FAB-10 (i.e. with 10 restarts) the "\# best" is computed excluding the results of FAB-100.
	} \vspace{2mm}
		\begin{tabular}{L{25mm}*{6}{|C{14mm} }}
			\multicolumn{7}{c}{\textbf{statistics on MNIST + CIFAR-10}}\\[2mm]
			\cellcolor{gray!20} \textbf{$l_\infty$-norm}& &DF &   DAA-50 &   PGD-100 &   FAB-10 & FAB-100\\
			\hline
			avg. rob. acc. &&58.81&60.67&46.07&46.18&\textbf{45.47}\\
			\# best &&0&8&12&13*&\textbf{17}\\	
			avg. diff. to best & &14.58&16.45&1.85&1.96&\textbf{1.25}\\	
			max diff. to best & &78.10&49.00&\textbf{10.70}&20.30&17.10\\
			\hline
			\hline
			%\multicolumn{7}{c}{}\\
			\cellcolor{gray!20}\textbf{$l_2$-norm}& CW&DF &    LRA&   PGD-100 &   FAB-10 & FAB-100\\
			\hline
			avg. rob. acc. &45.09&56.10&36.97&44.94&36.41&\textbf{35.57}\\\# best &4&1&9&11&19*&\textbf{23}\\avg. diff. to best &9.65&20.67&1.54&9.51&0.98&\textbf{0.13}\\max diff. to best &65.40&91.40&13.60&64.80&8.40&\textbf{1.60}\\
			\hline
			\hline
			%\multicolumn{7}{c}{}\\
			\cellcolor{gray!20}\textbf{$l_1$-norm}& &SF &    EAD&   PGD-100 &   FAB-10 & FAB-100\\
			\hline
			avg. rob. acc. &&64.47&35.79&49.51&33.26&\textbf{29.46}\\\# best &&0&13&0&10*&\textbf{17}\\avg. diff. to best &&35.31&6.63&20.35&4.10&\textbf{0.30}\\max diff. to best &&95.90&58.40&74.00&21.80&\textbf{1.60}\\

\end{tabular}

	}%
	
	\end{table*}

\begin{table*}[h!]
	\centering
	%\vspace{-5mm}
	\extrarowheight=0.5mm
	{%\scriptsize
		\small
		\caption{\label{tab:perf_4}As in Table \ref{tab:perf_2} statistics of the performance of different attacks on Restricted ImageNet (on the first 500 points of the validation set). In total we consider 5 thresholds $\times$ 3 models = 15 cases for each of the 3 norms.} \vspace{2mm}
		\begin{tabular}{L{20mm} |*{4}{|C{7mm}} | *{3}{|C{7mm}} |*{3}{|C{7mm} }}
			\multicolumn{11}{c}{\textbf{statistics on Restricted ImageNet}}\\[2mm]
			\multicolumn{1}{c}{}& \multicolumn{4}{c||}{\cellcolor{gray!20} \textbf{$l_\infty$-norm}} &\multicolumn{3}{c||}{\cellcolor{gray!20} \textbf{$l_2$-norm}} & \multicolumn{3}{c}{\cellcolor{gray!20} \textbf{$l_1$-norm}}\\
			%\multicolumn{1}{c}{}\\
			& DF &   DAA-10 &   PGD-10 &   FAB-10 & DF & PGD-10 &   FAB-10 & SF & PGD-5 &   FAB-5\\ 
			\hline avg. rob. acc. &35.61&38.44&\textbf{26.91}&27.83 & 45.69&\textbf{31.75}&33.24 &71.31&40.64&\textbf{38.12}\\
			\# best &0&1&\textbf{13}&3& 0&\textbf{14}&1 & 0&3&\textbf{12}\\
			avg. diff. best &8.75&11.57&\textbf{0.04}&0.96&13.99&\textbf{0.04}&1.53&33.52&2.85&\textbf{0.33}\\
			max diff. best &14.60&37.20&\textbf{0.40}&2.00&25.40&\textbf{0.60}&3.40 & 59.00&6.20&\textbf{2.40}\\
	 \end{tabular}
	}%
	%\vspace{-2mm}
	\end{table*}

\paragraph{Attacks:}
We compare the performance of FAB-attack\footnote{\url{https://github.com/fra31/fab-attack}} to those of attacks representing the state-of-the-art in each norm: DeepFool (DF) \cite{MooFawFro2016}, Carlini-Wagner $l_2$-attack (CW) \cite{CarWag2016}, Linear Region $l_2$-Attack (LRA) \cite{CroRauHei2019}, 
Projected Gradient Descent on the cross-entropy function (PGD) \citep{KurGooBen2016a,MadEtAl2018,TraBon2019}, Distributionally Adversarial Attack (DAA) \cite{ZheEtAl2019}, SparseFool (SF) \cite{ModEtAl19}, Elastic-net Attack (EAD) \cite{CheEtAl2018}. 
We use DF from \cite{foolbox}, CW and EAD as in \cite{Cleverhans2017},
DAA and LRA with the code from the original papers, while we reimplemented SF and PGD.
For MNIST and CIFAR-10 we used DAA with 50 restarts, PGD and FAB with 100 restarts.
For  Restricted ImageNet, we used DAA, PGD and FAB with 10 restarts (for $l_1$ we used  5 restarts, since the methods benefit from more iterations).
Moreover, we could not use LRA since it hardly scales to models of such scale and CW and EAD for compatibility issues between the implementations of  attacks and  models.
See \ifpaper \ref{sec:app_attacks} \else the supplementary material \fi for more details e.g. regarding number of iterations and hyperparameters of all attacks. In particular, we provide \ifpaper in Section \ref{sec:app_pgd_step_size} \fi a detailed analysis of the dependency of PGD on the step size. Indeed the optimal
choice of the step size is quite important for PGD.
In order to select the optimal step size for PGD for each norm, we performed a grid search on the step size parameter in
$\nicefrac{\epsilon}{t}$ for $t\in\{1, 2, 4, 10, 25, 75\}$ for different models and thresholds, and took the values working best on average, see Figure \ref{fig:PGDStepsizeShort}
%and the supplementary material
for an illustration (similar plots for other datasets and thresholds are presented in \ifpaper Section \ref{sec:app_pgd_step_size}\else the supplements\fi). As a result we use for PGD wrt $l_\infty$ step size $\nicefrac{\epsilon}{10}$ and the direction is the sign of the gradient of the cross entropy loss, for PGD wrt $l_2$ we do a step in the direction of the $l_2$-normalized gradient with step size $\nicefrac{\epsilon}{4}$, for PGD wrt $l_1$ we use the gradient step suggested in \cite{TraBon2019} (with sparsity levels of 1\% for MNIST and 10\% for CIFAR-10 and Restricted ImageNet) with step size $\nicefrac{\epsilon}{2}$.
For FAB-attack we use always $\beta=0.9$ and on MNIST and CIFAR-10: $\alpha_\textrm{max}=0.1$, $\eta=1.05$ and on Restricted ImageNet: $\alpha_{\max}=0.05$, $\eta=1.3$. These parameters are the same for all norms.

%ensure we provided a fair comparison with a strong competitor.

\paragraph{Evaluation metrics:}
The \textit{robust accuracy} for a threshold $\epsilon$ is the classification accuracy (in percentage) when an adversary is allowed to change every test input with perturbations of $l_p$-norm smaller than $\epsilon$ in order to change the decision. Thus stronger attacks produce lower robust accuracies.
For each model and dataset we fix five thresholds at which we compute the robust accuracy for each attack (we choose the thresholds so that the robust accuracy covers the range between clean accuracy and 0). We evaluate the attacks by the following statistics:
i) \textbf{avg. rob. accuracy}: the mean of the robust accuracies achieved by the attack over all models and thresholds (lower is better), ii) \textbf{\# best}: how many times the attack achieves the lowest robust accuracy (it is the most effective),
iii) \textbf{avg. difference to best}: for each model/threshold we compute the difference between the robust accuracy of the attack and the best one across all the attacks, then we average over all models/thresholds,
iv) \textbf{max difference to best}: as "avg. difference to best", but with the maximum difference instead of the average one.
In \ifpaper Section \ref{sec:app_results} \else the supplement \fi we report additionally the average $l_p$-norm of the adversarial perturbations given by the attacks.

\paragraph{Results:} We report the complete results \ifpaper in Tables \ref{tab:MNIST_plain} to \ref{tab:ImageNet_l2} of the appendix\else in the supplementary material\fi, while we summarize them in Table \ref{tab:perf_2} (MNIST and CIFAR-10 aggregated, as we used the same attacks) and Table \ref{tab:perf_4} (Restricted ImageNet). Our FAB-attack achieves the best results in all statistics for every norm (with the only exception of "max diff. to best" in $l_\infty$) on MNIST+CIFAR-10. In particular, while on $l_\infty$ the "avg. robust accuracy" of PGD is not far from that of FAB, the gap is large when considering $l_2$ and $l_1$ (in the appendix we provide \ifpaper  in Table \ref{tab:perf_withoutMadry} \fi the aggregate statistics without the result on $l_\infty$-AT 
MNIST, showing that FAB is still the best attack even if one leaves out this failure case of PGD).
Interestingly, the second best attack in terms of average robust accuracy, is different for every norm (PGD for $l_\infty$, LRA for $l_2$, EAD for $l_1$), which implies that FAB outperforms algorithms specialized in the individual norms.\\
We also report the results of FAB-10, that is our attack with only 10 restarts, to show that FAB yields high quality results already with a low budget in terms of time/computational cost. In fact, FAB-10 has "avg. robust accuracy" better than or very close to that of the strongest versions of the other attacks (see below for a runtime analysis, where one observes that FAB-10 is the fastest attack when excluding the significantly worse DF and SF attacks).
On Restricted ImageNet, FAB-attack gets the best results in all statistics for $l_1$, while for $l_\infty$ and $l_2$ PGD performs on average better, but the difference in "avg. robust accuracy" is small.\\
In general, both average and maximum \textit{difference to best} of FAB-attack are small for all the datasets and norms, implying that it does not suffer from severe failures, which makes it an efficient, high quality technique to evaluate the robustness of classifiers for all $l_p$-norms.
Finally, we show in \ifpaper Table \ref{tab:avg_dist} \else the supplementary material \fi that FAB-attack outperforms or matches the competitors in 16 out of 18 cases when comparing the average $l_p$-norms of the generated adversarial perturbations.\\

\begin{table}[h] \vspace{-5mm}\centering {\small \caption{We attack the ResNet-110 in \cite{pang2020rethinking} on CIFAR-10 at $\epsilon=\nicefrac{8}{255}$. The performance of the PGD attack on the cross entropy loss (CE) heavily depends on both scale of the classifier and the step size. In contrast, the scaling invariant FAB-attack works well even on the original (unscaled) model.} \label{tab:grad_obf_below}
		\vspace{2mm}
		\begin{tabular}{l r r}
		\textit{attack} & \textit{step size} & \textit{robust accuracy}\\
		\hline
		PGD-CE & $\nicefrac{\epsilon}{10}$ &90.2\% \\	
		PGD-CE & $\nicefrac{\epsilon}{2}$ &90.2\% \\ PGD-CE rescaled & $\nicefrac{\epsilon}{10}$ &12.9\% \\ PGD-CE rescaled & $\nicefrac{\epsilon}{2}$ & 2.5\%\\
		FAB & - & \textbf{0.3\%}\\
		\hline
\end{tabular}}\end{table}
\commentout{
\begin{table}[b]
	\centering
	{\small \caption{Importance of scale invariance and absence of step size.} \label{tab:grad_obf_above}
	\begin{tabular}{l r r}
		\textit{attack} & \textit{step size} & \textit{robust accuracy}\\
		\hline
		PGD-CE & $\epsilon/10$ &90.2\% \\	
		PGD-CE & $\epsilon/2$ &90.2\% \\ PGD-CE rescaled & $\epsilon/10$ &12.9\% \\ PGD-CE rescaled & $\epsilon/2$ & 2.5\%\\
		FAB & - & \textbf{0.3\%}\\
		\hline
		\end{tabular}} \end{table}}

\paragraph{Resistance to gradient masking:} It has been argued \cite{TraBon2019} that models trained with first-order methods to be resistant wrt $l_\infty$-attacks on MNIST (adv. training) give a false sense of robustness in $l_1$ and $l_2$ due to \textit{gradient masking}. This means that standard gradient-based methods like PGD have problems to find adversarial examples while they still exist. In contrast, FAB %, while relies on first-order approximations of the classifier, 
does not suffer from gradient masking. In Table \ifpaper \ref{tab:MNIST_linf} of the appendix\else 8 (supplement) \fi we see that it is extremely effective also wrt $l_1$ and $l_2$ on the $l_\infty$-robust model, outperforming by a large margin the competitors. The reason is that FAB is not dependent on the norm of the gradient but just its direction matters for the definition of the hyperplane in \eqref{eq:lin_db}. 
While we believe that resistance to gradient masking is a key property of a solid attack,
%as an additional sanity check,
we recompute 
%the statistics mentioned
the statistics of Table \ref{tab:perf_2} excluding $l_1$ and $l_2$ attacks on the $l_\infty$-AT model on MNIST \ifpaper in Table \ref{tab:perf_withoutMadry} in the appendix\else (see supplements)\fi. FAB still achieves in most of the cases the best aggregated statistics, implying that our attack is effective whether or not the attacked classifier tends to "mask" the gradient.
\\
We have shown in Section \ref{sec:scale_inv} that FAB-attack is invariant under rescaling of the classifier. We provide an example why this is a desirable
property of an adversarial attack. 
We consider the defense proposed in \cite{pang2020rethinking}, in particular their ResNet-110 (without adversarial training) for CIFAR-10. In Table 5 in \cite{pang2020rethinking} it is claimed that this model has a robust accuracy of 31.4\% for $\nicefrac{8}{255}$ obtained by a PGD attack on their new loss function. They say that a standard PGD attack on the cross-entropy loss performs much worse. We test the performance of PGD on the cross-entropy loss,
both using the original classifier and the same
scaled down by a factor of $10^6$. Moreover, we use the default step size $\nicefrac{\epsilon}{10}$ together with $\nicefrac{\epsilon}{2}$. The results are reported in Table \ref{tab:grad_obf_below}. We can see that PGD on the original model yields more than 90\% robust accuracy which confirms the statement in \cite{pang2020rethinking}
about the cross-entropy loss being unsuitable for this case.
However, PGD applied to the rescaled classifier reduces robust accuracy below 13\%. The better step size $\nicefrac{\epsilon}{2}$ decreases it to 2.5\% which shows that tuning the stepsize is important for PGD. At the same time, FAB achieves a robust accuracy of 0.3\% without any need of parameter tuning %and without
or rescaling of the classifier. This exemplifies the benefit of the scaling invariance of FAB. Moreover, as a side result this shows that the new loss alone in \cite{pang2020rethinking} is an ineffective defense.

\paragraph{Runtime comparison:}
DF and SF are much faster than the other attacks as their primary goal is to find as fast as possible adversarial examples, without emphasis on minimizing their norms, while LRA is rather expensive as noted in the original paper.
PGD needs a forward and a backward pass of the network per iteration whereas FAB requires three passes for each iteration. Thus PGD is given 1.5 times more iterations than FAB, so that overall they have same budget of forward/backward passes (and thus runtime).
Below we report the runtimes (for 1000 points on MNIST and CIFAR-10, 50 on R-ImageNet) for the attacks as used in the experiments (if not specified otherwise, it includes all the restarts). For PGD and DAA this is the time for evaluating the robust accuracy at 5 thresholds, while for the other methods a single run is sufficient to compute the robust accuracy for all five thresholds. 
\textbf{MNIST}: DAA 11736s, PGD 3825s for $l_\infty$/$l_2$ and 14106s for $l_1$, CW 944s, EAD 606s, FAB-10 161s, FAB-100 1613s.
\textbf{CIFAR-10}: DAA 11625s, PGD 31900s for $l_\infty$/$l_2$ and 70110s for $l_1$, CW 3691s, EAD 3398s, FAB-10 1209s, FAB-100 12093s.
\textbf{R-ImageNet}: DAA 6890s, PGD 4738s for $l_\infty$/$l_2$ and 24158s for $l_1$, FAB 2268s for $l_\infty$/$l_2$ and 3146s for $l_1$  (note that for $l_1$ different numbers of restarts/iterations are used on R-ImageNet).\\
We note that for PGD the robust accuracy for the five thresholds can be computed faster by exploiting the fact that points which are non-robust
for a thresholds $\epsilon$ are also non-robust for thresholds larger than $\epsilon$. However, even when taking this into account FAB-10 would still be significantly faster than
PGD-100 and has better quality on MNIST and CIFAR-10. Moreover, when just considering a fixed number of thresholds, one can stop FAB-attack whenever it finds
an adversarial example for the smallest threshold which also leads to a speed-up. However, in real world applications a full picture of robustness
as a continuous function of the threshold is the most interesting evaluation scenario.

\subsection{Additional results} \label{sec:add_results}
In \ifpaper Section~\ref{sec:evol_iter} \else the supplementary materials \fi we show how the robust accuracy provided by either PGD or FAB-attack evolves over iterations, when only one start it used. In particular, we compare the two methods when the same number of passes, forward or backward, of the networks are used. One can observe that a few steps are usually sufficient for FAB-attack to achieve good results, often faster than PGD, although there are cases where a higher number of iterations leads to significantly better robust accuracy.

Finally, \cite{croce2020reliable} use FAB-attack together with other white- and black-box attacks to evaluate the robustness of over 50 classifiers trained with recently proposed adversarial defenses wrt $l_\infty$ and $l_2$ on different datasets. With fixed hyperparameters, FAB-attack yields the best results in most of the cases on CIFAR-10, CIFAR-100 and ImageNet in both norms, in particular compared to different variations of PGD (with and without a momentum term, with different step sizes and using various losses). This shows again that FAB-attack is very effective for testing the robustness of adversarial defenses.

\section{FAB-attack with a large number of classes} \label{sec:targeted_version}
The standard algorithm of FAB-attack requires to compute at each iteration the Jacobian matrix of the classifier $f$ wrt the input $x$ %which has dimension $K \times d$ (recall $K$ is the number of classes and $d$ the input dimension),
and then the closest approximated decision hyperplane. The Jacobian matrix has dimension $K \times d$, recalling that $K$ is the number of classes and $d$ the input dimension. Although this can be in principle obtained with a single backward pass of the network, it becomes computationally expensive on datasets with many classes. Moreover, the memory consumption of FAB-attack increases with $K$. As a consequence, using FAB-attack in the normal formulation on datasets like ImageNet which has $K=1000$ classes may be inefficient.

Then, we propose a \emph{targeted} version of our attack which performs at each iteration the projection onto the linearized decision boundary between the original class and a fixed target class. This means that in \eqref{eq:iter_step_noextra} the hyperplane $\pi_s$ is not selected via \eqref{eq:min_dist} as the closest one to the current iterate but rather $s\equiv t$, with $t$ the target class used. Note that in practice we do not constrain the final outcome of the algorithm to be assigned to class $t$, but any misclassification is sufficient to have a valid adversarial example. The target class $t$ is selected as the second most likely one according to the score given by the model to the target point, and if $k$ restarts are allowed one can use the classes with the $k+1$ highest scores as target (excluding the correct one $c$). In this way, only the gradient of $f_t - f_c : \R^d \rightarrow \R$ needs to be computed, which is a cheaper operation than getting the full Jacobian of $f$ and with computational cost independent of the total number of classes.

This targeted version of FAB-attack has been used in  \cite{croce2020reliable} considering the top-10 classes where it yields on CIFAR-10, CIFAR-100 and ImageNet  almost always better robust accuracy than normal FAB-attack, which instead is almost always better on MNIST. 

\section{Conclusion}
In summary, our geometrically motivated FAB-attack outperforms in terms of average quality the state-of-the-art attacks, already with a limited computational effort, and works for all $l_p$-norms in $p \in \{1,2,\infty\}$ unlike most competitors. Thanks to its scaling invariance and being step size free it is resistant to gradient masking and thus more reliable for assessing robustness than the standard PGD attack.

\section*{Acknowledgements}
We acknowledge support from the German Federal Ministry of Education and Research (BMBF) through the T\"{u}bingen AI Center (FKZ: 01IS18039A). This work was also supported by the DFG Cluster of Excellence “Machine Learning – New Perspectives for Science”, EXC 2064/1, project number 390727645, and by DFG grant 389792660 as part of TRR~248.

%\clearpage
%\newpage

\bibliographystyle{icml2020}
%\bibliography{Literatur}

\begin{thebibliography}{33}
\providecommand{\natexlab}[1]{#1}
\providecommand{\url}[1]{\texttt{#1}}
\expandafter\ifx\csname urlstyle\endcsname\relax
  \providecommand{\doi}[1]{doi: #1}\else
  \providecommand{\doi}{doi: \begingroup \urlstyle{rm}\Url}\fi

\bibitem[Athalye et~al.(2018)Athalye, Carlini, and Wagner]{AthEtAl2018}
Athalye, A., Carlini, N., and Wagner, D.~A.
\newblock Obfuscated gradients give a false sense of security: Circumventing
  defenses to adversarial examples.
\newblock In \emph{ICML}, 2018.

\bibitem[Bastani et~al.(2016)Bastani, Ioannou, Lampropoulos, Vytiniotis, Nori,
  and Criminisi]{BasEtAl2016}
Bastani, O., Ioannou, Y., Lampropoulos, L., Vytiniotis, D., Nori, A., and
  Criminisi, A.
\newblock Measuring neural net robustness with constraints.
\newblock In \emph{NeurIPS}, 2016.

\bibitem[Brendel et~al.(2018)Brendel, Rauber, and Bethge]{BreRauBet18}
Brendel, W., Rauber, J., and Bethge, M.
\newblock Decision-based adversarial attacks: Reliable attacks against
  black-box machine learning models.
\newblock In \emph{ICLR}, 2018.

\bibitem[Brown et~al.(2017)Brown, Man{\'e}, Roy, Abadi, and
  Gilmer]{BroEtAl2017}
Brown, T.~B., Man{\'e}, D., Roy, A., Abadi, M., and Gilmer, J.
\newblock Adversarial patch.
\newblock In \emph{NeurIPS 2017 Workshop on Machine Learning and Computer
  Security}, 2017.

\bibitem[Carlini \& Wagner(2017{\natexlab{a}})Carlini and Wagner]{CarWag2016}
Carlini, N. and Wagner, D.
\newblock Towards evaluating the robustness of neural networks.
\newblock In \emph{IEEE Symposium on Security and Privacy}, 2017{\natexlab{a}}.

\bibitem[Carlini \& Wagner(2017{\natexlab{b}})Carlini and Wagner]{CarWag2017}
Carlini, N. and Wagner, D.
\newblock Adversarial examples are not easily detected: Bypassing ten detection
  methods.
\newblock In \emph{ACM Workshop on Artificial Intelligence and Security},
  2017{\natexlab{b}}.

\bibitem[Chen et~al.(2018)Chen, Sharma, Zhang, Yi, and Hsieh]{CheEtAl2018}
Chen, P., Sharma, Y., Zhang, H., Yi, J., and Hsieh, C.
\newblock Ead: Elastic-net attacks to deep neural networks via adversarial
  examples.
\newblock In \emph{AAAI}, 2018.

\bibitem[Croce \& Hein(2020)Croce and Hein]{croce2020reliable}
Croce, F. and Hein, M.
\newblock Reliable evaluation of adversarial robustness with an ensemble of
  diverse parameter-free attacks.
\newblock In \emph{ICML}, 2020.

\bibitem[Croce et~al.(2019)Croce, Rauber, and Hein]{CroRauHei2019}
Croce, F., Rauber, J., and Hein, M.
\newblock Scaling up the randomized gradient-free adversarial attack reveals
  overestimation of robustness using established attacks.
\newblock \emph{International J. of Computer Vision (IJCV)}, 2019.

\bibitem[Engstrom et~al.(2017)Engstrom, Tran, Tsipras, Schmidt, and
  Madry]{EngEtAl2017}
Engstrom, L., Tran, B., Tsipras, D., Schmidt, L., and Madry, A.
\newblock A rotation and a translation suffice: Fooling {CNN}s with simple
  transformations.
\newblock In \emph{NeurIPS 2017 Workshop on Machine Learning and Computer
  Security}, 2017.

\bibitem[Gu \& Rigazio(2015)Gu and Rigazio]{GuRig2015}
Gu, S. and Rigazio, L.
\newblock Towards deep neural network architectures robust to adversarial
  examples.
\newblock In \emph{ICLR Workshop}, 2015.

\bibitem[He et~al.(2016)He, Zhang, Ren, and Sun]{HeZhaRen2015}
He, K., Zhang, X., Ren, S., and Sun, J.
\newblock Deep residual learning for image recognition.
\newblock In \emph{CVPR}, pp.\  770--778, 2016.

\bibitem[Hein \& Andriushchenko(2017)Hein and Andriushchenko]{HeiAnd2017}
Hein, M. and Andriushchenko, M.
\newblock Formal guarantees on the robustness of a classifier against
  adversarial manipulation.
\newblock In \emph{NeurIPS}, 2017.

\bibitem[Huang et~al.(2016)Huang, Xu, Schuurmans, and Szepesvari]{HuaEtAl2016}
Huang, R., Xu, B., Schuurmans, D., and Szepesvari, C.
\newblock Learning with a strong adversary.
\newblock In \emph{ICLR}, 2016.

\bibitem[Katz et~al.(2017)Katz, Barrett, Dill, Julian, and
  Kochenderfer]{KatzEtAl2017}
Katz, G., Barrett, C., Dill, D., Julian, K., and Kochenderfer, M.
\newblock Reluplex: An efficient smt solver for verifying deep neural networks.
\newblock In \emph{CAV}, 2017.

\bibitem[Krizhevsky et~al.(2014)Krizhevsky, Nair, and Hinton]{CIFAR10}
Krizhevsky, A., Nair, V., and Hinton, G.
\newblock Cifar-10 (canadian institute for advanced research).
\newblock 2014.
\newblock URL \url{http://www.cs.toronto.edu/~kriz/cifar.html}.

\bibitem[Kurakin et~al.(2017)Kurakin, Goodfellow, and Bengio]{KurGooBen2016a}
Kurakin, A., Goodfellow, I.~J., and Bengio, S.
\newblock Adversarial examples in the physical world.
\newblock In \emph{ICLR Workshop}, 2017.

\bibitem[Madry et~al.(2018)Madry, Makelov, Schmidt, Tsipras, and
  Valdu]{MadEtAl2018}
Madry, A., Makelov, A., Schmidt, L., Tsipras, D., and Valdu, A.
\newblock Towards deep learning models resistant to adversarial attacks.
\newblock In \emph{ICLR}, 2018.

\bibitem[Modas et~al.(2019)Modas, Moosavi-Dezfooli, and Frossard]{ModEtAl19}
Modas, A., Moosavi-Dezfooli, S., and Frossard, P.
\newblock Sparsefool: a few pixels make a big difference.
\newblock In \emph{CVPR}, 2019.

\bibitem[Moosavi-Dezfooli et~al.(2016)Moosavi-Dezfooli, Fawzi, and
  Frossard]{MooFawFro2016}
Moosavi-Dezfooli, S.-M., Fawzi, A., and Frossard, P.
\newblock Deepfool: a simple and accurate method to fool deep neural networks.
\newblock In \emph{CVPR}, pp.\  2574--2582, 2016.

\bibitem[Mosbach et~al.(2018)Mosbach, Andriushchenko, Trost, Hein, and
  Klakow]{MosEtAl18}
Mosbach, M., Andriushchenko, M., Trost, T., Hein, M., and Klakow, D.
\newblock Logit pairing methods can fool gradient-based attacks.
\newblock In \emph{NeurIPS 2018 Workshop on Security in Machine Learning},
  2018.

\bibitem[Narodytska \& Kasiviswanathan(2016)Narodytska and
  Kasiviswanathan]{NarKas16}
Narodytska, N. and Kasiviswanathan, S.~P.
\newblock Simple black-box adversarial perturbations for deep networks.
\newblock In \emph{CVPR 2017 Workshops}, 2016.

\bibitem[Pang et~al.(2020)Pang, Xu, Dong, Du, Chen, and
  Zhu]{pang2020rethinking}
Pang, T., Xu, K., Dong, Y., Du, C., Chen, N., and Zhu, J.
\newblock Rethinking softmax cross-entropy loss for adversarial robustness.
\newblock In \emph{ICLR}, 2020.

\bibitem[Papernot et~al.(2016)Papernot, McDonald, Wu, Jha, and
  Swami]{PapEtAl2016a}
Papernot, N., McDonald, P., Wu, X., Jha, S., and Swami, A.
\newblock Distillation as a defense to adversarial perturbations against deep
  networks.
\newblock In \emph{IEEE Symposium on Security \& Privacy}, 2016.

\bibitem[Papernot et~al.(2017)Papernot, Carlini, Goodfellow, Feinman, Faghri,
  Matyasko, Hambardzumyan, Juang, Kurakin, Sheatsley, Garg, and
  Lin]{Cleverhans2017}
Papernot, N., Carlini, N., Goodfellow, I., Feinman, R., Faghri, F., Matyasko,
  A., Hambardzumyan, K., Juang, Y.-L., Kurakin, A., Sheatsley, R., Garg, A.,
  and Lin, Y.-C.
\newblock cleverhans v2.0.0: an adversarial machine learning library.
\newblock preprint, arXiv:1610.00768, 2017.

\bibitem[Rauber et~al.(2017)Rauber, Brendel, and Bethge]{foolbox}
Rauber, J., Brendel, W., and Bethge, M.
\newblock Foolbox: A python toolbox to benchmark the robustness of machine
  learning models.
\newblock In \emph{ICML Reliable Machine Learning in the Wild Workshop}, 2017.

\bibitem[Su et~al.(2019)Su, Vargas, and Kouichi]{SuVarKou19}
Su, J., Vargas, D.~V., and Kouichi, S.
\newblock One pixel attack for fooling deep neural networks.
\newblock \emph{IEEE Transactions on Evolutionary Computation}, 23:\penalty0
  828--841, 2019.

\bibitem[Tjeng et~al.(2019)Tjeng, Xiao, and Tedrake]{TjeTed2017}
Tjeng, V., Xiao, K., and Tedrake, R.
\newblock Evaluating robustness of neural networks with mixed integer
  programming.
\newblock In \emph{ICLR}, 2019.

\bibitem[Tramèr \& Boneh(2019)Tramèr and Boneh]{TraBon2019}
Tramèr, F. and Boneh, D.
\newblock Adversarial training and robustness for multiple perturbations.
\newblock In \emph{NeurIPS}, 2019.

\bibitem[Tsipras et~al.(2019)Tsipras, Santurkar, Engstrom, Turner, and
  Madry]{TsiEtAl18}
Tsipras, D., Santurkar, S., Engstrom, L., Turner, A., and Madry, A.
\newblock Robustness may be at odds with accuracy.
\newblock In \emph{ICLR}, 2019.

\bibitem[Wong et~al.(2019)Wong, Schmidt, and Kolter]{WonEtAl2019}
Wong, E., Schmidt, F.~R., and Kolter, J.~Z.
\newblock Wasserstein adversarial examples via projected sinkhorn iterations.
\newblock In \emph{ICML}, 2019.

\bibitem[Zheng et~al.(2016)Zheng, Song, Leung, and Goodfellow]{ZheEtAl2016}
Zheng, S., Song, Y., Leung, T., and Goodfellow, I.~J.
\newblock Improving the robustness of deep neural networks via stability
  training.
\newblock In \emph{CVPR}, 2016.

\bibitem[Zheng et~al.(2019)Zheng, Chen, and Ren]{ZheEtAl2019}
Zheng, T., Chen, C., and Ren, K.
\newblock Distributionally adversarial attack.
\newblock In \emph{AAAI}, 2019.

\end{thebibliography}

\ifpaper
\clearpage
\appendix
\section{Scale invariance of FAB-attack}
\label{sec:app_scale_inv}
We here describe the proof of Proposition \ref{prop:scale_inv}.

\begin{customprop}{\ref{prop:scale_inv}} Let $f:\R^d \rightarrow \R^K$ be a classifier. Then for any $\alpha>0$ and $\beta\in\R$ the output $x_\textrm{out}$  of Algorithm \ref{alg:algorithm-label}
	for the classifier $f$ is the same as of the classifiers $g=\alpha f$ and $h =f+\beta$. \end{customprop}

\begin{proof} It holds $\nabla g_i = \nabla \alpha f_i = \alpha \nabla f_i$ and $\nabla h_i = \nabla(f_i + \beta) = \nabla f_i$
	for $i=1, \ldots, K$ and thus the definition of the hyperplane $\pi_l(z)$ in \eqref{eq:lin_db} is not affected by rescaling or shifting. The same holds for the distance of $x^{(i)}$ to the hyperplane $\pi_l(z)$ in \eqref{eq:min_dist}. Note that the rest of the steps just depends on geometric quantities derived from these quantities and are independent of $f$. Finally, also the iterates of the final
	search in \eqref{eq:linear_approx_search_2} are invariant under rescaling and shifting and thus the output $x_\textrm{out}$  of Algorithm \ref{alg:algorithm-label} is invariant
	under rescaling of $f$. \end{proof}

\section{Experiments}\label{sec:app_exps}
%\paragraph{Models:}
\subsection{Models}\label{sec:app_models}
The \textit{plain} and $l_\infty$-AT models on MNIST are those available at \url{https://github.com/MadryLab/mnist_challenge} and consist of two convolutional and two fully-connected layers. The architecture of the CIFAR-10 models has 8 convolutional layers (with number of filters increasing from 96 to 384) and 2 dense layers and we make the classifiers available at \url{https://github.com/fra31/fab-attack}, while on Restricted ImageNet we use the models (ResNet-50 \cite{HeZhaRen2015}) from \cite{TsiEtAl18} and available at \url{https://github.com/MadryLab/robust-features-code}.\\
The models on MNIST achieve the following clean accuracy: \textit{plain} 98.7\%, $l_\infty$-AT 98.5\%, $l_2$-AT 98.6\%. The models on CIFAR-10 achieve the following clean accuracy: \textit{plain} 89.2\%, $l_\infty$-AT 79.4\%, $l_2$-AT 81.2\%.\\
%The architecture used on Restricted ImageNet is ResNet-50 \cite{HeZhaRen2015}.\\

%\paragraph{Attacks:}
\subsection{Attacks}\label{sec:app_attacks}
We use CW with 40 binary search steps, 10000  iterations and confidence 0, EAD with 1000 iterations, $l_1$ decision rule and $\beta=0.05$. In both cases we set the parameters to achieve minimally (wrt $l_2$ for CW and $l_1$ for EAD) distorted adversarial examples.
%with 10000 and 1000 iterations respectively, CW with confidence 0 and EAD with the $l_1$ decision rule and $\beta=0.05$ to achieve the minimally distorted adversarial examples
We could not use these methods on Restricted ImageNet since, to be compatible with the attacks from \cite{Cleverhans2017}, it would be necessary to reimplement from scratch the models of \cite{TsiEtAl18}, as done in \url{https://github.com/tensorflow/cleverhans/tree/master/cleverhans/model_zoo/madry_lab_challenges} for a similar situation.

For DAA we use 200 iterations for MNIST, 50 for the other datasets and, given a threshold $\epsilon$, a step size of $\nicefrac{\epsilon}{30}$ for MNIST, $\nicefrac{\epsilon}{10}$ otherwise.

We apply PGD with 150 iterations, except for the case of $l_1$ on Restricted ImageNet where we use 450 iterations.
To choose the step size for each norm, we performed a grid search on the step size in
$\nicefrac{\epsilon}{t}$ for $t\in\{1, 2, 4, 10, 25, 75\}$ for different models and took the values working best on average (note that this gives an advantage
to PGD). A more detailed analysis of the dependency of PGD on the step size is given in Section \ref{sec:app_pgd_step_size}. As a result we use for PGD wrt $l_\infty$ a step size of $\nicefrac{\epsilon}{10}$ and the direction is the sign of the gradient of the cross-entropy loss, for PGD wrt $l_2$ we do a step in the direction of the normalized gradient of size $\nicefrac{\epsilon}{4}$, for PGD wrt $l_1$ we use the gradient step suggested in \cite{TraBon2019} (with sparsity levels of 1\% for MNIST and 10\% for CIFAR-10 and Restricted ImageNet) with step size $\nicefrac{\epsilon}{2}$. 

For FAB-attack we set 100 iterations, except for the case of $l_1$ on Restricted ImageNet where we use 300 iterations.
Moreover, we use the following parameters for all the cases on MNIST and CIFAR-10: $\alpha_\textrm{max}=0.1$, $\eta=1.05$, $\beta=0.9$. On Restricted ImageNet we set $\alpha_{\max}=0.05$, $\eta=1.3$, $\beta=0.9$.
When using random restarts, FAB-attack needs a value for the parameters $\epsilon$. It represents the radius of the $l_p$-ball around the original point inside which we sample the starting point of the algorithm, at least until a sufficiently small adversarial perturbation is found (see Algorithm \ref{alg:algorithm-label}). We use the values of $\epsilon$ reported in Table \ref{tab:eps_app}. Note that the attack usually finds at the first run an adversarial perturbation small enough so that $\epsilon$ in practice rarely comes into play.

\begin{table*}[h]
	\centering
	\caption{We report the values of $\epsilon$ used for sampling in case our FAB-attack uses random restarts.}
	\label{tab:eps_app}
	%\vspace{2mm}
	\extrarowheight=0.5mm
	{\footnotesize
		%\small
		%\scriptsize
		%\parbox{.45\linewidth}{
		%\setlength\extrarowheight{1mm}
		%\parbox[c][][c]{100mm}{\textbf{Robust accuracy by FAB-10 with intermediate search}}
		\begin{tabular}{C{5mm} *{3}{|*{3}{R{10mm}}}}
			\multicolumn{10}{c}{\textbf{values of $\epsilon$ used for random restarts}}\\[2mm]
			\multicolumn{1}{c}{}&\multicolumn{3}{c}{MNIST} &\multicolumn{3}{c}{CIFAR-10}&\multicolumn{3}{c}{Restricted ImageNet} \\[2mm]
			& plain & $l_\infty$-AT & $l_2$-AT& plain & $l_\infty$-AT & $l_2$-AT& plain & $l_\infty$-AT & $l_2$-AT\\
			\hline
			$l_\infty$ &0.15&0.3&0.3&0.0& 0.02&0.02&0.02&0.08&0.08\\$l_2$&2.0&2.0&2.0&0.5&4.0&4.0&5.0&5.0&5.0\\$l_1$&40.0&40.0&40.0&10.0&10.0&10.0&100.0&250.0&250.0\\
			\hline
			
		\end{tabular}
	}
\end{table*}

\subsection{Complete results}\label{sec:app_complete_results}
In Tables \ref{tab:MNIST_plain} to \ref{tab:ImageNet_l2} we report the complete values of the robust accuracy, wrt either $l_\infty$, $l_2$ or $l_1$, computed by every attack, for 3 datasets, 3 models for each dataset, 5 thresholds for each model (135 evaluations overall). In Table \ref{tab:perf_withoutMadry}
we provide the aggreate statistics on MNIST and CIFAR-10 without the $l_\infty$-AT model of Madry, as there PGD fails completely for $l_1$- and $l_2$-attacks.
FAB has still the better aggregate statistics if one leaves out these cases.

\begin{table*}[t]
	\centering
	%\vspace{-5mm}
	\extrarowheight=0.3mm
	{%\scriptsize
		\small
		\caption{\label{tab:perf_withoutMadry}Performance summary of all attacks on MNIST and CIFAR-10 (aggregated).
			We report, for each norm, "avg. rob. acc.",  the mean of the robust accuracies across all models and datasets (lower is better), "\# best", number of times the attack is the best one, "avg. diff. to best" and "max diff. to best", the mean and maximum difference of the robust accuracy of the attack to the robust
			accuracy of the best attack for each model/threshold (on the first 1000 points for $l_\infty$ and $l_1$, 500 for $l_2$, of the test sets). The numbers after the name of the attacks indicate the number of restarts.
			In total we have 5 thresholds $\times$ 6 models = 30 cases for each of the 3 norms.
			*Note that for FAB-10 (i.e. with 10 restarts) the "\# best" is computed excluding the results of FAB-100.
			We also recompute the statistics excluding the $l_\infty$-AT model on MNIST from \cite{MadEtAl2018} where PGD fails in order to show 
	that even excluding this case FAB is still the best.
	} \vspace{2mm}
		\begin{tabular}{L{25mm}*{6}{|C{14mm} }}
			\multicolumn{7}{c}{\textbf{statistics on MNIST + CIFAR-10}}\\[2mm]
			\cellcolor{gray!20} \textbf{$l_\infty$-norm}& &DF &   DAA-50 &   PGD-100 &   FAB-10 & FAB-100\\
			\hline
			avg. rob. acc. &&58.81&60.67&46.07&46.18&\textbf{45.47}\\
			\# best &&0&8&12&13*&\textbf{17}\\	
			avg. diff. to best & &14.58&16.45&1.85&1.96&\textbf{1.25}\\	
			max diff. to best & &78.10&49.00&\textbf{10.70}&20.30&17.10\\
			\hline
			\hline
			%\multicolumn{7}{c}{}\\
			\cellcolor{gray!20}\textbf{$l_2$-norm}& CW&DF &    LRA&   PGD-100 &   FAB-10 & FAB-100\\
			\hline
			avg. rob. acc. &45.09&56.10&36.97&44.94&36.41&\textbf{35.57}\\\# best &4&1&9&11&19*&\textbf{23}\\avg. diff. to best &9.65&20.67&1.54&9.51&0.98&\textbf{0.13}\\max diff. to best &65.40&91.40&13.60&64.80&8.40&\textbf{1.60}\\
			\hline\hline
			%\multicolumn{7}{c}{}\\
			%\cellcolor{yellow!40}
			\cellcolor{red!20}$l_2$-norm wo/ $l_\infty$-AT model on MNIST &CW&DF &    LRA&   PGD-100 &   FAB-10 & FAB-100\\
			\hline
			avg. rob. acc. &40.85&49.18&40.25&42.95&39.98&\textbf{39.57}\\\# best &4&1&8&11&14*&\textbf{18}\\avg. diff. to best &1.44&9.78&0.84&3.54&0.57&\textbf{0.16}\\max diff. to best &8.80&44.00&4.00&22.00&4.20&\textbf{1.60}\\
			\hline
			\hline
			%\multicolumn{7}{c}{}\\
			\cellcolor{gray!20}\textbf{$l_1$-norm}& &SF &    EAD&   PGD-100 &   FAB-10 & FAB-100\\
			\hline
			avg. rob. acc. &&64.47&35.79&49.51&33.26&\textbf{29.46}\\\# best &&0&13&0&10*&\textbf{17}\\avg. diff. to best &&35.31&6.63&20.35&4.10&\textbf{0.30}\\max diff. to best &&95.90&58.40&74.00&21.80&\textbf{1.60}\\
			\hline
			\hline
\cellcolor{red!20}$l_1$-norm wo/ $l_\infty$-AT model on MNIST& &SF &    EAD&   PGD-100 &   FAB-10 & FAB-100\\
\hline
avg. rob. acc. &&58.06&32.56&43.82&33.79&\textbf{32.06}\\\# best &&0&\textbf{13}&0&5*&12\\avg. diff. to best &&26.36&0.87&12.12&2.10&\textbf{0.36}\\max diff. to best &&53.10&4.80&31.90&3.90&\textbf{1.60}\\
\end{tabular}
}
\end{table*}

\subsection{Further results}\label{sec:app_results}
In Table \ref{tab:avg_dist} we report the average $l_p$-norm of the adversarial perturbations found by the different attacks, computed on the originally correctly classified points on which the attack is successful. Note that we cannot show this statistic for the attacks which do not minimize the distance of the adversarial example to the clean input (PGD and DAA). FAB-attack produces also in this metric the best results in most of the cases, being the best for every model when considering $l_\infty$ and $l_2$, and the best in 4 out of 6 cases in $l_1$ (lower values mean a stronger attack).
\begin{table*}[h]
	\centering
	\caption{We report mean $l_p$-norm of the adversarial perturbations found by the attacks (when successful, excluding the already misclassified points) for every model.
	\label{tab:avg_dist}}
	%
	%\vspace{2mm}
	\extrarowheight=0.5mm
	{%\footnotesize
		\small
		%\scriptsize
		%\parbox{.45\linewidth}{
		%\setlength\extrarowheight{1mm}
		%\parbox[c][][c]{100mm}{\textbf{Robust accuracy by FAB-10 with intermediate search}}
		\begin{tabular}{L{15mm} L{12mm}| *{4}{C{12mm}}}
			\multicolumn{6}{c}{\textbf{average norm of adversarial perturbations}}\\[2mm]
			
			\cellcolor{gray!20} \textbf{$l_\infty$-norm}& &&&DF     &   FAB \\
			\hline
			\multirow{3}{*}{MNIST}&plain&&&0.078&\textbf{0.066}\\&$l_\infty$-at&&&0.508&\textbf{0.326}\\&$l_2$-at&&&0.249&\textbf{0.170}\\
			\hline
			\multirow{3}{*}{CIFAR-10}&plain&&&0.008&\textbf{0.006}\\&$l_\infty$-at&&&0.032&\textbf{0.024}\\&$l_2$-at&&&0.026&\textbf{0.019}\\\hline \multicolumn{6}{c}{}\\
			\cellcolor{gray!20} \textbf{$l_2$-norm}& &DF&CW&LRA     &   FAB \\
			\hline
			\multirow{3}{*}{MNIST}&plain&1.13&1.01&\textbf{1.00}&\textbf{1.00}\\&$l_\infty$-at&4.95&1.76&1.25&\textbf{1.12}\\&$l_2$-at&3.10&2.35&2.25&\textbf{2.24}\\
			\hline
			\multirow{3}{*}{CIFAR-10}&plain&0.28&\textbf{0.21}&0.22&\textbf{0.21}\\&$l_\infty$-at&0.96&0.74&0.74&\textbf{0.73}\\&$l_2$-at&0.91&0.71&0.72&\textbf{0.70}\\\hline \multicolumn{6}{c}{}\\
			\cellcolor{gray!20} \textbf{$l_1$-norm}& &&& EAD & FAB\\
			\hline
			\multirow{3}{*}{MNIST}&plain&&&6.38&\textbf{6.04}\\&$l_\infty$-at&&&8.26&\textbf{3.36}\\&$l_2$-at&&&12.18&\textbf{12.16}\\
			\hline
			\multirow{3}{*}{CIFAR-10}&plain&&&3.01&\textbf{2.87}\\&$l_\infty$-at&&&\textbf{5.79}&6.03\\&$l_2$-at&&&\textbf{7.94}&8.05\\\hline
			
		\end{tabular}
		}
	\end{table*}

\begin{table*}
	\centering
	\caption{Comparison of $l_\infty$-, $l_2$- and $l_1$-attacks on a naturally trained model on MNIST. We report the accuracy in percentage of the classifier on the test set if the attack is allowed to perturb the test points of $\epsilon$ in $l_p$-distance. The statistics are computed on the first 1000 points on the test set for $l_\infty$ and $l_1$, on 500 points for $l_2$.}
	\label{tab:MNIST_plain}
	\vspace{2mm}
	{\footnotesize
		%\small
		%\scriptsize
		%\parbox{.45\linewidth}{
		\begin{tabular}{C{8mm} C{8mm}| *{9}{R{8mm}}}
			\multicolumn{11}{c}{\textbf{Robust accuracy of MNIST plain model}}\\[2mm]
			metric&$\epsilon$ & DF & DAA-1 & DAA-50 & PGD-1 & PGD-10 & PGD-100 & FAB-1 & FAB-10 & FAB-100\\
			\hline
			\multirow{5}{*}{$l_\infty$} &0.03&93.2& \textbf{91.9}& \textbf{91.9}&92.0& \textbf{91.9}& \textbf{91.9}&92.0&92.0&92.0\\&0.05&83.4&78.2&76.7&76.0&74.9& \textbf{74.6}&77.2&76.8&76.1\\&0.07&61.5&59.8&56.3&43.8&41.8& \textbf{40.4}&44.3&43.1&42.6\\&0.09&33.2&46.7&41.0&16.5&14.2& \textbf{12.8}&16.2&14.8&14.4\\&0.11&13.1&34.4&26.2&4.0&2.8& \textbf{2.4}&3.3&3.1& \textbf{2.4}\\
			\hline
			\multicolumn{11}{c}{}\\
			& & 
			CW & DF & LRA & PGD-1 & PGD-10 & PGD-100 & FAB-1 & FAB-10 & FAB-100\\
			\hline
			\multirow{5}{*}{$l_2$}& 0.5& \textbf{92.6}&93.6& \textbf{92.6}& \textbf{92.6}& \textbf{92.6}& \textbf{92.6}& \textbf{92.6}& \textbf{92.6}& \textbf{92.6}\\&1&47.4&58.6&47.4&48.4&47.4& \textbf{46.2}&47.0&46.8& \textbf{46.2}\\&1.5&8.8&19.8&7.8&9.8&8.8&8.2&7.8&7.2& \textbf{7.0}\\&2&0.6&1.8& \textbf{0.2}&1.2&0.6&0.6& \textbf{0.2}& \textbf{0.2}& \textbf{0.2}\\&2.5& \textbf{0.0}& \textbf{0.0}& \textbf{0.0}&0.6&0.2&0.2& \textbf{0.0}& \textbf{0.0}& \textbf{0.0}\\
			\hline
			\multicolumn{11}{c}{}\\
			
			& & \multicolumn{2}{r}{SparseFool} & EAD & PGD-1 & PGD-10 & PGD-100 & FAB-1 & FAB-10 & FAB-100\\
			\hline
			\multirow{5}{*}{$l_1$} &2&&95.5&93.6&94.4&93.9&93.7&94.2&93.7& \textbf{93.5}\\&4&&88.9&76.7&79.8&77.5&76.9&80.2&76.6& \textbf{75.2}\\&6&&75.8&48.1&57.4&52.2&49.3&54.5&47.2& \textbf{43.3}\\&8&&60.3&26.6&46.7&36.3&31.6&31.3&25.3& \textbf{22.4}\\&10&&43.8&11.2&40.0&27.4&22.1&15.2&9.8& \textbf{8.4}\\

			\hline
		\end{tabular}
	}
	\end{table*}
\begin{table*}
	\centering
	\caption{Comparison of $l_\infty$-, $l_2$- and $l_1$-attacks on an $l_\infty$-robust model on MNIST. We report the accuracy on the test set if the attack is allowed to perturb the test points of $\epsilon$ in $l_p$-distance. The statistics are computed on the first 1000 points on the test set for $l_\infty$ and $l_1$, on 500 points for $l_2$.}
	\label{tab:MNIST_linf}
	\vspace{2mm}
	{\footnotesize
		%\small
		%\scriptsize
		%\parbox{.45\linewidth}{
		\begin{tabular}{C{8mm} C{8mm}| *{9}{R{8mm}}}
			\multicolumn{11}{c}{\textbf{Robust accuracy of MNIST $l_\infty$-robust model}}\\[2mm]
			metric&$\epsilon$ & DF & DAA-1 & DAA-50 & PGD-1 & PGD-10 & PGD-100 & FAB-1 & FAB-10 & FAB-100\\
			\hline
			\multirow{5}{*}{$l_\infty$} &0.2&95.2&94.6& \textbf{93.7}&95.0&94.2& \textbf{93.7}&94.6&94.4&93.9\\&0.25&94.7&92.7& \textbf{91.1}&93.1&91.8&91.4&93.3&92.1&91.7\\&0.3&93.9&89.5& \textbf{87.2}&91.3&88.3&87.6&91.2&89.2&88.5\\&0.325&92.5&72.1& \textbf{64.2}&74.9&68.4&64.7&86.2&83.1&81.3\\&0.35&89.8&19.7& \textbf{11.7}&32.1&19.3&13.8&48.7&32.0&23.8\\
			\hline
			\multicolumn{11}{c}{}\\
			& &
			CW & DF & LRA & PGD-1 & PGD-10 & PGD-100 & FAB-1 & FAB-10 & FAB-100\\
			\hline
			\multirow{5}{*}{$l_2$}&1&88.8&94.6&73.6&92.2&90.8&89.8&84.2&70.6& \textbf{65.4}\\&1.5&77.6&93.0&25.8&86.0&81.2&77.0&47.0&20.6& \textbf{12.2}\\&2&64.4&91.6&3.2&77.8&67.0&57.8&15.6&1.8& \textbf{0.2}\\&2.5&53.8&89.6&0.4&68.2&49.6&36.4&3.8& \textbf{0.0}& \textbf{0.0}\\&3&46.8&84.6& \textbf{0.0}&59.8&29.6&13.4&1.4& \textbf{0.0}& \textbf{0.0}\\
			\hline
			\multicolumn{11}{c}{}\\
			
			& & \multicolumn{2}{r}{SparseFool} & EAD & PGD-1 & PGD-10 & PGD-100 & FAB-1 & FAB-10 & FAB-100\\
			\hline
			\multirow{5}{*}{$l_1$} &2.5&&96.8&92.2&94.1&93.7&93.6&88.0&74.3& \textbf{56.9}\\&5&&96.5&76.0&90.9&88.9&88.2&78.3&39.4& \textbf{17.6}\\&7.5&&96.4&49.5&85.2&81.4&79.0&66.1&19.8& \textbf{5.0}\\&10&&96.4&27.4&80.2&73.5&70.3&49.3&11.9& \textbf{2.4}\\&12.5&&96.4&14.6&74.9&65.6&58.7&35.6&7.7& \textbf{0.5}\\

			\hline
		\end{tabular}
	}
\end{table*}

\begin{table*}
	\centering
	\caption{Comparison of $l_\infty$-, $l_2$- and $l_1$-attacks on an $l_2$-robust model on MNIST. We report the accuracy in percentage of the classifier on the test set if the attack is allowed to perturb the test points of $\epsilon$ in $l_p$-distance. The statistics are computed on the first 1000 points on the test set for $l_\infty$ and $l_1$, on 500 points for $l_2$.}
	\label{tab:MNIST_l2}
	\vspace{2mm}
	{\footnotesize
		%\small
		%\scriptsize
		%\parbox{.45\linewidth}{
		\begin{tabular}{C{8mm} C{8mm}| *{9}{R{8mm}}}
			\multicolumn{11}{c}{\textbf{Robust accuracy of MNIST $l_2$-robust model}}\\[2mm]
			metric&$\epsilon$ & DF & DAA-1 & DAA-50 & PGD-1 & PGD-10 & PGD-100 & FAB-1 & FAB-10 & FAB-100\\
			\hline
			\multirow{5}{*}{$l_\infty$} & 0.05&96.7&96.4& \textbf{96.3}&96.4& \textbf{96.3}& \textbf{96.3}&96.4& \textbf{96.3}& \textbf{96.3}\\&0.1&93.4&91.0& \textbf{90.2}&90.7&90.4& \textbf{90.2}&90.8&90.4&90.4\\&0.15&86.4&74.3&72.3&74.6&73.2&72.4&74.0&72.3& \textbf{72.0}\\&0.2&73.8&34.5&27.2&36.2&29.8&26.5&34.1&28.2& \textbf{24.4}\\&0.25&55.1&1.5&0.9&2.6&1.5&1.0&1.9&0.9& \textbf{0.8}\\
			\hline
			\multicolumn{11}{c}{}\\
			& & 
			CW & DF & LRA & PGD-1 & PGD-10 & PGD-100 & FAB-1 & FAB-10 & FAB-100\\
			\hline
			\multirow{5}{*}{$l_2$}& 1& \textbf{92.6}&93.8& \textbf{92.6}&93.0&93.0&93.0& \textbf{92.6}& \textbf{92.6}& \textbf{92.6}\\&1.5&84.8&87.2& \textbf{83.4}&83.8& \textbf{83.4}& \textbf{83.4}&83.8&83.6&83.6\\&2&70.6&79.0&68.0&68.8&68.0& \textbf{67.6}&69.8&69.0&67.8\\&2.5&46.4&67.4&41.6&45.6&40.4& \textbf{37.6}&45.6&41.8&39.2\\&3&17.2&54.2&11.2&17.4&12.4& \textbf{10.2}&18.6&13.4&11.0\\
			\hline
			\multicolumn{11}{c}{}\\
			
			& & \multicolumn{2}{r}{SparseFool} & EAD & PGD-1 & PGD-10 & PGD-100 & FAB-1 & FAB-10 & FAB-100\\
			\hline
			\multirow{5}{*}{$l_1$} &5&&94.9& \textbf{89.8}&90.3&90.2&90.2&90.5&90.2&90.0\\&8.75&&89.1& \textbf{71.2}&75.5&74.0&72.7&75.3&73.7&72.2\\&12.5&&81.0&45.9&61.1&57.5&54.9&55.6&49.2& \textbf{45.7}\\&16.25&&72.8& \textbf{20.6}&49.2&42.3&38.4&32.2&24.1&20.8\\&20&&60.8&8.3&41.4&29.6&23.2&15.2&9.4& \textbf{7.7}\\

			\hline
		\end{tabular}
	}
	
\end{table*}

\begin{table*}
	\centering
	\caption{Comparison of $l_\infty$-, $l_2$- and $l_1$-attacks on a naturally trained model on CIFAR-10. We report the accuracy in percentage of the classifier on the test set if the attack is allowed to perturb the test points of $\epsilon$ in $l_p$-distance. The statistics are computed on the first 1000 points on the test set for $l_\infty$ and $l_1$, on 500 points for $l_2$.}
	\label{tab:CIFAR-10_plain}
	\vspace{2mm}
	{\footnotesize
		\begin{tabular}{C{8mm} C{8mm}| *{9}{R{8mm}}}
			\multicolumn{11}{c}{\textbf{Robust accuracy of CIFAR-10 plain model}}\\[2mm]
			metric&$\epsilon$ & DF & DAA-1 & DAA-50 & PGD-1 & PGD-10 & PGD-100 & FAB-1 & FAB-10 & FAB-100\\
			\hline
			\multirow{5}{*}{$l_\infty$} & $\nicefrac{1}{255}$ &62.6&65.7&64.1&56.1&55.8& \textbf{55.6}&56.5&55.9&55.7 \\&$\nicefrac{1.5}{255}$&49.3&63.2&60.8&38.9&37.9& \textbf{37.4}&38.5&37.7& \textbf{37.4}\\&$\nicefrac{2}{255}$&37.3&62.4&58.5&24.3&23.3&22.9&23.4&21.9& \textbf{21.2}\\&$\nicefrac{2.5}{255}$&26.4&61.2&56.3&16.2&14.8&14.0&13.2&12.0& \textbf{11.8}\\&$\nicefrac{3}{255}$&19.0&60.2&54.4&10.7&9.2&8.6&7.4&5.8& \textbf{5.4}\\

			\hline
			\multicolumn{11}{c}{}\\
			& & 
			CW & DF & LRA & PGD-1 & PGD-10 & PGD-100 & FAB-1 & FAB-10 & FAB-100\\
			\hline
			\multirow{5}{*}{$l_2$}& 0.1&69.4&72.2&69.0&68.4& \textbf{67.6}& \textbf{67.6}&68.4&68.4&68.4\\&0.15&55.4&62.6&55.0&54.6& \textbf{53.8}& \textbf{53.8}&54.6&54.0& \textbf{53.8}\\&0.2&43.4&51.2&43.4&43.8&42.8&42.0&42.4&42.0& \textbf{41.8}\\&0.3&21.6&33.8&22.0&24.8&24.2&23.6&21.6&20.8& \textbf{20.6}\\&0.4&9.4&20.8&9.8&18.2&16.2&15.4&9.6&8.2& \textbf{8.0}\\
			
			\hline
			\multicolumn{11}{c}{}\\
			
			& & \multicolumn{2}{r}{SparseFool} & EAD & PGD-1 & PGD-10 & PGD-100 & FAB-1 & FAB-10 & FAB-100\\
			\hline
			\multirow{5}{*}{$l_1$} & 2&&72.1&54.7&54.9&54.4&53.9&55.5&52.2& \textbf{50.8}\\&4&&58.6&24.1&30.0&29.1&28.9&30.7&25.1& \textbf{22.4}\\&6&&45.6&8.9&18.8&18.6&18.4&17.0&10.5& \textbf{8.1}\\&8&&34.3&3.0&14.2&14.1&14.0&7.8&3.8& \textbf{2.5}\\&10&&27.2& \textbf{0.7}&12.9&12.5&12.3&4.7&1.5&1.0\\

			\hline
		\end{tabular}
	}
\end{table*}

\begin{table*}
	\centering
	\caption{Comparison of $l_\infty$-, $l_2$- and $l_1$-attacks on an $l_\infty$-robust model on CIFAR-10. We report the accuracy in percentage of the classifier on the test set if the attack is allowed to perturb the test points of $\epsilon$ in $l_p$-distance. The statistics are computed on the first 1000 points on the test set for $l_\infty$ and $l_1$, on 500 points for $l_2$.}
	\label{tab:CIFAR-10_linf}
	\vspace{2mm}
	{\footnotesize
		\begin{tabular}{C{8mm} C{8mm}| *{9}{R{8mm}}}
			\multicolumn{11}{c}{\textbf{Robust accuracy of CIFAR-10 $l_\infty$-robust model}}\\[2mm]
			metric&$\epsilon$ & DF & DAA-1 & DAA-50 & PGD-1 & PGD-10 & PGD-100   & FAB-1 & FAB-10 & FAB-100\\
			\hline
			\multirow{5}{*}{$l_\infty$} & $\nicefrac{2}{255}$ &66.8&66.9&66.3& \textbf{65.5}& \textbf{65.5}& \textbf{65.5}&65.8&65.8&65.7\\&$\nicefrac{4}{255}$&53.2&63.8&61.4&49.8&49.3&49.0&49.2&49.1& \textbf{48.9}\\&$\nicefrac{6}{255}$&42.9&63.1&58.4&38.0&36.9&36.6&35.4&34.7& \textbf{34.6}\\&$\nicefrac{8}{255}$&32.9&61.2&56.3&30.5&30.0&29.6&23.8&23.5& \textbf{23.3}\\&$\nicefrac{10}{255}$ & 24.5&59.8&54.1&25.8&23.7&22.4&15.4&14.7& \textbf{14.4}\\

			\hline
			\multicolumn{11}{c}{}\\
			& & 
			CW & DF & LRA & PGD-1 & PGD-10 & PGD-100 & FAB-1 & FAB-10 & FAB-100\\
			\hline
			\multirow{5}{*}{$l_2$}& 0.25&64.6&67.0& \textbf{64.4}& \textbf{64.4}& \textbf{64.4}& \textbf{64.4}&64.8&64.6& \textbf{64.4}\\&0.5&48.4&53.0&48.8&49.0&48.4& \textbf{48.0}&48.4&48.4&48.2\\&0.75&33.4&41.4&33.4&39.0&38.2&37.4&33.6&33.2& \textbf{33.0}\\&1&22.8&32.6&22.8&35.0&34.4&33.8&22.2&21.6& \textbf{21.4}\\&1.25&12.0&24.2&13.0&34.6&34.2&33.2&12.2& \textbf{11.2}& \textbf{11.2}\\

			\hline
			\multicolumn{11}{c}{}\\
			
			& & \multicolumn{2}{r}{SparseFool} & EAD & PGD-1 & PGD-10 & PGD-100 & FAB-1 & FAB-10 & FAB-100\\
			\hline
			\multirow{5}{*}{$l_1$} &5&&57.8& \textbf{36.8}&47.3&46.6&46.2&43.1&39.9&37.9\\&8.75&&44.7& \textbf{19.2}&37.4&37.0&36.8&25.7&22.5&20.2\\&12.5&&34.9& \textbf{7.1}&34.0&33.9&33.9&13.7&10.9&8.7\\&16.25&&27.6& \textbf{3.0}&33.3&33.2&33.1&7.1&4.3&3.5\\&20&&20.2& \textbf{0.9}&32.9&32.8&32.8&3.8&1.7&1.3\\

			\hline
		\end{tabular}
	}
\end{table*}
\begin{table*}
	\centering
	\caption{Comparison of $l_\infty$-, $l_2$- and $l_1$-attacks on an $l_2$-robust model on CIFAR-10. We report the accuracy in percentage of the classifier on the test set if the attack is allowed to perturb the test points of $\epsilon$ in $l_p$-distance. The statistics are computed on the first 1000 points on the test set for $l_\infty$ and $l_1$, on 500 points for $l_2$.}
	\label{tab:CIFAR-10_l2}
	\vspace{2mm}
	{\footnotesize
		\begin{tabular}{C{8mm} C{8mm}| *{9}{R{8mm}}}
			\multicolumn{11}{c}{\textbf{Robust accuracy of CIFAR-10 $l_2$-robust model}}\\[2mm]
			metric&$\epsilon$ & DF & DAA-1 & DAA-50 & PGD-1 & PGD-10 & PGD-100 & FAB-1 & FAB-10 & FAB-100\\
			\hline
			\multirow{5}{*}{$l_\infty$} & $\nicefrac{2}{255}$ & 64.1&67.2&66.3&62.6&62.5& \textbf{62.4}&62.7&62.6&62.6\\
			&$\nicefrac{4}{255}$&49.0&65.0&62.8&45.3&45.0&44.9&44.4& \textbf{44.2}& \textbf{44.2}\\ &$\nicefrac{6}{255}$&36.9&64.2&60.8&32.9&31.6&31.1&27.2&26.8& \textbf{26.7}\\
			&$\nicefrac{8}{255}$&25.8&62.3&58.0&25.7&24.9&23.9&14.8&14.1& \textbf{13.8}\\ &$\nicefrac{10}{255}$&  17.6&61.9&54.8&21.9&19.8&18.6&8.6&8.0& \textbf{7.9}\\
			\hline

			\multicolumn{11}{c}{}\\
			& & 
			CW & DF & LRA & PGD-1 & PGD-10 & PGD-100 & FAB-1 & FAB-10 & FAB-100\\
			\hline
			\multirow{5}{*}{$l_2$}&0.25&66.0&67.0& \textbf{65.6}&65.8& \textbf{65.6}& \textbf{65.6}& \textbf{65.6}& \textbf{65.6}& \textbf{65.6}\\&0.5&48.2&53.8& \textbf{47.8}&49.6&48.8&48.8&48.4&48.2&48.0\\&0.75&32.6&42.2&32.4&38.4&37.2&36.4&32.8&32.4& \textbf{32.2}\\&1&21.6&30.0&21.6&35.4&33.6&33.0&21.8&21.4& \textbf{21.0}\\&1.25& \textbf{11.4}&22.4&12.4&34.2&31.6&31.2&12.2&12.2& \textbf{11.4}\\

			\hline
			\multicolumn{11}{c}{}\\
			
			& & \multicolumn{2}{r}{SparseFool} & EAD & PGD-1 & PGD-10 & PGD-100 & FAB-1 & FAB-10 & FAB-100\\
			\hline
			\multirow{5}{*}{$l_1$} &3&&69.5& \textbf{62.2}&64.5&64.4&64.4&63.4&63.2&63.0\\&6&&61.6& \textbf{45.5}&51.9&51.9&51.8&48.6&47.2&45.6\\&9&&53.1& \textbf{27.7}&42.8&42.5&42.5&33.9&30.6&28.8\\&12&&44.4&17.9&38.5&38.3&38.0&23.8&19.8& \textbf{17.3}\\&15&&37.0& \textbf{10.4}&35.8&35.8&35.4&16.0&12.4&11.2\\

			\hline
		\end{tabular}
	}
	
	\end{table*}

\begin{table*}
	\centering
	\caption{Comparison of $l_\infty$-, $l_2$- and $l_1$-attacks on a naturally trained model on Restricted ImageNet. We report the accuracy in percentage of the classifier on the test set if the attack is allowed to perturb the test points of $\epsilon$ in $l_p$-distance. The statistics are computed on the first 500 points of the test set.}
	\label{tab:ImageNet_plain}
	\vspace{2mm}
	{\footnotesize
		\begin{tabular}{C{8mm} C{8mm}| *{7}{R{8mm}}}
			\multicolumn{9}{c}{\textbf{Robust accuracy of Restricted ImageNet plain model}}\\[2mm]
			metric&$\epsilon$ & DF & DAA-1 & DAA-10 & PGD-1 & PGD-10 & FAB-1 & FAB-10\\
			\hline
			\multirow{5}{*}{$l_\infty$} & $\nicefrac{0.25}{255}$ &76.6&74.8&74.8&74.8& \textbf{74.6}&75.2&75.2\\&$\nicefrac{0.5}{255}$ &52.0&51.8&48.2&38.2& \textbf{37.8}&39.6&39.6\\&$\nicefrac{0.75}{255}$&26.8&46.0&41.0& \textbf{12.2}& \textbf{12.2}&14.2&14.2\\&$\nicefrac{1}{255}$&11.2&43.2&39.4&3.8&3.8& \textbf{3.6}& \textbf{3.6}\\&$\nicefrac{1.25}{255}$&5.0&41.2&38.2& \textbf{1.0}& \textbf{1.0}&1.2& \textbf{1.0}\\

			\hline
			\multicolumn{9}{c}{}\\
			& & &
			&DF & PGD-1 & PGD-10 & FAB-1 & FAB-10\\
			\hline
			\multirow{5}{*}{$l_2$} & 0.2&&&80.2& \textbf{76.0}& \textbf{76.0}&77.0&76.8\\&0.4&&&58.4&40.8& \textbf{40.6}&43.0&42.2\\&0.6&&&33.8&15.4& \textbf{14.8}&19.0&18.2\\&0.8&&&18.8& \textbf{4.0}& \textbf{4.0}&4.6&4.4\\&1&&&8.6&1.6&1.6&1.2& \textbf{1.0}\\

			\hline
			\multicolumn{9}{c}{}\\
			
			& & &\multicolumn{2}{r}{SparseFool} & PGD-1 & PGD-5& FAB-1 & FAB-5\\
			\hline
			\multirow{5}{*}{$l_1$} &5&&&88.6&81.8&81.8&79.6& \textbf{78.0}\\&16&&&80.0&45.2&45.2&46.8& \textbf{40.0}\\&27&&&70.6&17.8& \textbf{17.4}&25.6&19.8\\&38&&&65.0&6.2& \textbf{6.0}&13.6&7.0\\&49&&&55.4& \textbf{2.2}& \textbf{2.2}&6.8&3.8\\

			\hline
		\end{tabular}
	}
\end{table*}

\begin{table*}
	\centering
	\caption{Comparison of $l_\infty$-, $l_2$- and $l_1$-attacks on an $l_\infty$-robust model on Restricted ImageNet. We report the accuracy in percentage of the classifier on the test set if the attack is allowed to perturb the test points of $\epsilon$ in $l_p$-distance. The statistics are computed on the first 500 points of the test set.}
	\label{tab:ImageNet_linf}
	\vspace{2mm}
	{\footnotesize
		\begin{tabular}{C{8mm} C{8mm}| *{7}{R{8mm}}}
			\multicolumn{9}{c}{\textbf{Robust accuracy of Restricted ImageNet $l_\infty$-robust model}}\\[2mm]
			metric&$\epsilon$ & DF & DAA-1 & DAA-10 & PGD-1 & PGD-10 & FAB-1 & FAB-10\\
			\hline
			\multirow{5}{*}{$l_\infty$} & $\nicefrac{2}{255}$ &75.8&75.0&75.0& \textbf{74.6}& \textbf{74.6}&75.2&75.2\\&$\nicefrac{4}{255}$&53.0&46.2&46.2& \textbf{45.4}& \textbf{45.4}&47.4&47.4\\&$\nicefrac{6}{255}$&32.4&24.6&23.8& \textbf{19.4}& \textbf{19.4}&21.2&21.0\\&$\nicefrac{8}{255}$&19.4&17.0&14.6& \textbf{6.2}& \textbf{6.2}&6.8&6.8\\&$\nicefrac{10}{255}$&10.8&12.8&11.6&1.0& \textbf{0.8}&1.2&1.2\\

			\hline
			\multicolumn{9}{c}{}\\
			& & &
			&DF & PGD-1 & PGD-10 & FAB-1 & FAB-10\\
			\hline
			\multirow{5}{*}{$l_2$} &1&&&79.4& \textbf{76.6}& \textbf{76.6}&77.0&76.8\\&2&&&65.0&46.8& \textbf{46.2}&49.8&49.2\\&3&&&46.8&22.4& \textbf{21.4}&24.4&23.8\\&4&&&32.8&9.0& \textbf{8.6}&10.8&10.6\\&5&&&20.4&3.0& \textbf{2.8}&3.2&3.2\\

			\hline
			\multicolumn{9}{c}{}\\
			
			& & &\multicolumn{2}{r}{SparseFool} & PGD-1 & PGD-5& FAB-1 & FAB-5\\
			\hline
			\multirow{5}{*}{$l_1$} &15&&&81.8&69.8&69.8&69.6& \textbf{68.2}\\&25&&&76.4&58.6&58.2&56.2& \textbf{53.6}\\&40&&&71.4&41.0&41.0&41.2& \textbf{37.6}\\&60&&&63.2&28.8&28.8&28.2& \textbf{23.8}\\&100&&&49.2&12.0&11.6&14.6& \textbf{11.2}\\
			
			%15&&&81.8&72.4&72.4&69.4& \textbf{66.8}\\&25&&&76.4&61.6&61.2&55.0& \textbf{52.6}\\&40&&&71.4&47.2&46.6&41.6& \textbf{37.2}\\&60&&&63.2&33.0&32.6&29.4& \textbf{24.4}\\&100&&&49.2&14.0&13.6&15.8& \textbf{10.8}\\
			\hline
		\end{tabular}
	}
\end{table*}

\begin{table*}
	\centering
	\caption{Comparison of $l_\infty$-, $l_2$- and $l_1$-attacks on an $l_2$-robust model on Restricted ImageNet. We report the accuracy in percentage of the classifier on the test set if the attack is allowed to perturb the test points of $\epsilon$ in $l_p$-distance. The statistics are computed on the first 500 points of the test set.}
	\label{tab:ImageNet_l2}
	\vspace{2mm}
	{\footnotesize
		\begin{tabular}{C{8mm} C{8mm}| *{7}{R{8mm}}}
			\multicolumn{9}{c}{\textbf{Robust accuracy of Restricted ImageNet $l_2$-robust model}}\\[2mm]
			metric&$\epsilon$ & DF & DAA-1 & DAA-10 & PGD-1 & PGD-10 & FAB-1 & FAB-10\\
			\hline
			\multirow{5}{*}{$l_\infty$} & $\nicefrac{2}{255}$ &74.4& \textbf{73.0}& \textbf{73.0}& \textbf{73.0}& \textbf{73.0}&73.8&73.8\\&$\nicefrac{4}{255}$&49.0&39.6&39.2& \textbf{37.6}& \textbf{37.6}&39.6&39.4\\&$\nicefrac{6}{255}$&27.4&22.6&21.0& \textbf{13.2}& \textbf{13.2}&15.0&15.0\\&$\nicefrac{8}{255}$&13.8&18.6&16.8&3.6&3.6&3.6& \textbf{3.2}\\&$\nicefrac{10}{255}$&6.6&15.6&13.8& \textbf{0.4}& \textbf{0.4}&0.8&0.8\\

			\hline
			\multicolumn{9}{c}{}\\
			& & &
			&DF & PGD-1 & PGD-50 & FAB-1 & FAB-10\\
			\hline
			\multirow{5}{*}{$l_2$} &2&&&74.2& \textbf{71.8}& \textbf{71.8}&72.8&72.8\\&3&&&61.6&51.4& \textbf{51.0}&52.4&52.4\\&4&&&45.6&31.0& \textbf{30.8}&34.4&33.8\\&5&&&34.6& \textbf{20.4}& \textbf{20.4}&22.6&21.8\\&6&&&25.2& \textbf{9.6}& \textbf{9.6}&11.8&11.6\\

			\hline
			\multicolumn{9}{c}{}\\
			
			& & &\multicolumn{2}{r}{SparseFool} & PGD-1 & PGD-5& FAB-1 & FAB-5\\
			\hline
			\multirow{5}{*}{$l_1$} &50&&&85.4&81.0&81.0&78.6& \textbf{78.4}\\&100&&&79.6&63.8&63.6&60.8& \textbf{59.0}\\&150&&&74.4&48.4&48.4&45.2& \textbf{42.2}\\&200&&&68.6&32.2&32.2&31.0& \textbf{29.0}\\&250&&&60.0&23.0&22.4&22.8& \textbf{20.2}\\

			\hline
		\end{tabular}
	}
\end{table*}

\clearpage
%\hl{

\section{Analysis of the attacks}

\begin{figure*}[p]\centering \begin{tabular}{c}
		\textbf{MNIST - $l_\infty$-attack} \\[2mm]
		\includegraphics[width=0.5\columnwidth, clip, trim=10mm 0mm 15mm 0mm]{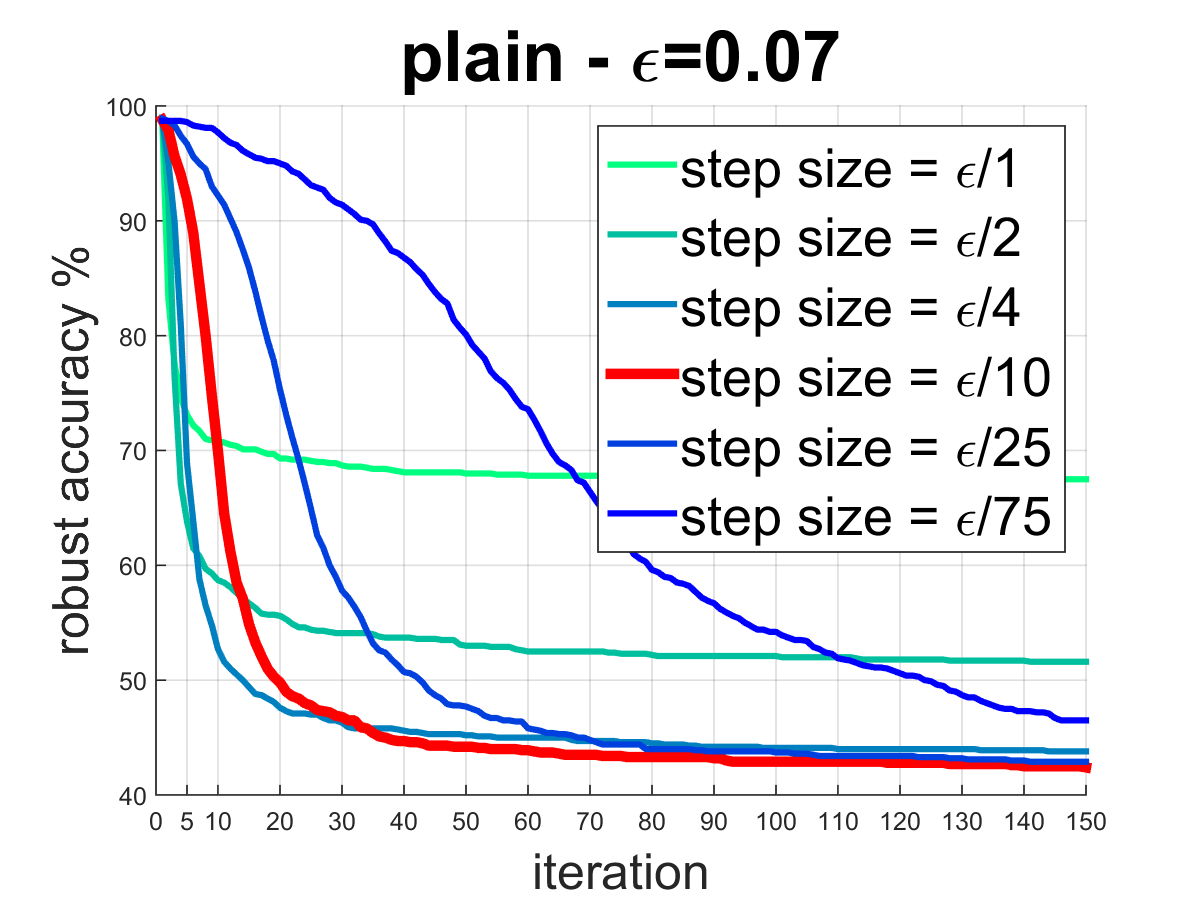} \includegraphics[width=0.5\columnwidth, clip, trim=10mm 0mm 15mm 0mm]{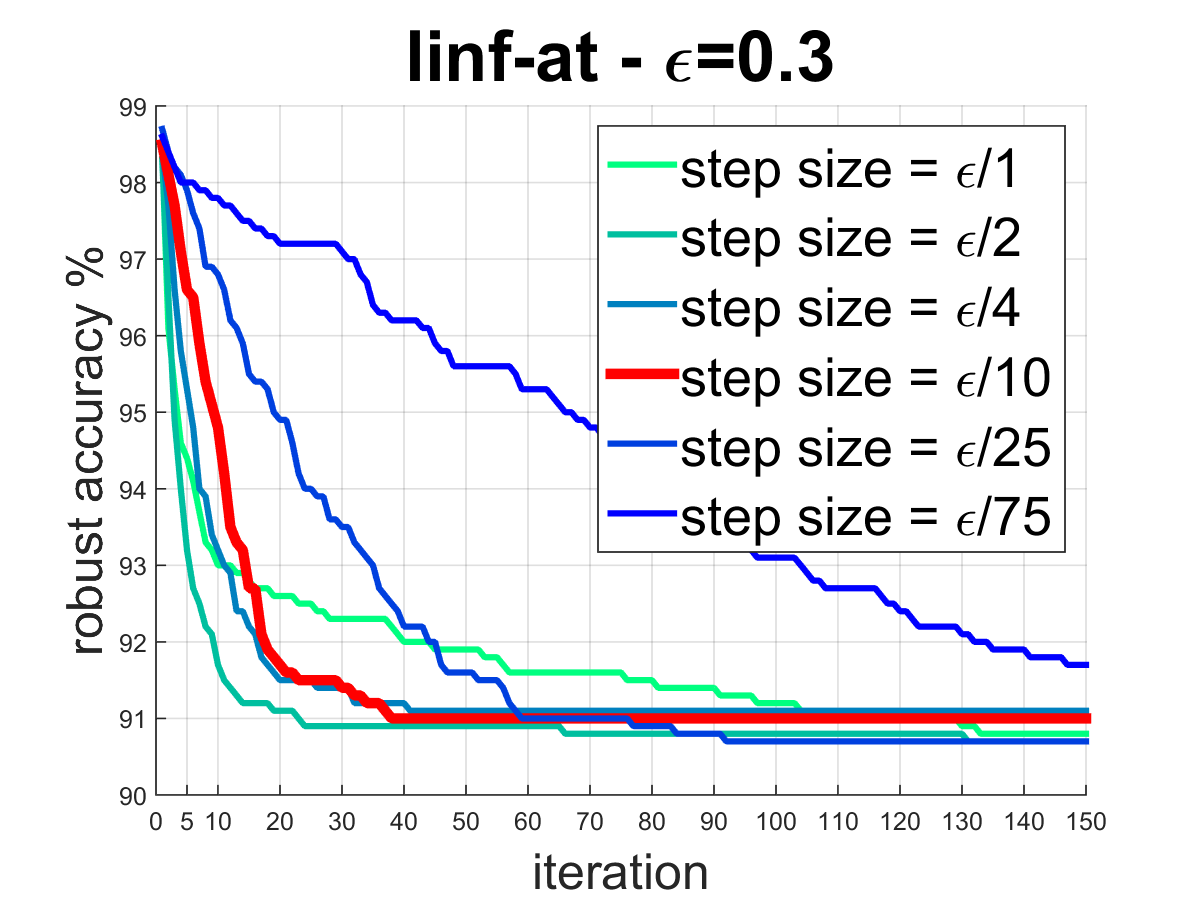} \includegraphics[width=0.5\columnwidth, clip, trim=10mm 0mm 15mm 0mm]{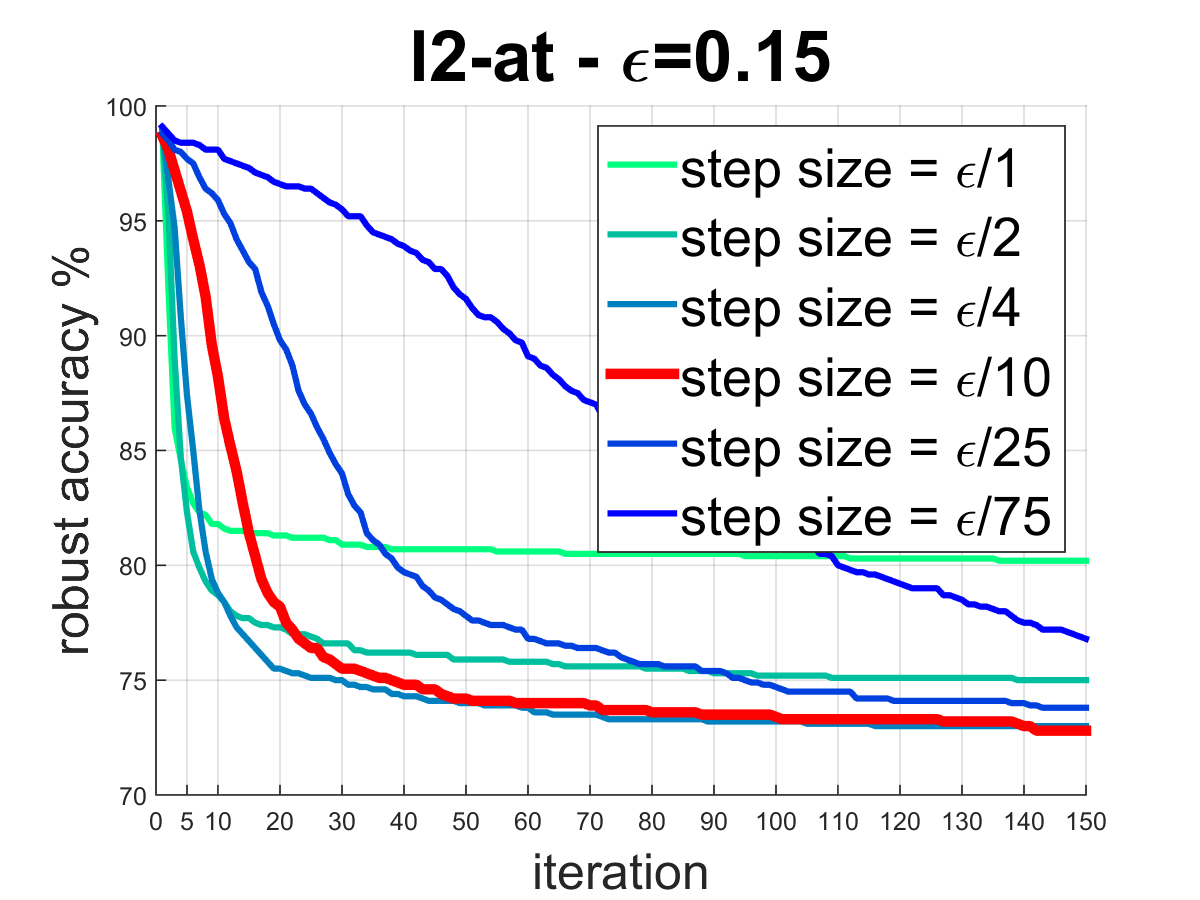}\\
		\includegraphics[width=0.5\columnwidth, clip, trim=10mm 0mm 15mm 0mm]{imgs_3/pl_stepsize_linf_mnist_plain} \includegraphics[width=0.5\columnwidth, clip, trim=10mm 0mm 15mm 0mm]{imgs_3/pl_stepsize_linf_mnist_linf-at} \includegraphics[width=0.5\columnwidth, clip, trim=10mm 0mm 15mm 0mm]{imgs_3/pl_stepsize_linf_mnist_l2-at}\\
		\hline
		\\
		\textbf{CIFAR-10 - $l_\infty$-attack}\\[2mm]
		\includegraphics[width=0.5\columnwidth, clip, trim=10mm 0mm 15mm 0mm]{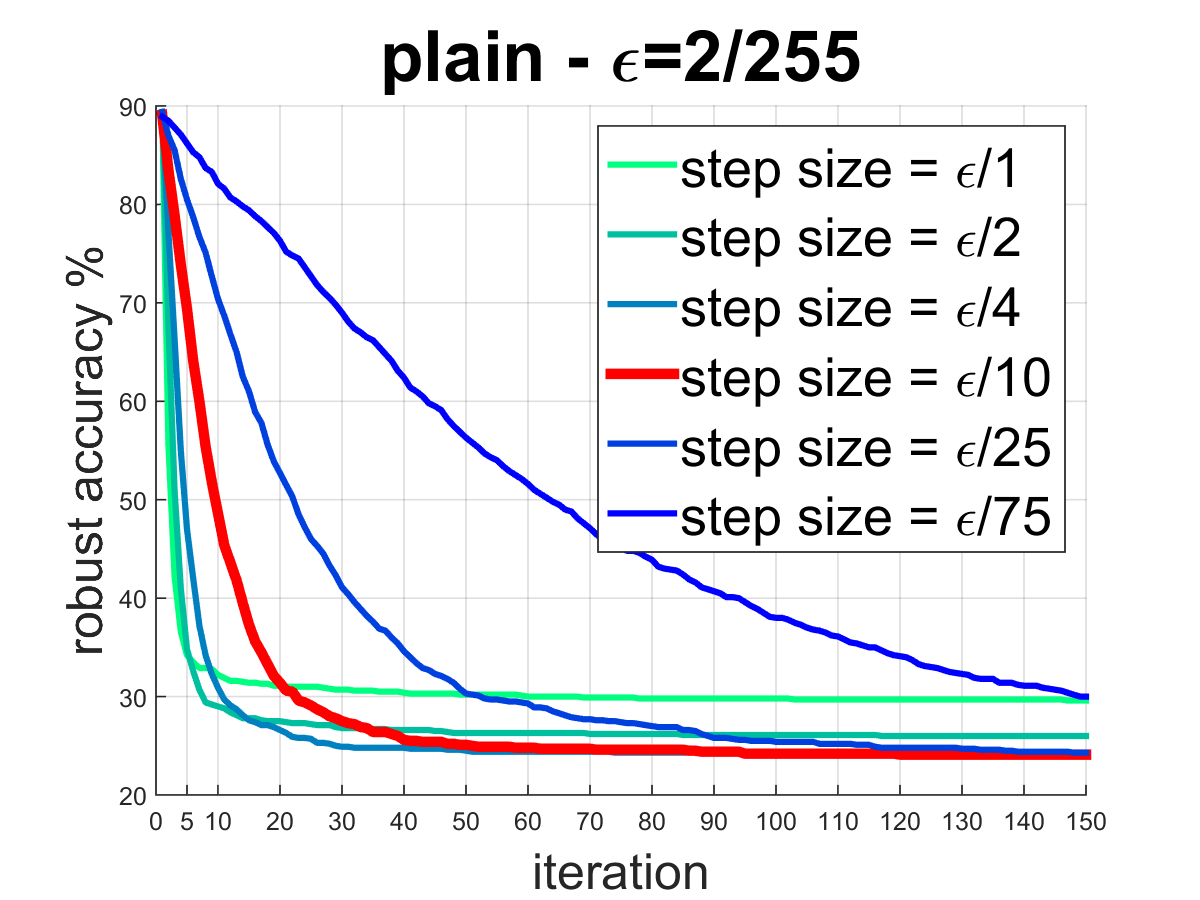} \includegraphics[width=0.5\columnwidth, clip, trim=10mm 0mm 15mm 0mm]{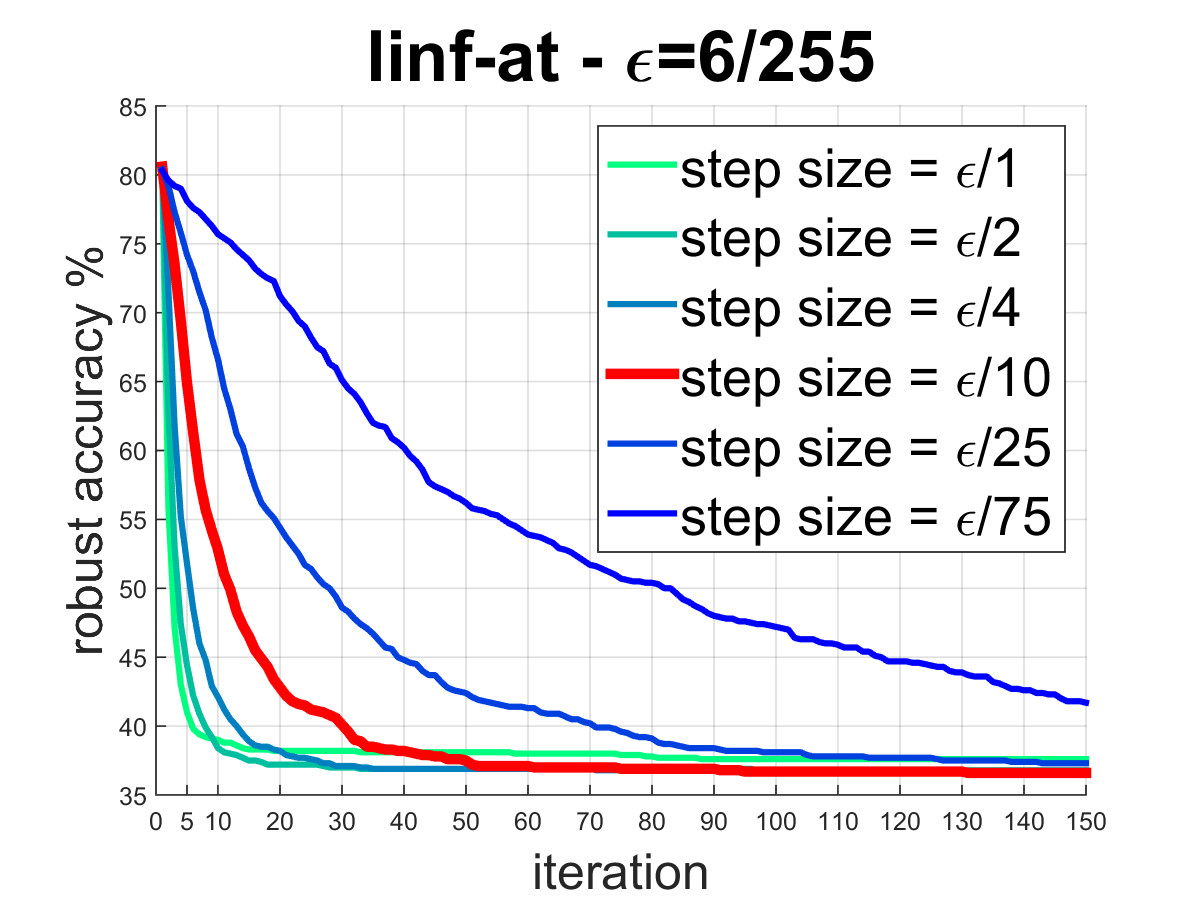} \includegraphics[width=0.5\columnwidth, clip, trim=10mm 0mm 15mm 0mm]{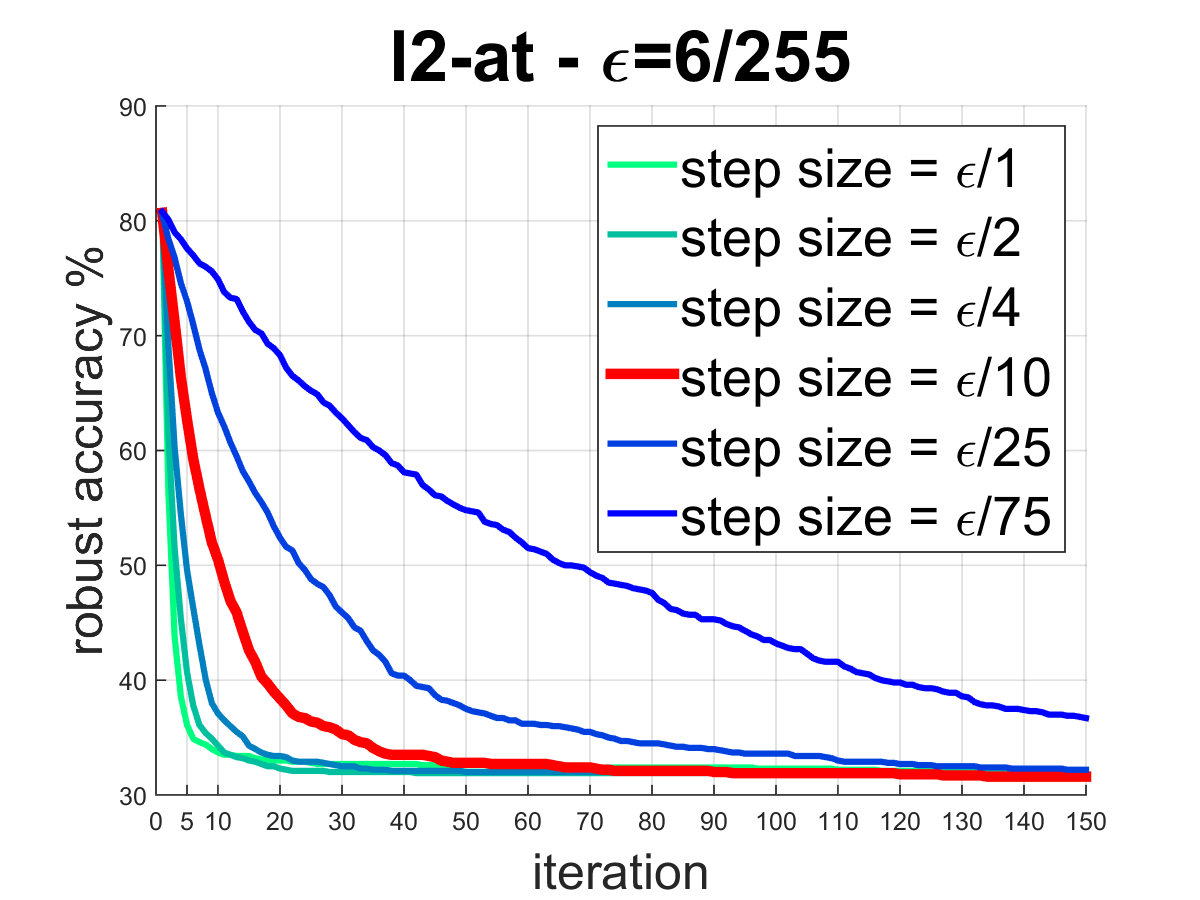}\\
		\includegraphics[width=0.5\columnwidth, clip, trim=10mm 0mm 15mm 0mm]{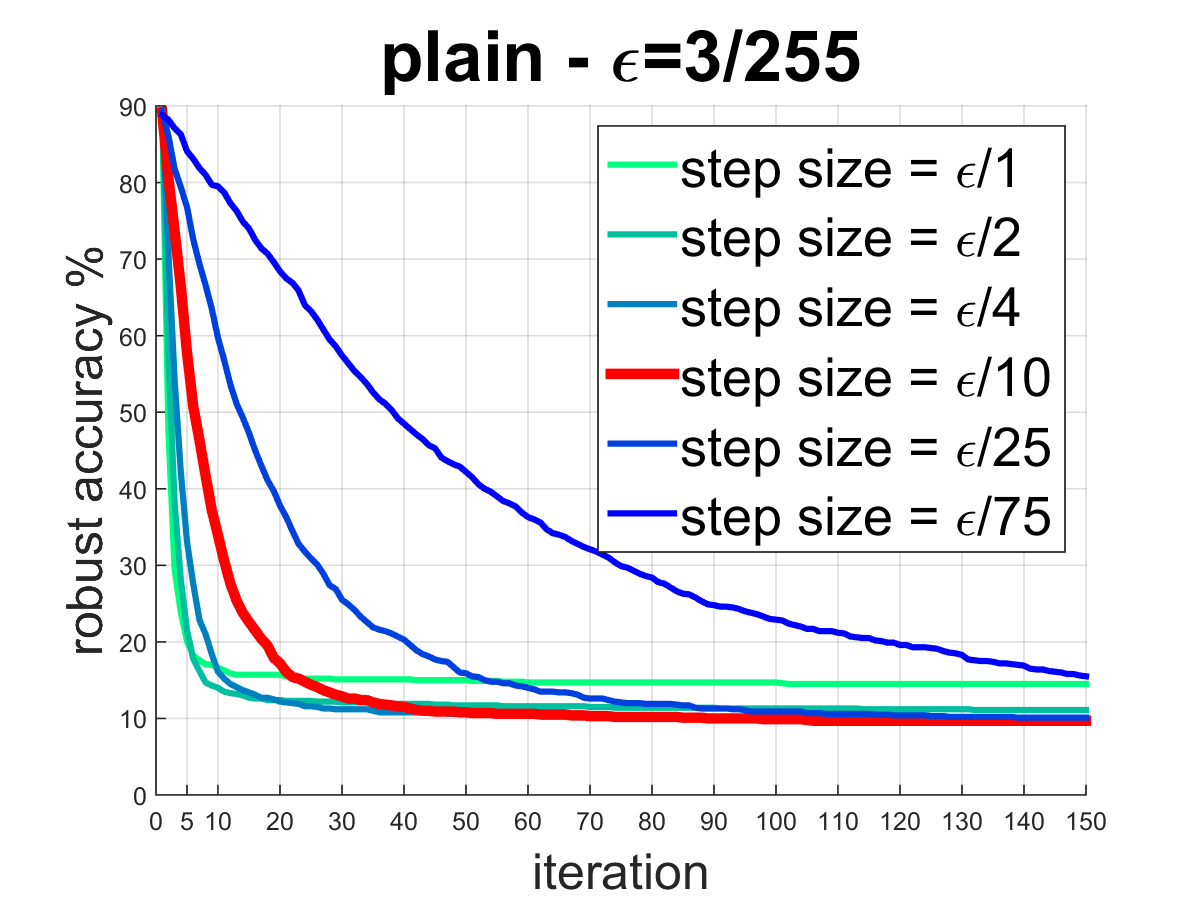} \includegraphics[width=0.5\columnwidth, clip, trim=10mm 0mm 15mm 0mm]{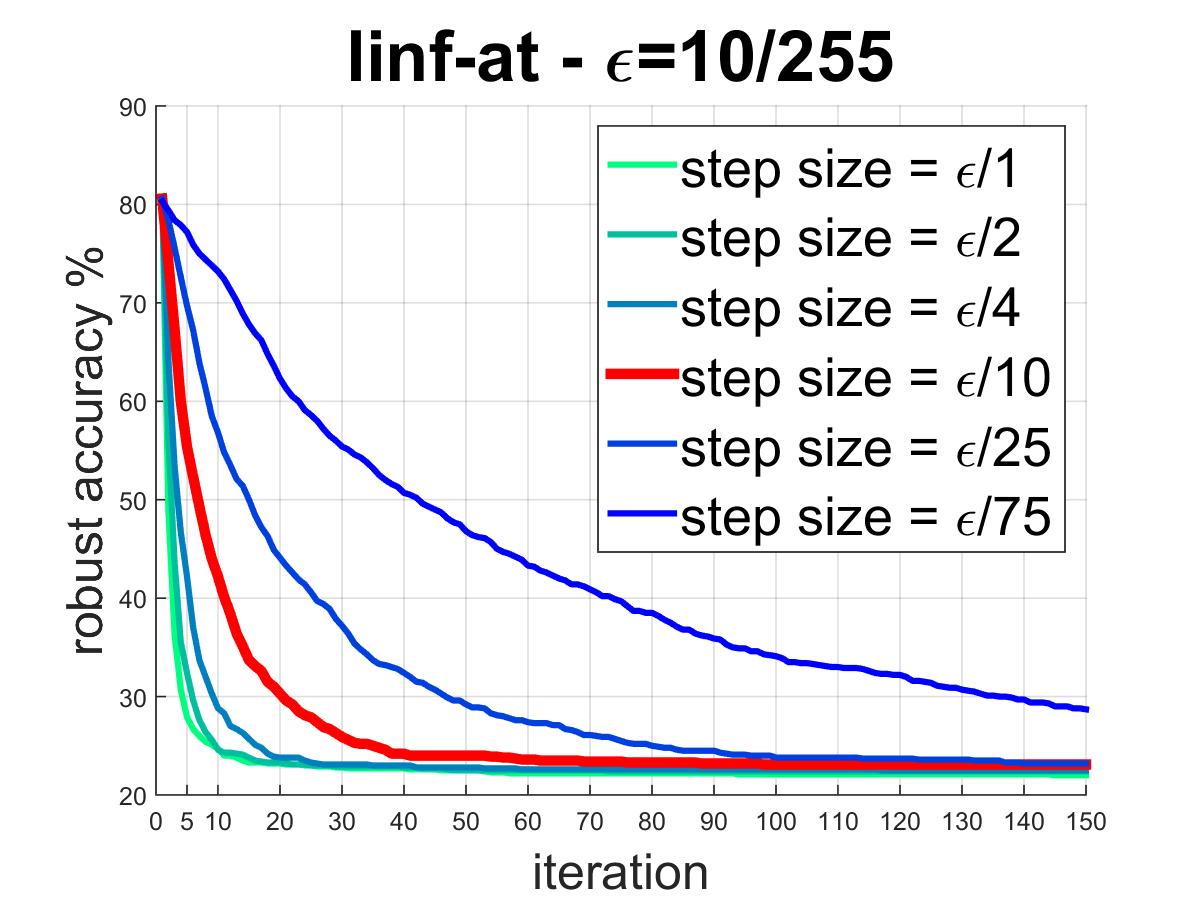} \includegraphics[width=0.5\columnwidth, clip, trim=10mm 0mm 15mm 0mm]{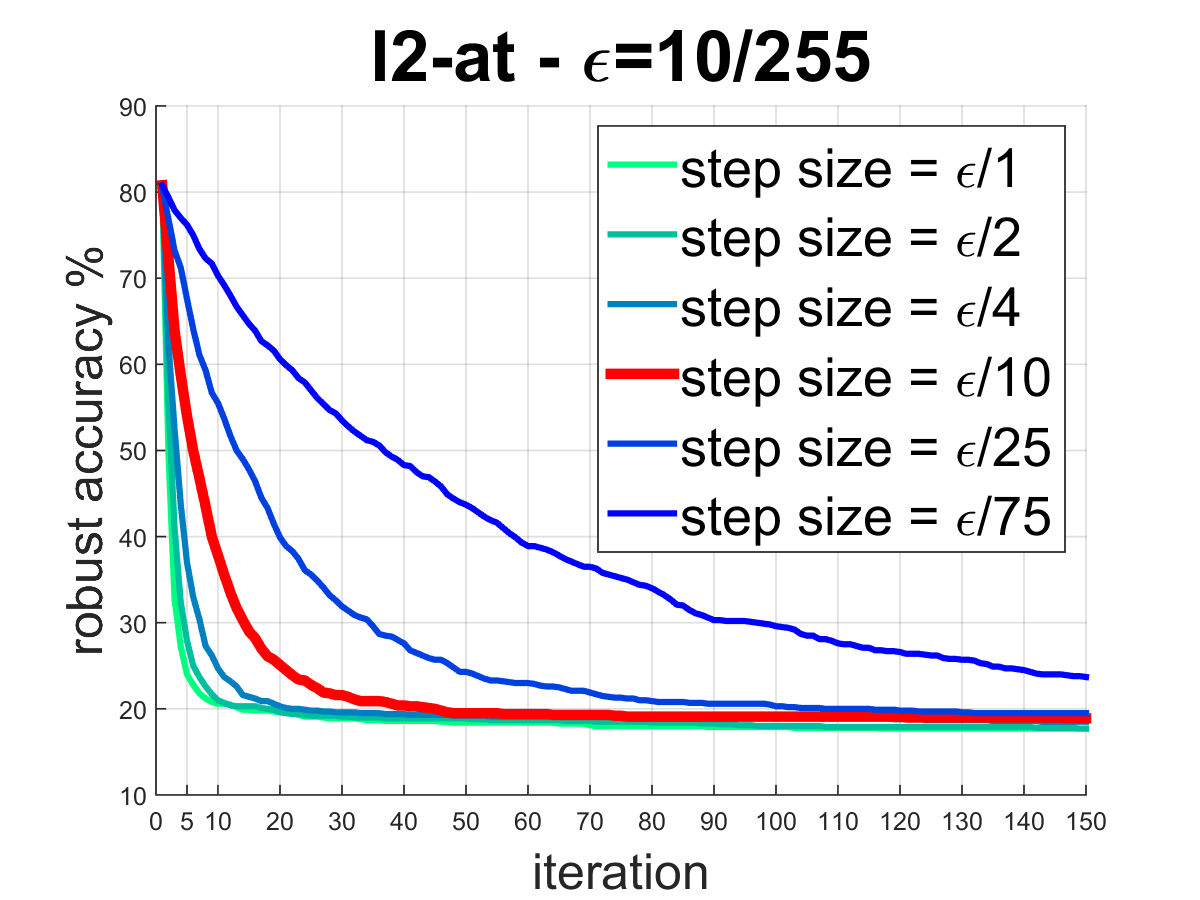}\\
		\hline
		\end{tabular} \caption{Evolution of robust accuracy as a function of the iterations for different step sizes for PGD wrt $l_\infty$. In red the step size we used in the experiments of Section \ref{sec:exps}. The models used are those trained on MNIST (top row) and CIFAR-10 (bottom row).}\label{fig:step_size_pgd_linf} \end{figure*}

\begin{figure*}[p]\centering \begin{tabular}{c}
\textbf{MNIST - $l_2$-attack} \\[2mm]
\includegraphics[width=0.5\columnwidth, clip, trim=10mm 0mm 15mm 0mm]{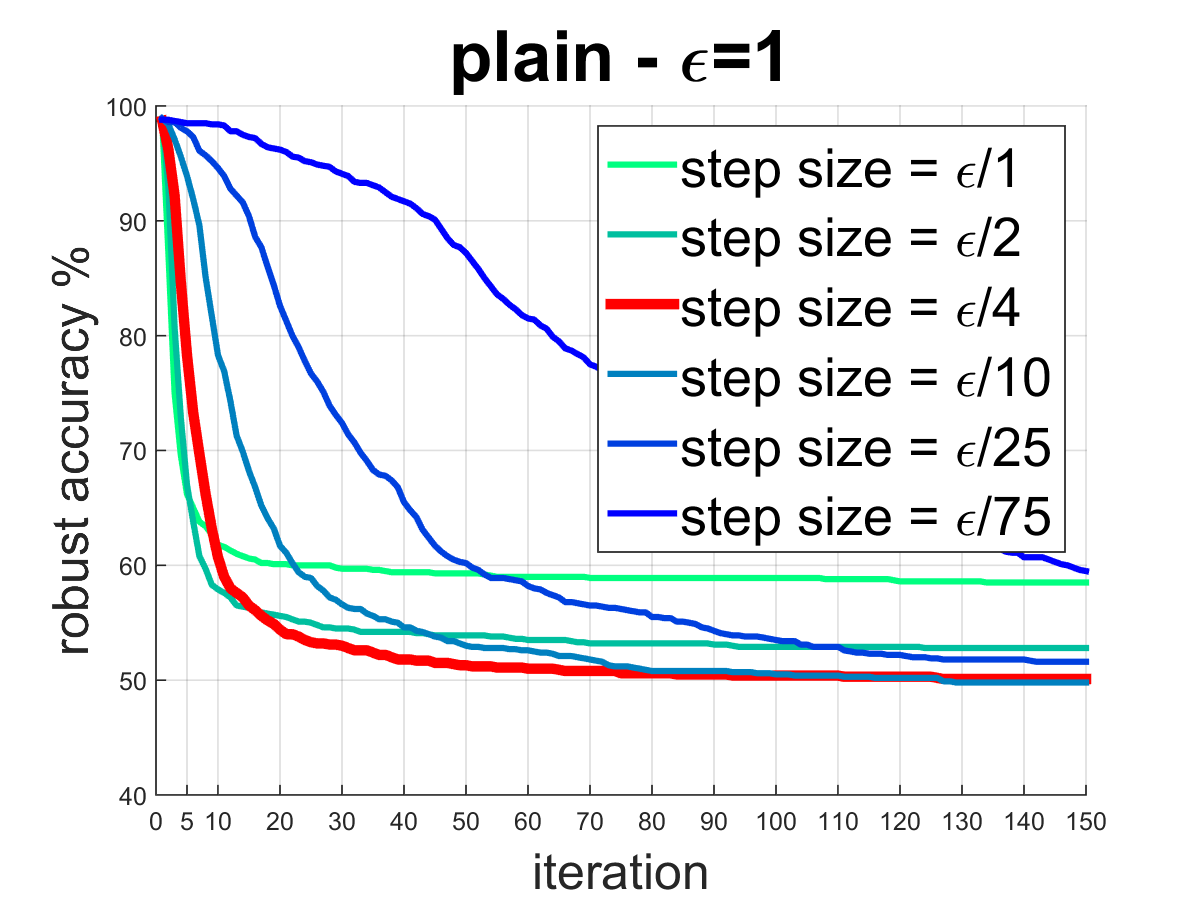} \includegraphics[width=0.5\columnwidth, clip, trim=10mm 0mm 15mm 0mm]{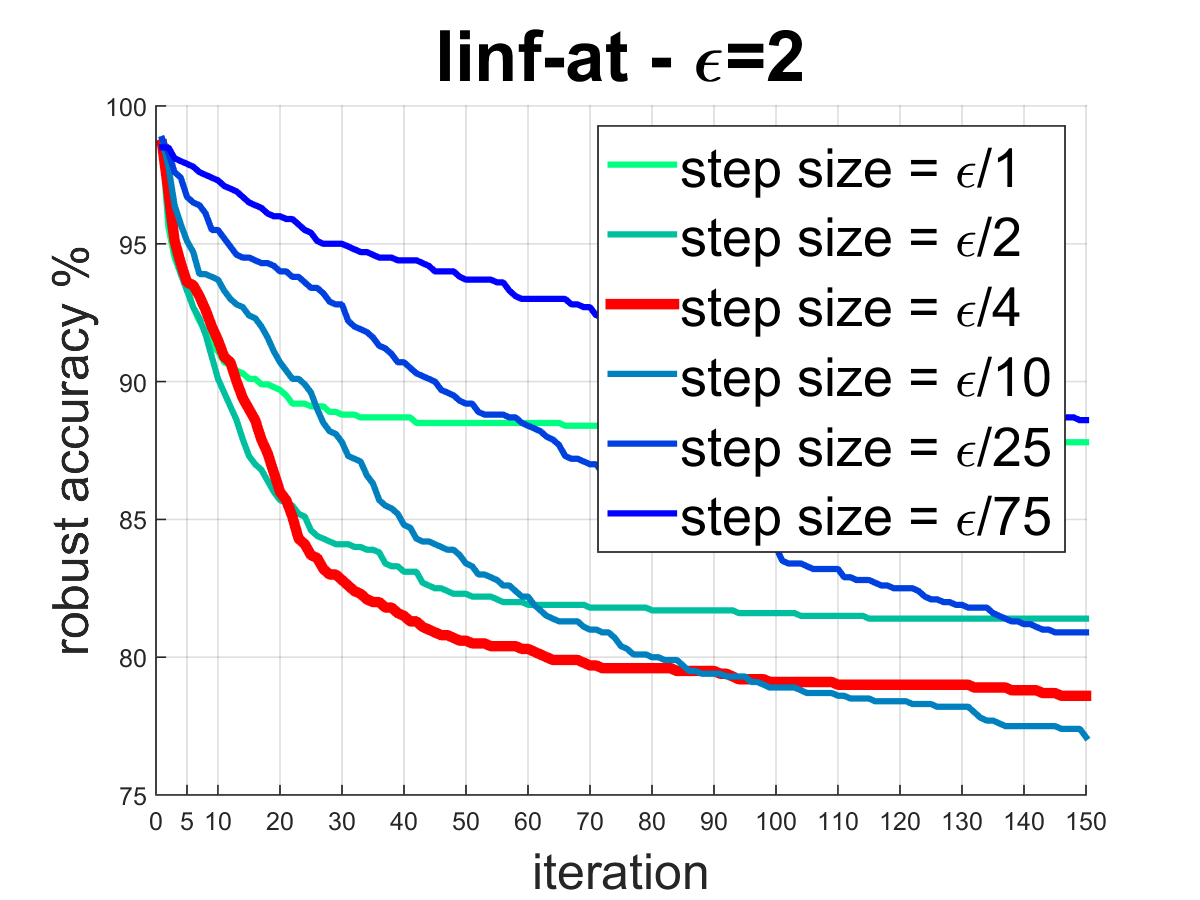} \includegraphics[width=0.5\columnwidth, clip, trim=10mm 0mm 15mm 0mm]{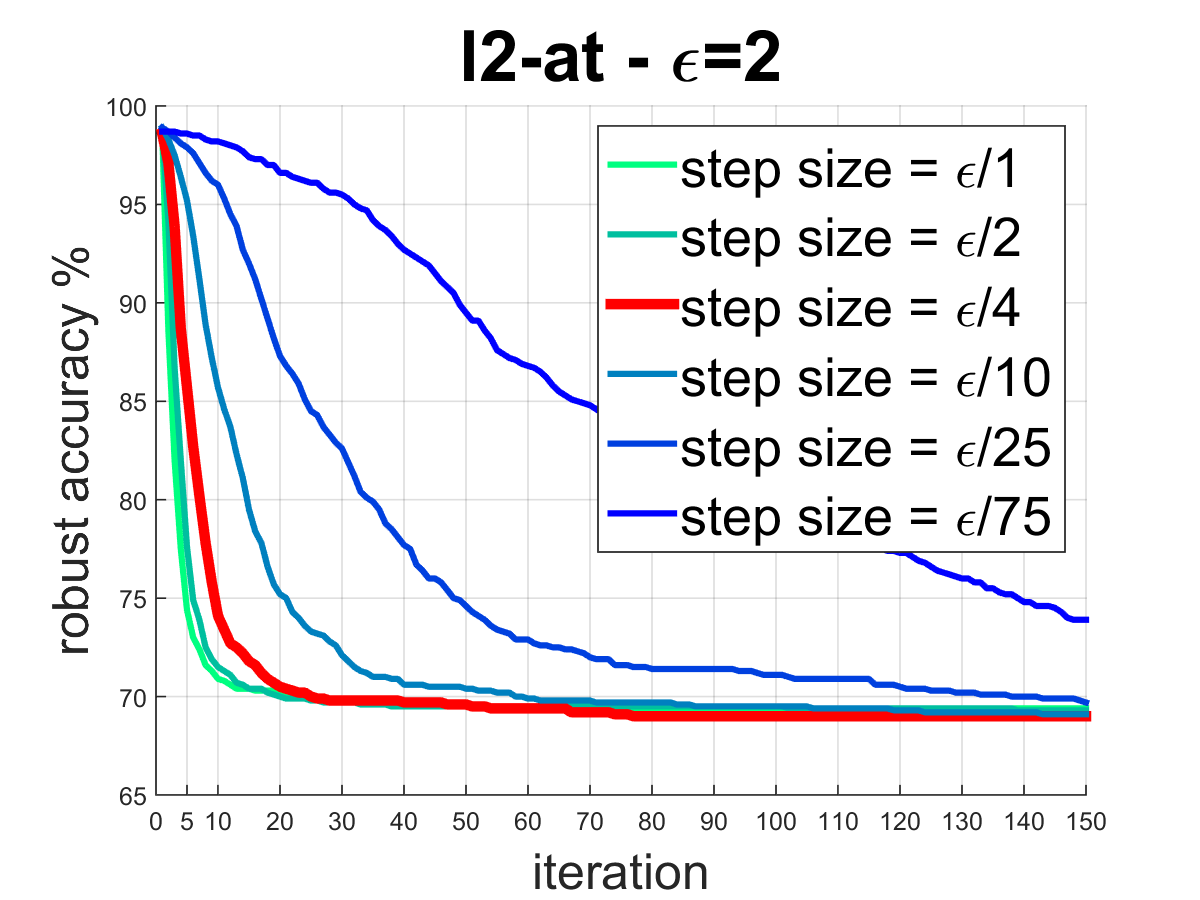}\\
\includegraphics[width=0.5\columnwidth, clip, trim=10mm 0mm 15mm 0mm]{imgs_3/pl_stepsize_l2_mnist_plain} \includegraphics[width=0.5\columnwidth, clip, trim=10mm 0mm 15mm 0mm]{imgs_3/pl_stepsize_l2_mnist_linf-at} \includegraphics[width=0.5\columnwidth, clip, trim=10mm 0mm 15mm 0mm]{imgs_3/pl_stepsize_l2_mnist_l2-at}\\
\hline
\\
\textbf{CIFAR-10 - $l_2$-attack}\\[2mm]
\includegraphics[width=0.5\columnwidth, clip, trim=10mm 0mm 15mm 0mm]{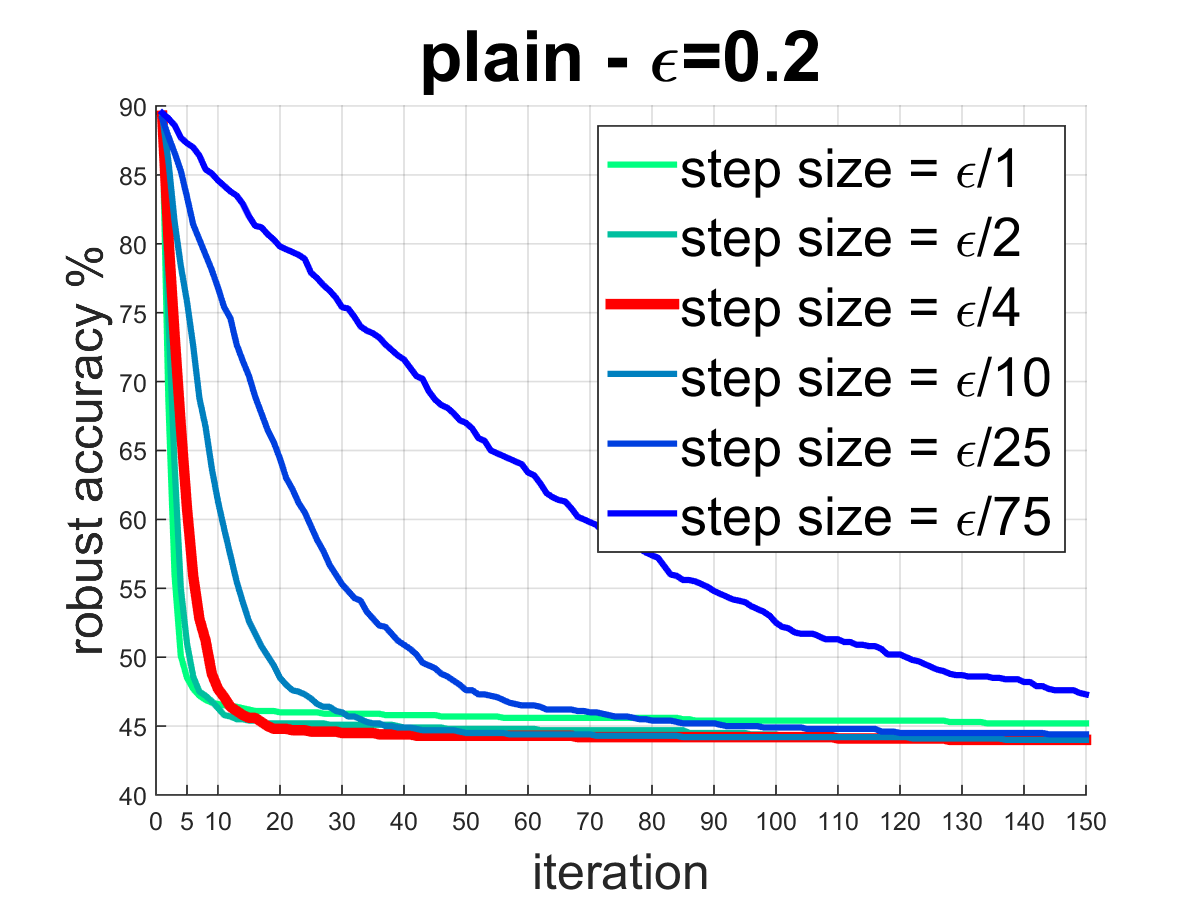} \includegraphics[width=0.5\columnwidth, clip, trim=10mm 0mm 15mm 0mm]{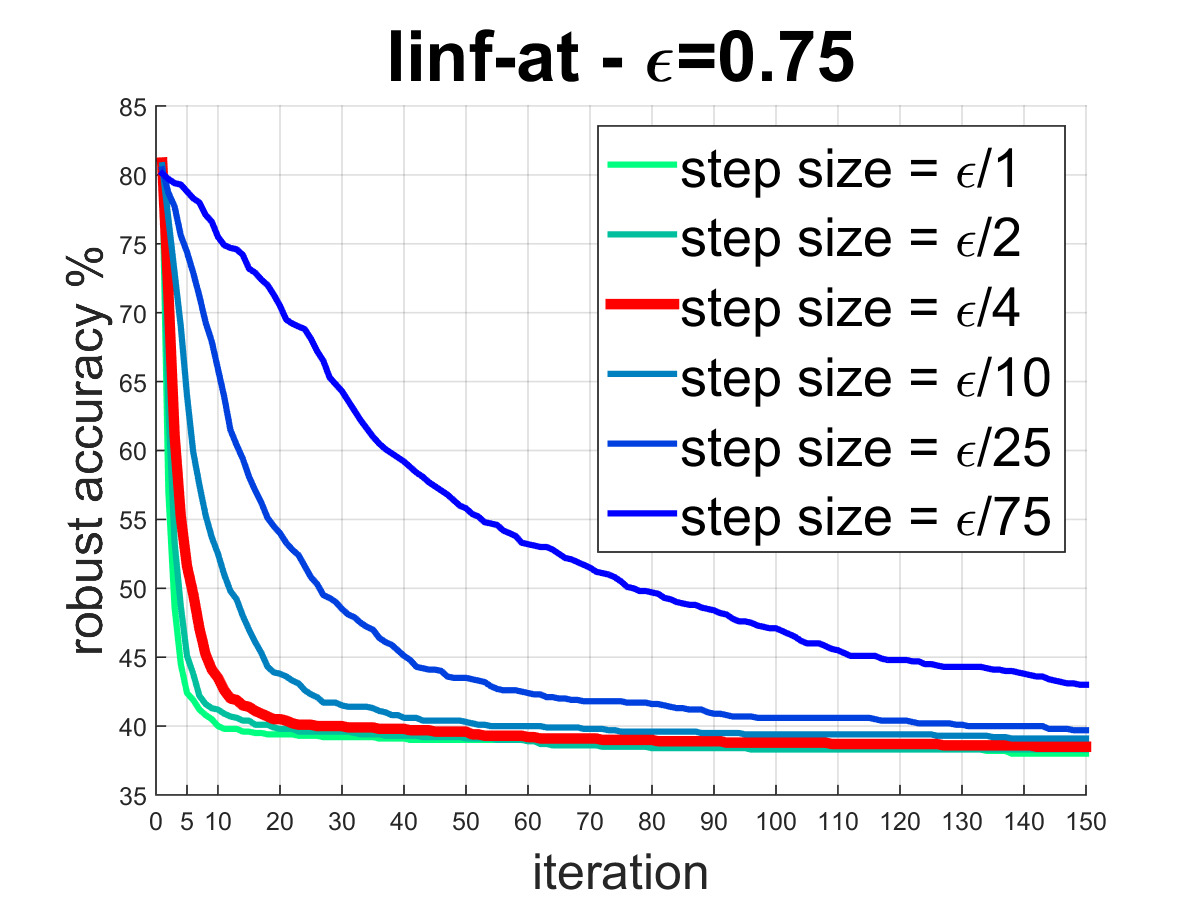} \includegraphics[width=0.5\columnwidth, clip, trim=10mm 0mm 15mm 0mm]{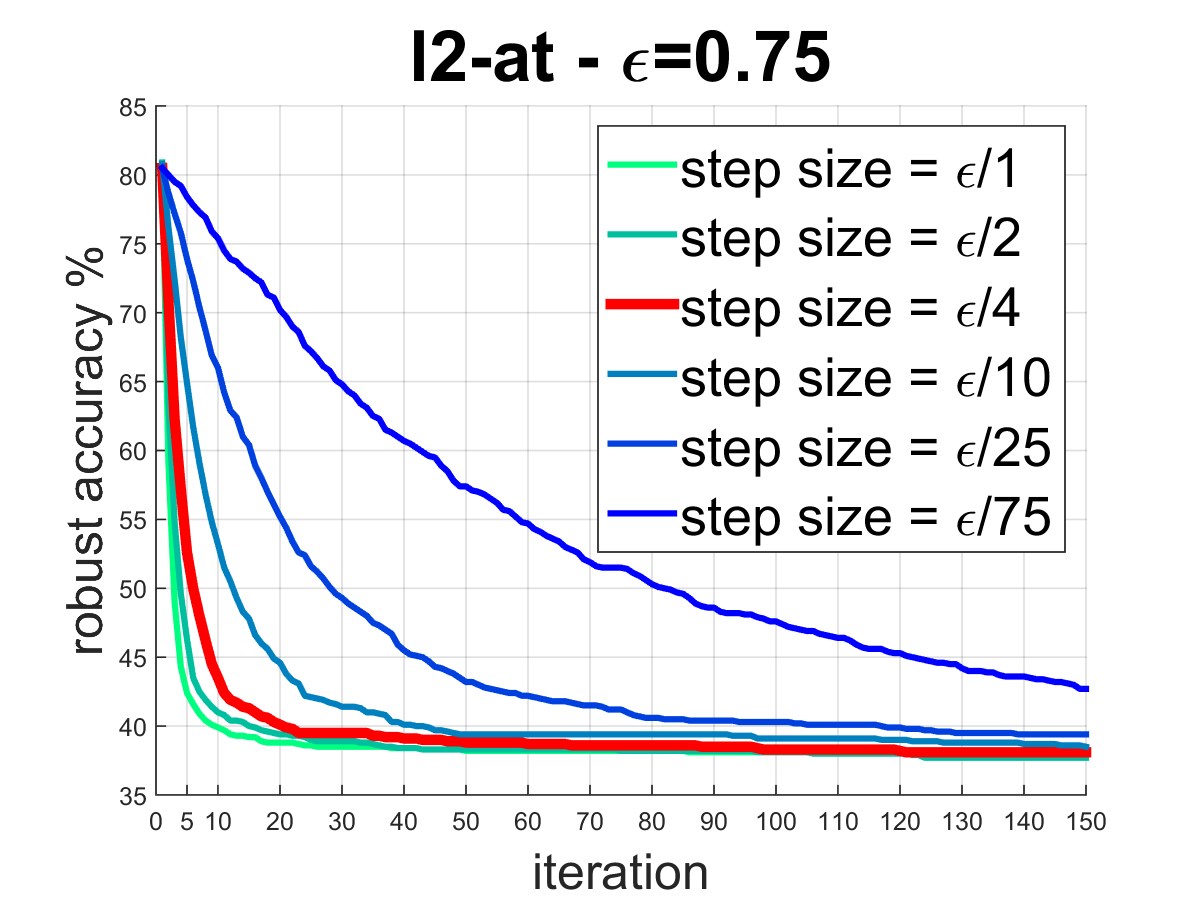}\\
\includegraphics[width=0.5\columnwidth, clip, trim=10mm 0mm 15mm 0mm]{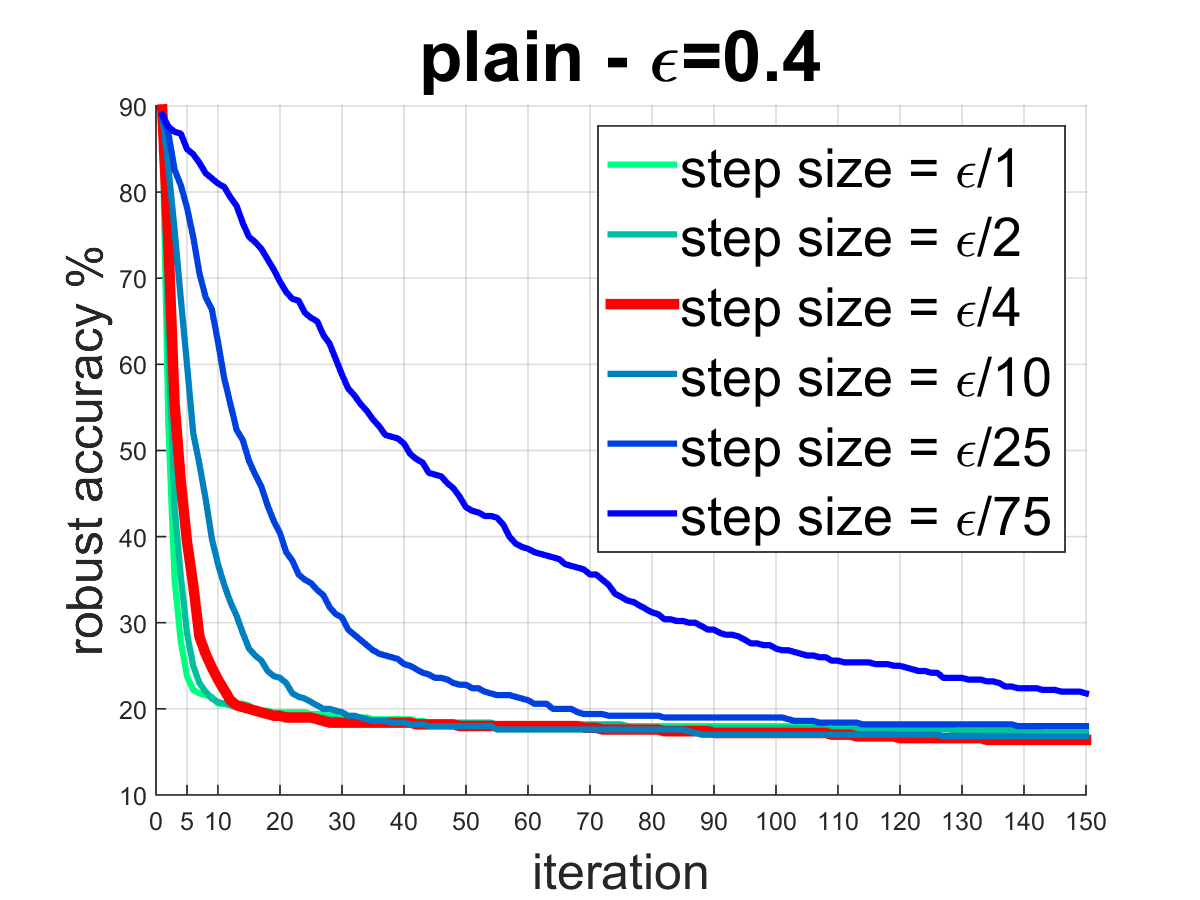} \includegraphics[width=0.5\columnwidth, clip, trim=10mm 0mm 15mm 0mm]{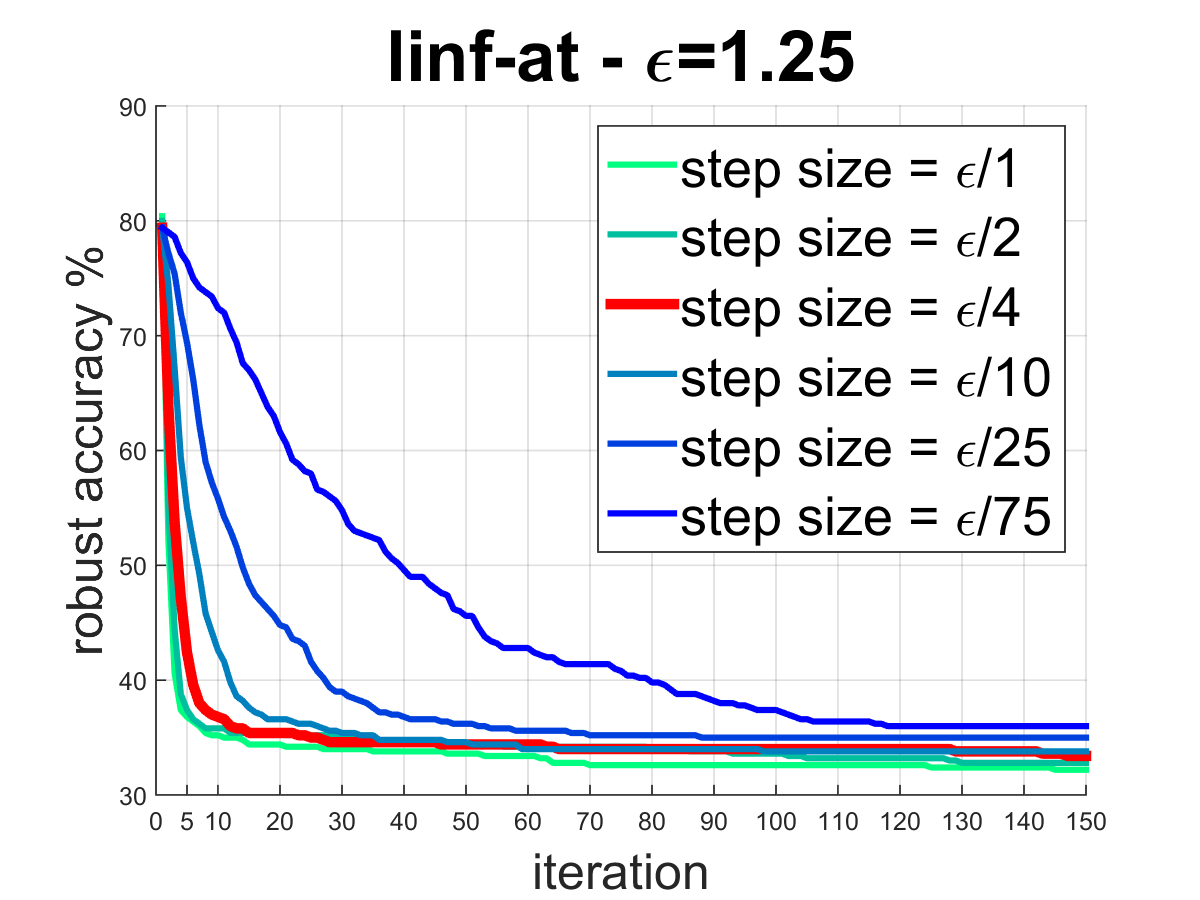} \includegraphics[width=0.5\columnwidth, clip, trim=10mm 0mm 15mm 0mm]{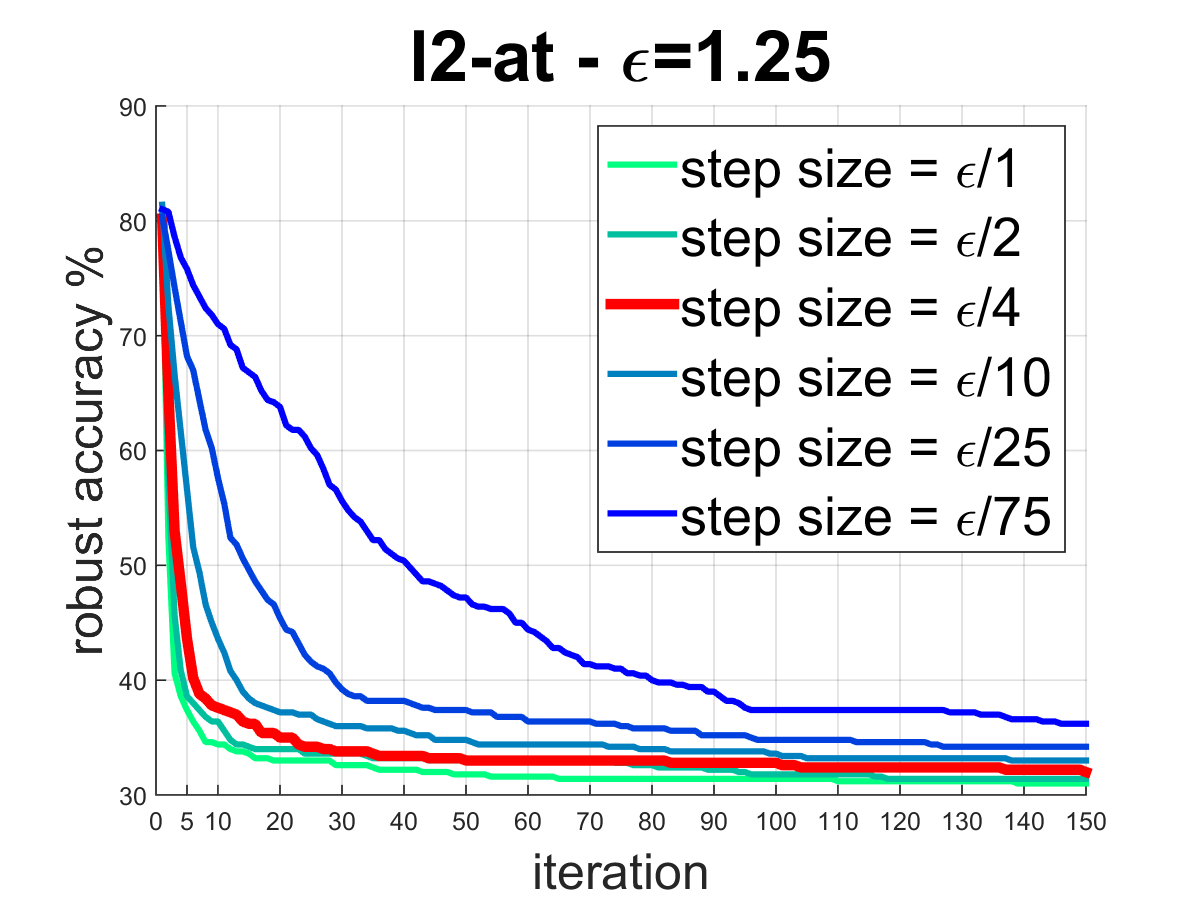}\\
\hline
\end{tabular}
\caption{Evolution of robust accuracy as a function of the iterations for different step sizes for PGD wrt $l_2$. In red the step size we used in the experiments of Section \ref{sec:exps}. The models used are those trained on MNIST (top row) and CIFAR-10 (bottom row).}\label{fig:step_size_pgd}
\end{figure*}

\begin{figure*}[p]\centering \begin{tabular}{c}
		\textbf{MNIST - $l_1$-attack} \\[2mm]
		\includegraphics[width=0.5\columnwidth, clip, trim=10mm 0mm 15mm 0mm]{imgs_3/pl_stepsize_l1_mnist_plain_smalleps} \includegraphics[width=0.5\columnwidth, clip, trim=10mm 0mm 15mm 0mm]{imgs_3/pl_stepsize_l1_mnist_linf-at_smalleps} \includegraphics[width=0.5\columnwidth, clip, trim=10mm 0mm 15mm 0mm]{imgs_3/pl_stepsize_l1_mnist_l2-at_smalleps}\\
		\includegraphics[width=0.5\columnwidth, clip, trim=10mm 0mm 15mm 0mm]{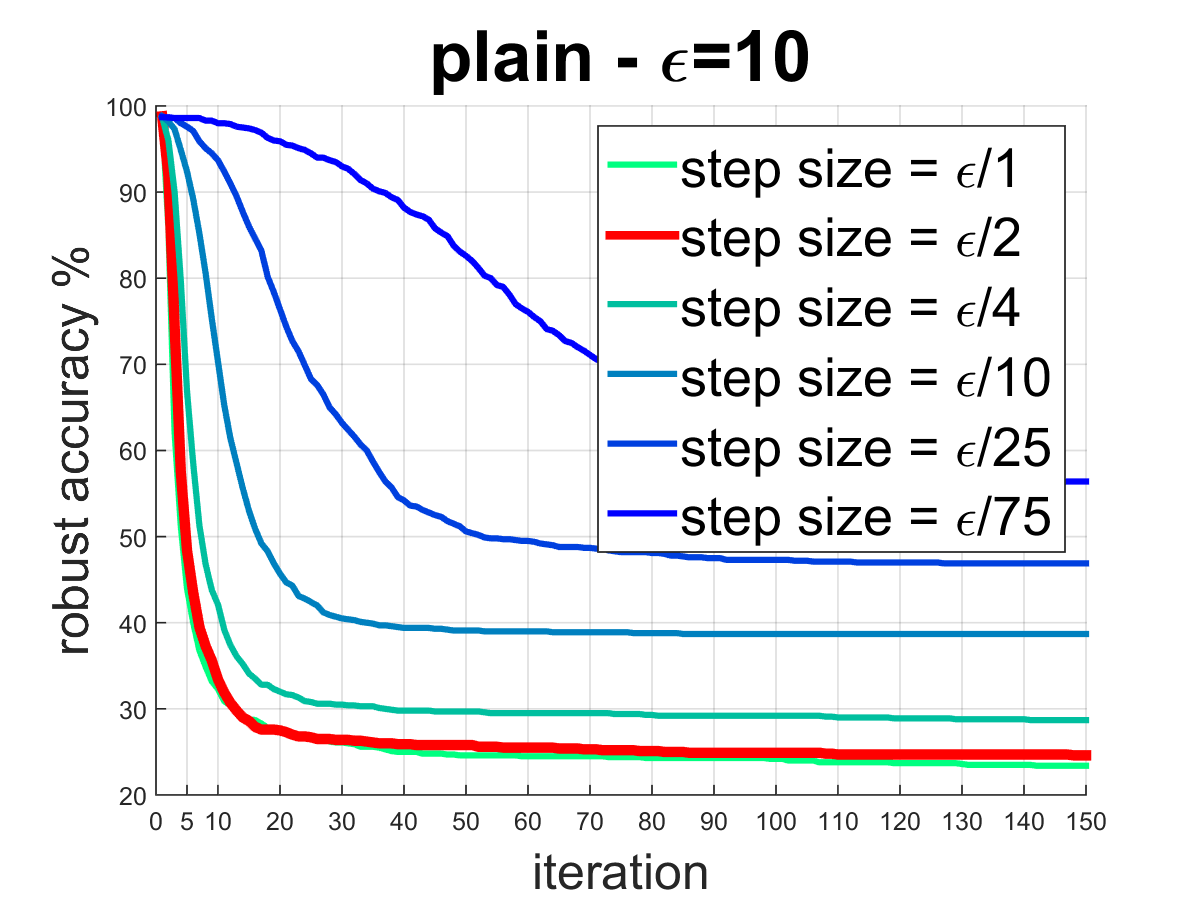} \includegraphics[width=0.5\columnwidth, clip, trim=10mm 0mm 15mm 0mm]{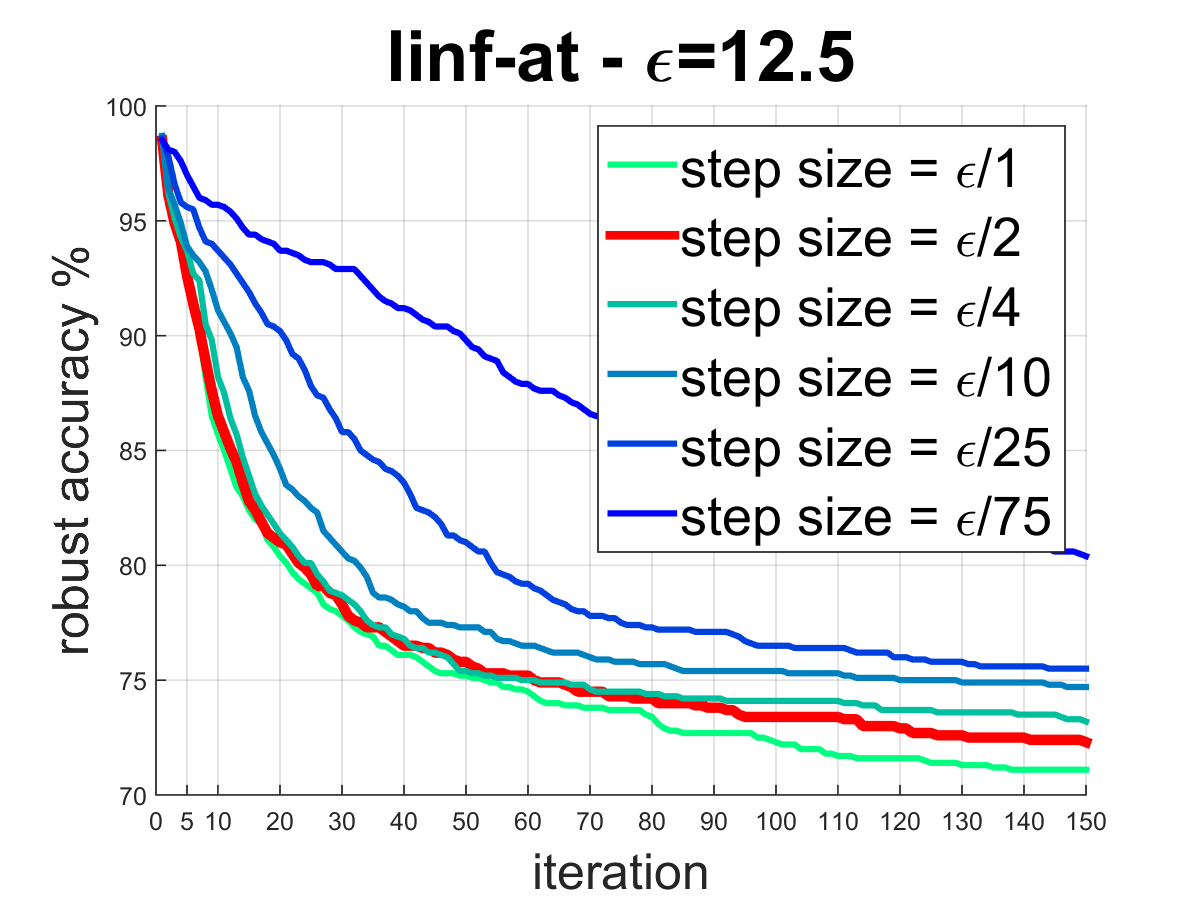} \includegraphics[width=0.5\columnwidth, clip, trim=10mm 0mm 15mm 0mm]{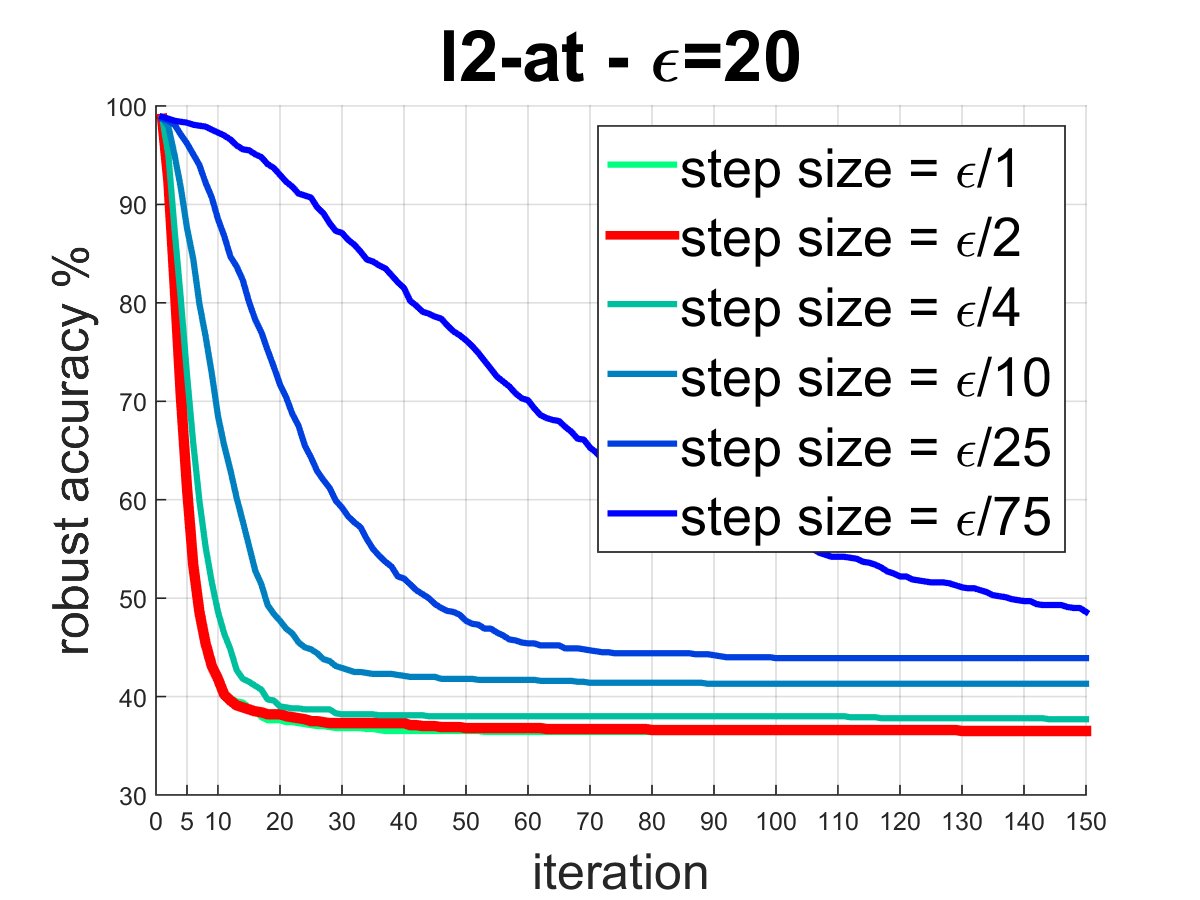}\\
		\hline
		\\
		%
		%\commentout{
		\textbf{CIFAR-10 - $l_1$-attack}\\[2mm]
		\includegraphics[width=0.5\columnwidth, clip, trim=10mm 0mm 15mm 0mm]{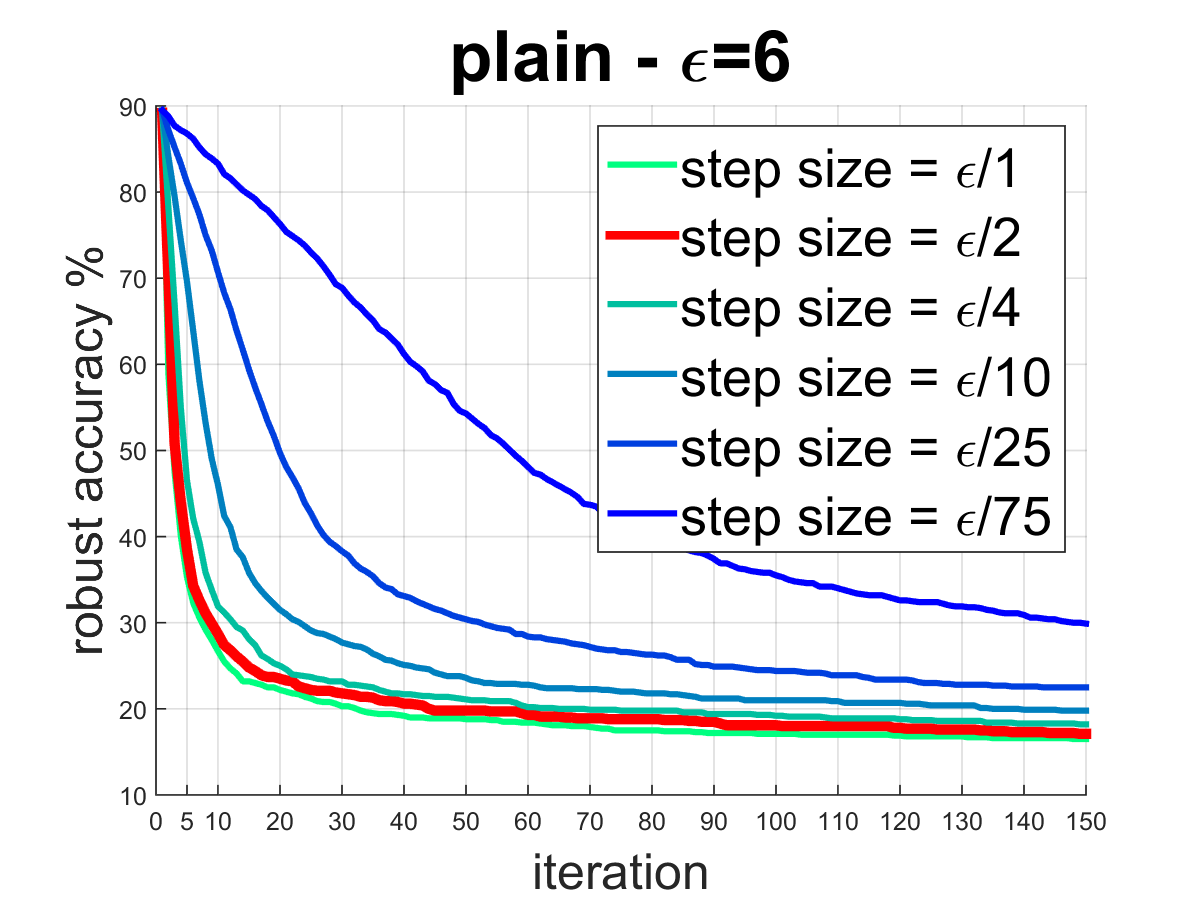} \includegraphics[width=0.5\columnwidth, clip, trim=10mm 0mm 15mm 0mm]{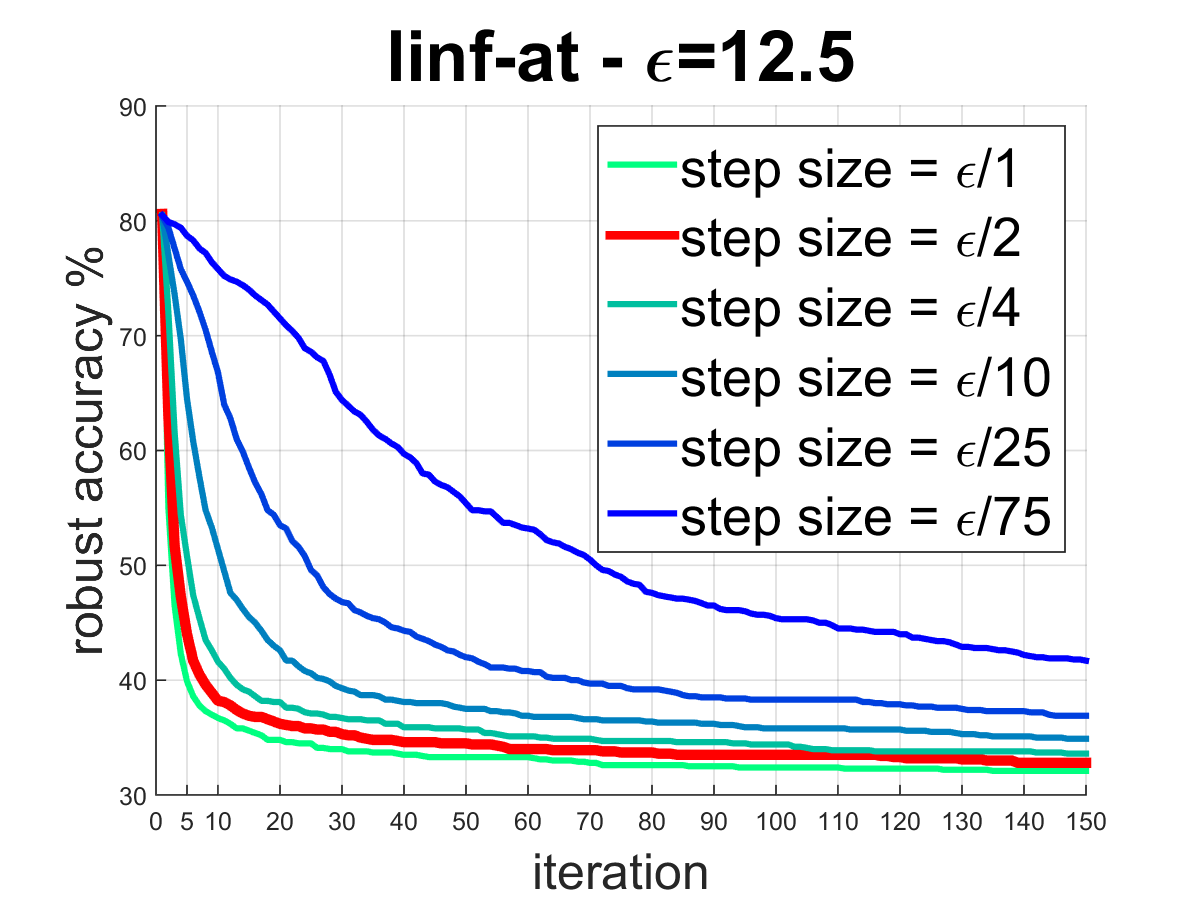} \includegraphics[width=0.5\columnwidth, clip, trim=10mm 0mm 15mm 0mm]{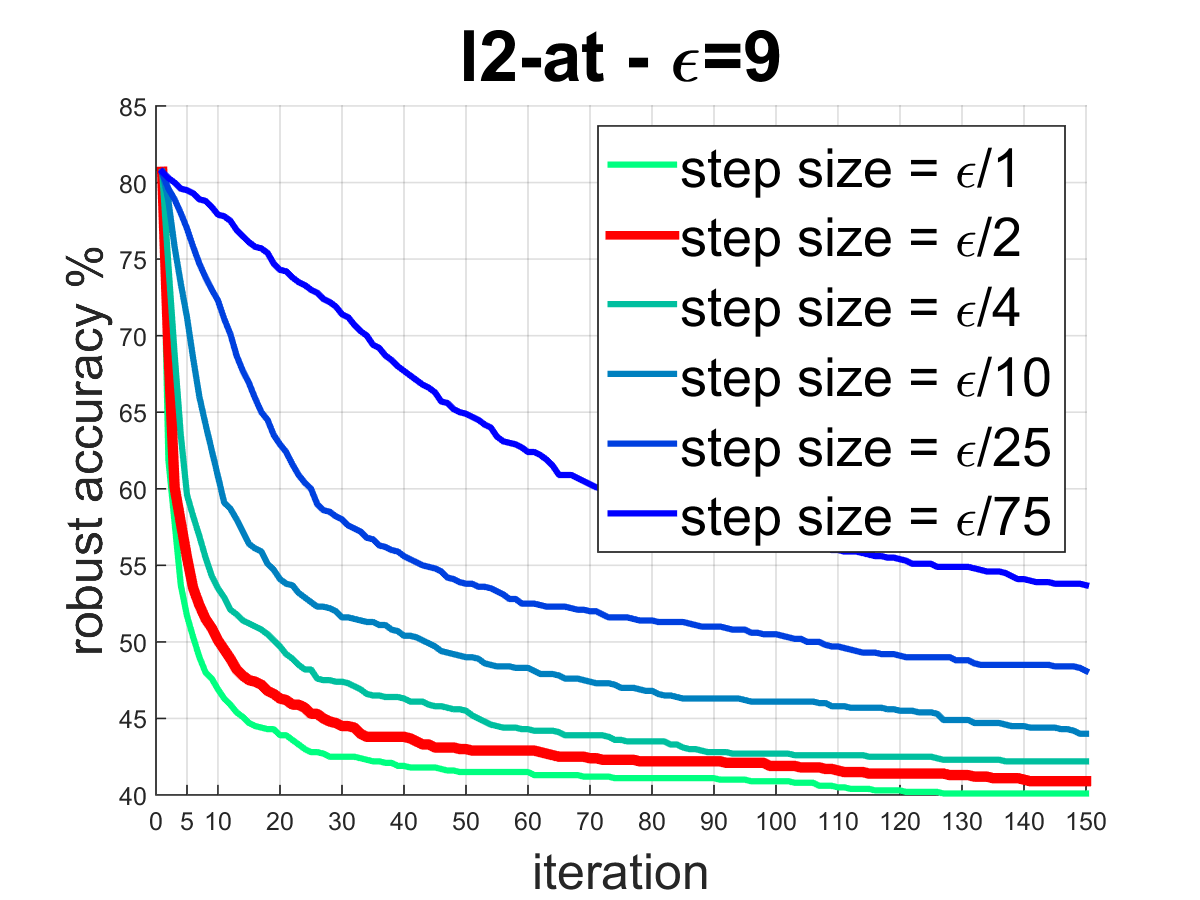}\\
		\includegraphics[width=0.5\columnwidth, clip, trim=10mm 0mm 15mm 0mm]{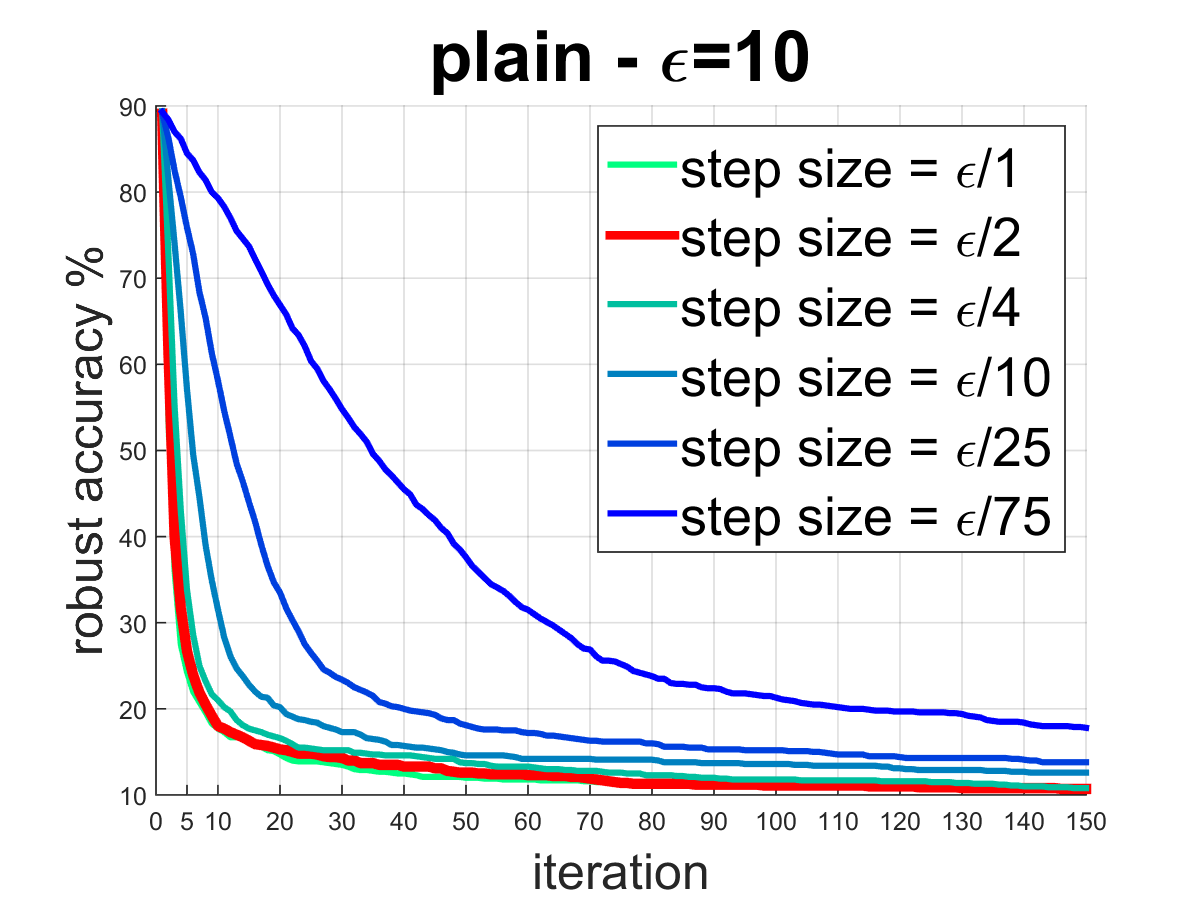} \includegraphics[width=0.5\columnwidth, clip, trim=10mm 0mm 15mm 0mm]{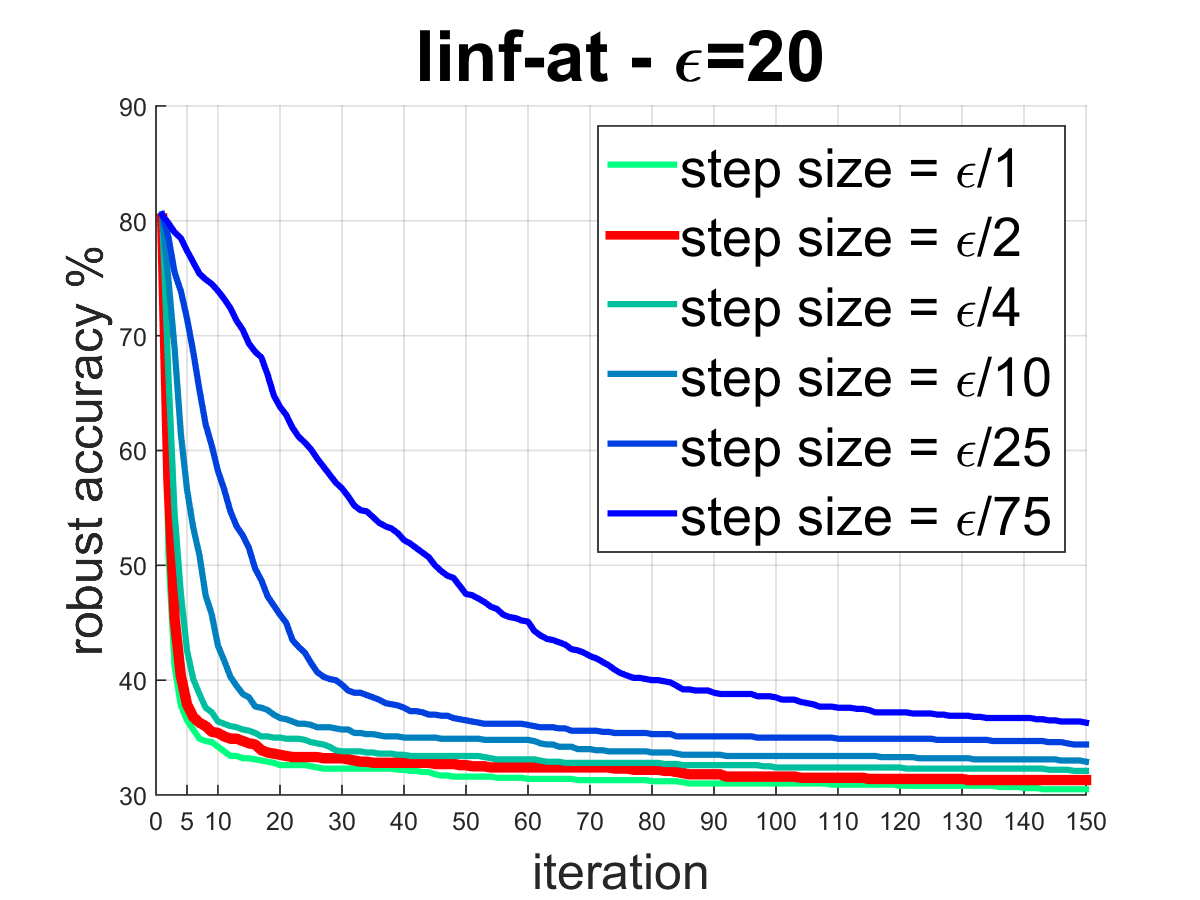} \includegraphics[width=0.5\columnwidth, clip, trim=10mm 0mm 15mm 0mm]{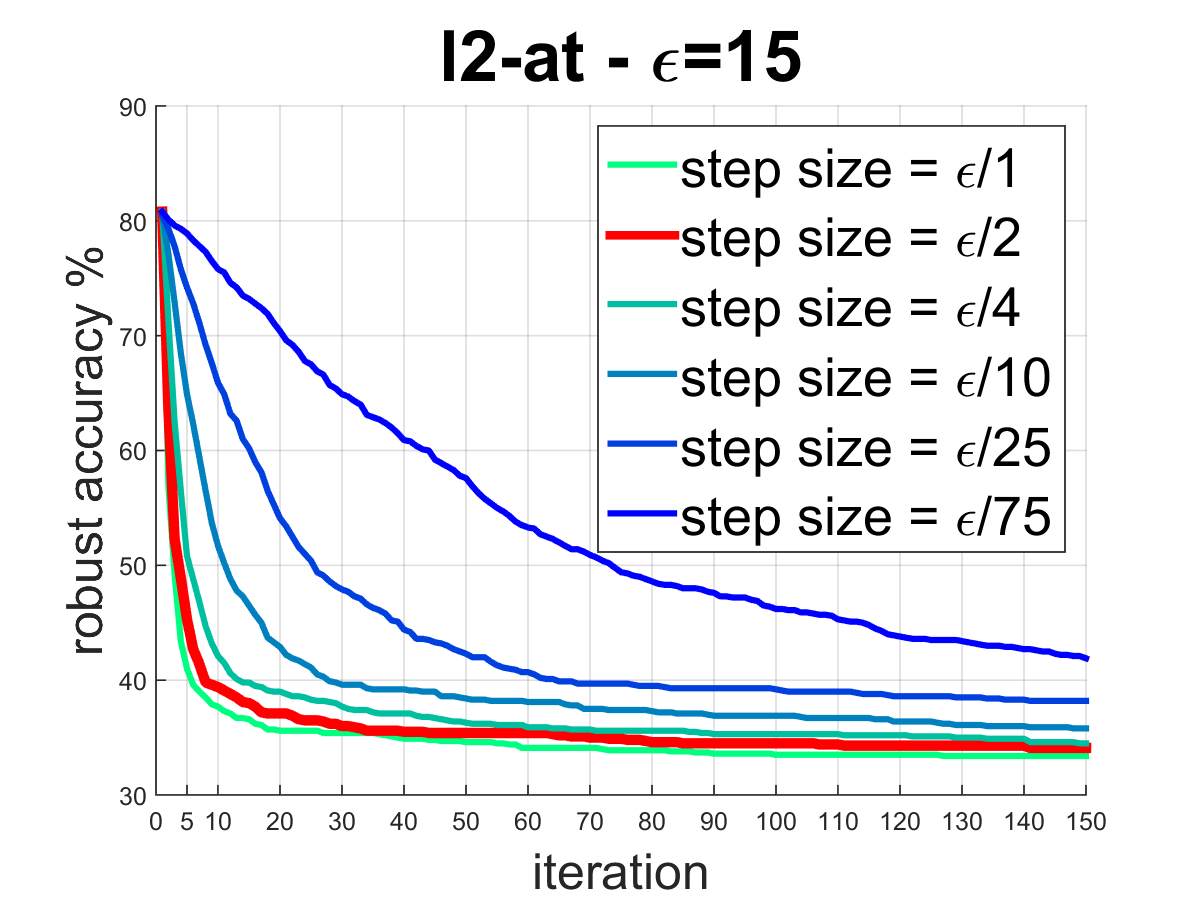}\\
		%}
		\hline
		\end{tabular} \caption{Evolution of robust accuracy as a function of the iterations for different step sizes for PGD wrt $l_1$. In red the step size we used in the experiments of Section \ref{sec:exps}. The models used are those trained on MNIST (top row) and CIFAR-10 (bottom row).} \label{fig:step_size_pgd_l1}
	\end{figure*}

\subsection{Choice of the step size of PGD} \label{sec:app_pgd_step_size}
We here show the performance of PGD wrt $l_1,l_2,l_\infty$ on MNIST (top row of Figure \ref{fig:step_size_pgd}) and CIFAR-10 (bottom row of Figure \ref{fig:step_size_pgd}) for different choices of the step size. In particular we focus on large $\epsilon$.
We report the robust accuracy as a function of iterations (in total 150 iterations). We test step sizes $\nicefrac{\epsilon}{t}$ for $t\in\{1, 2, 4, 10, 25, 75\}$. For each step size we show the \emph{best} run out of 10, with random initialization, which achieve the lowest robust accuracy after 150 iterations.
%Note however that the behaviour of different runs vary minimally.
\begin{itemize}
\item $l_2$: In Figure \ref{fig:step_size_pgd} we show, for each dataset, the results for the three models used in Section \ref{sec:exps}, with decreasing step size corresponding to darker shades of blue, while our chosen step size for $l_2$, that is $\nicefrac{\epsilon}{4}$, is highlighted in red. We see that it achieves in all the models best or close to best robust accuracy.
\item $l_\infty$: We show in Figure \ref{fig:step_size_pgd_linf} the same plots as for $l_2$ but instead for $l_\infty$, the chosen stepsize $\nicefrac{\epsilon}{10}$ for $l_\infty$-attacks is highlighted in red. Again, we see that it is on average performing best.
\item $l_1$: We show in Figure \ref{fig:step_size_pgd_l1} the same plots as for $l_2$ but instead for $l_1$, the chosen stepsize $\nicefrac{\epsilon}{2}$ for $l_1$-attacks is highlighted in red. Again, we see that it is on average performing best.
\end{itemize}

\begin{figure*}[p]\centering
	\begin{tabular}{c}
		\multicolumn{1}{c}{\textbf{MNIST, $l_\infty$}}\\[2mm]
		\includegraphics[width=0.5\columnwidth, clip, trim=10mm 0mm 15mm 0mm]{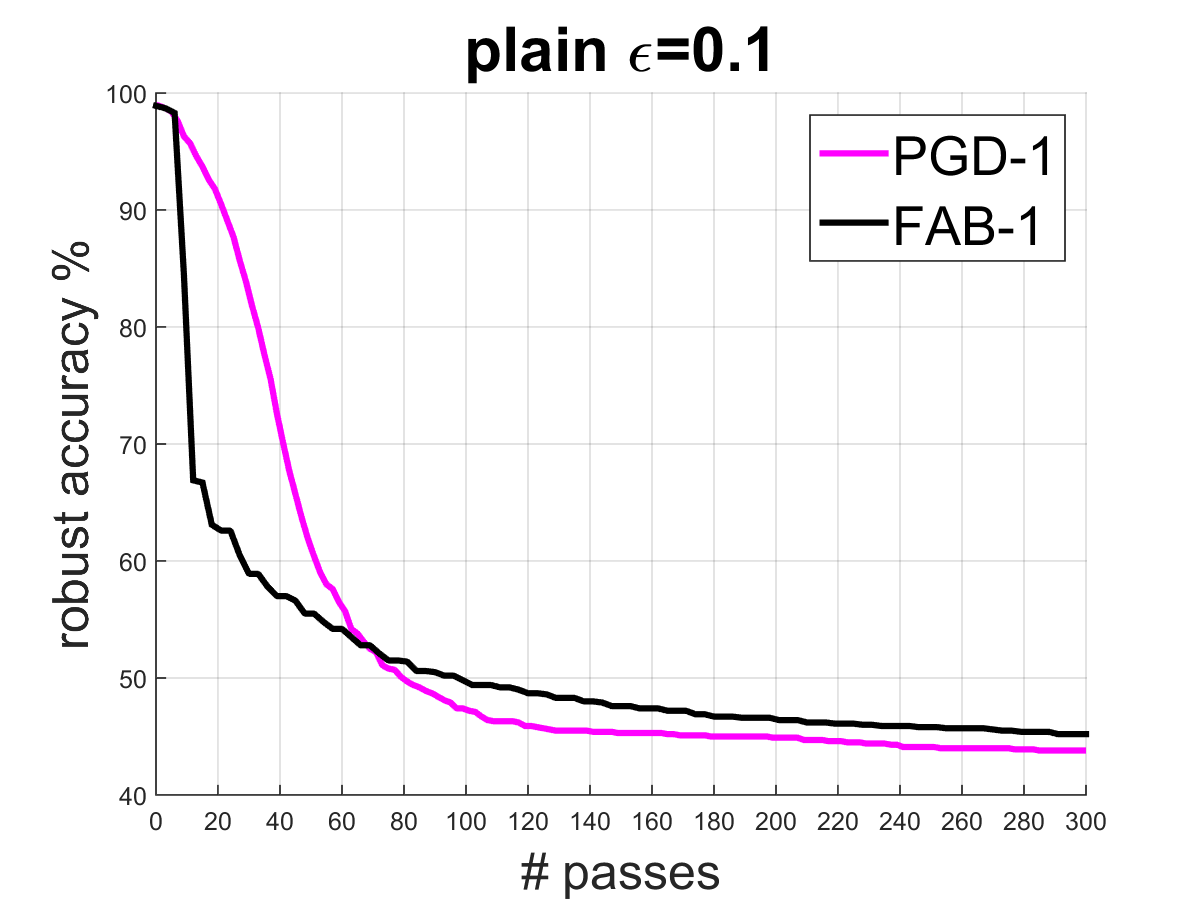}\includegraphics[width=0.5\columnwidth, clip, trim=10mm 0mm 15mm 0mm]{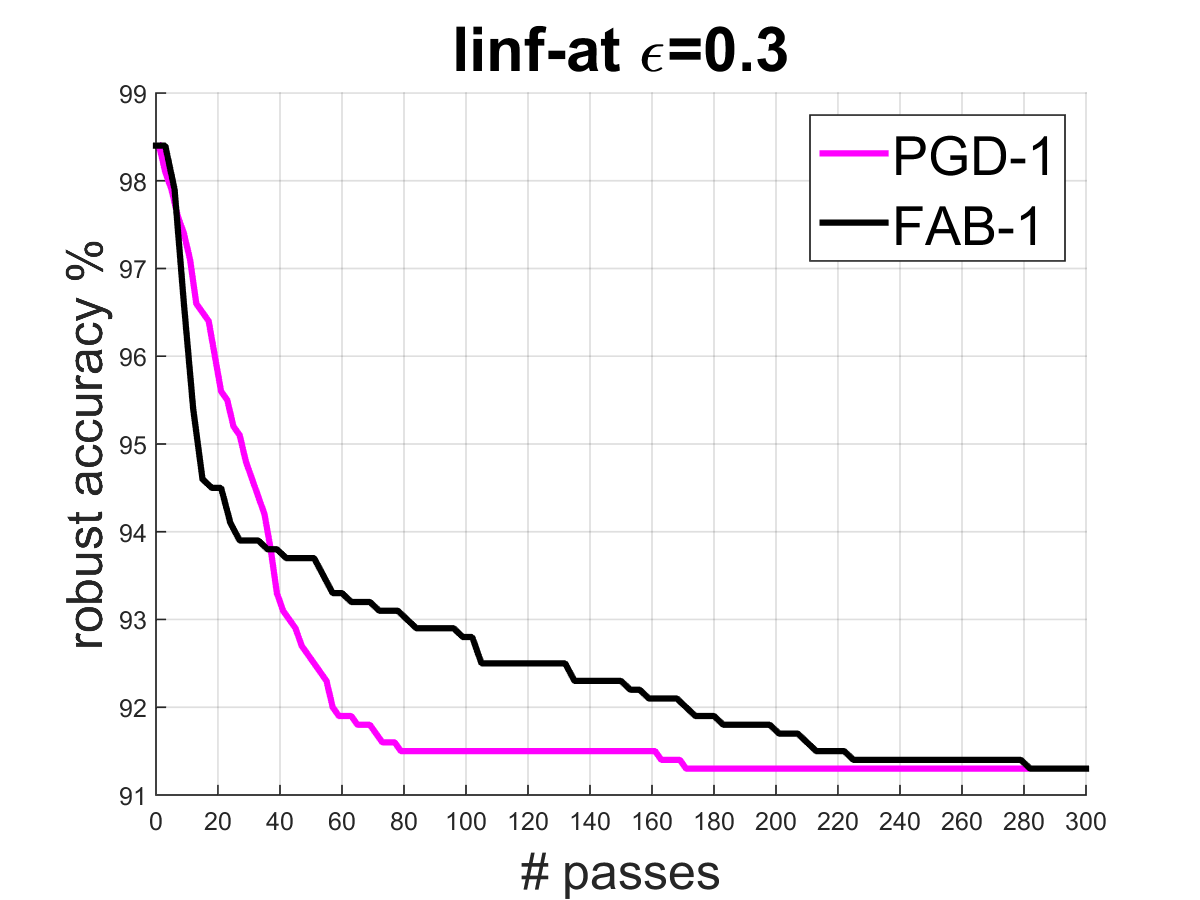} \includegraphics[width=0.5\columnwidth, clip, trim=10mm 0mm 15mm 0mm]{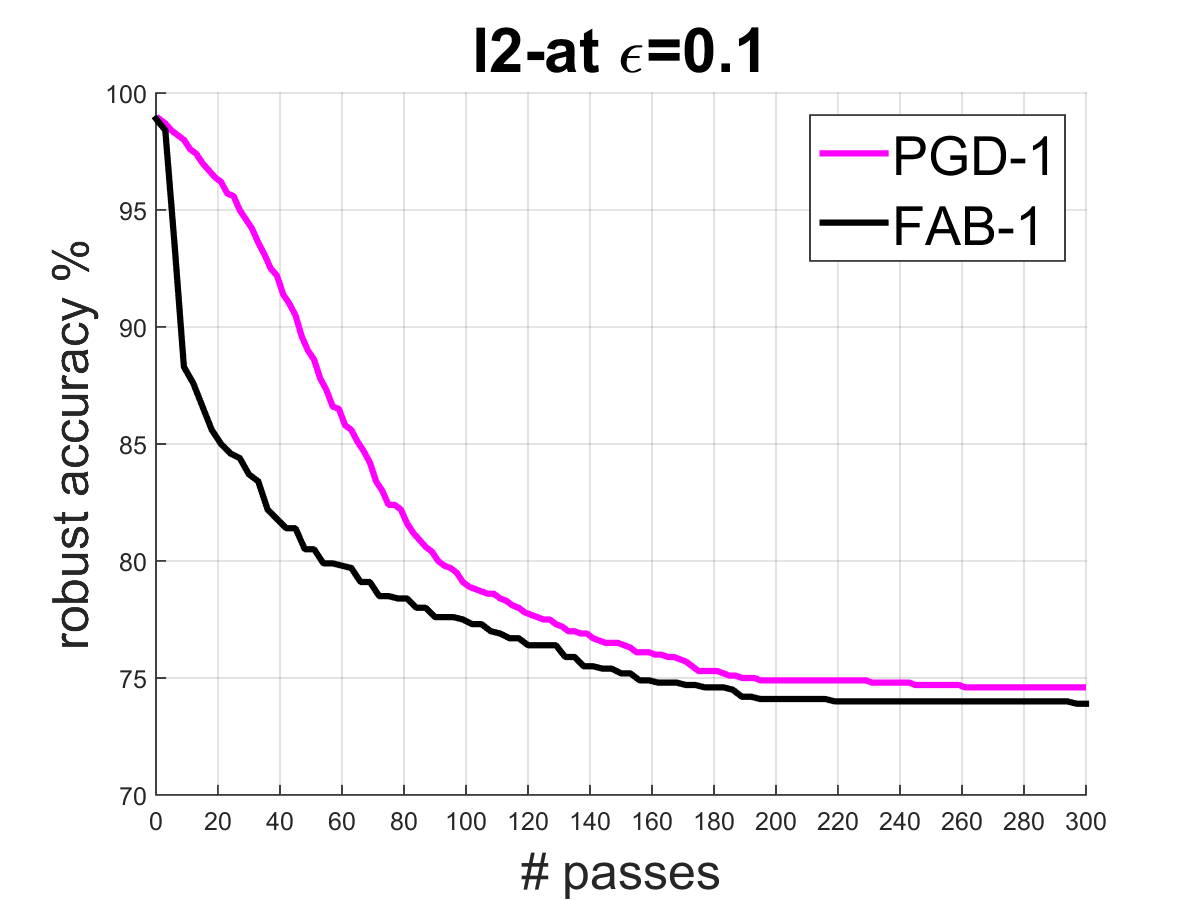}\\
		\hline
		\\
		\multicolumn{1}{c}{\textbf{MNIST, $l_2$}}\\[2mm]
		\includegraphics[width=0.5\columnwidth, clip, trim=10mm 0mm 15mm 0mm]{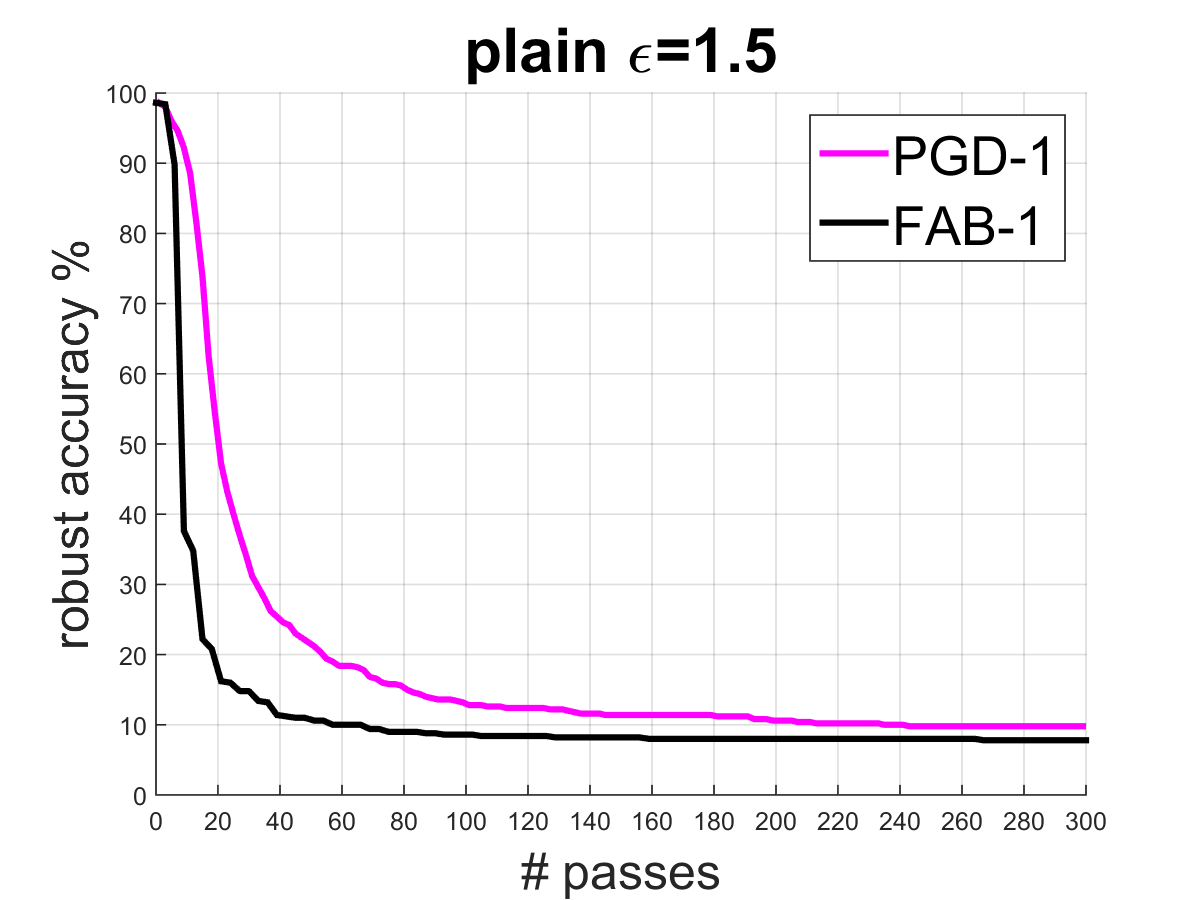}\includegraphics[width=0.5\columnwidth, clip, trim=10mm 0mm 15mm 0mm]{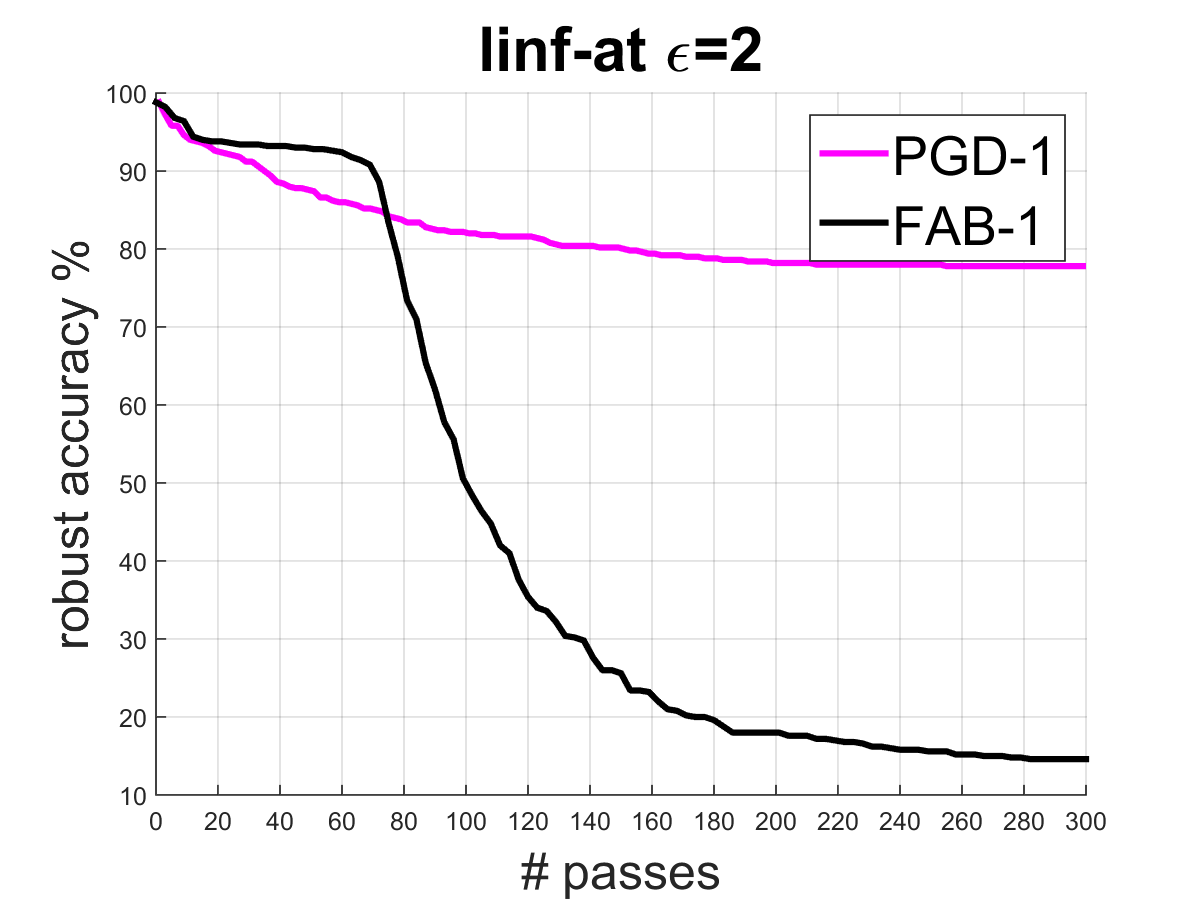} \includegraphics[width=0.5\columnwidth, clip, trim=10mm 0mm 15mm 0mm]{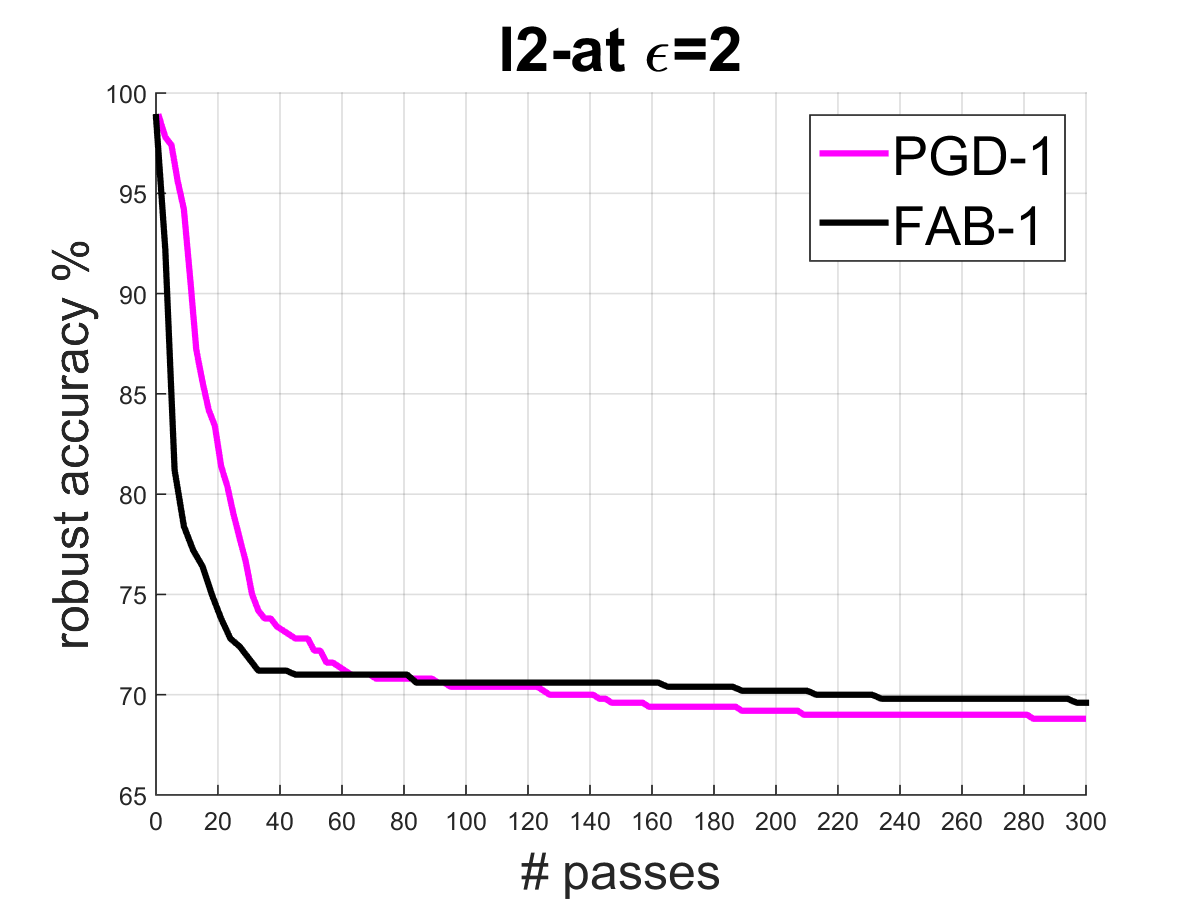}\\
		\hline
		\\
		\multicolumn{1}{c}{\textbf{MNIST, $l_1$}}\\[2mm]
		\includegraphics[width=0.5\columnwidth, clip, trim=10mm 0mm 15mm 0mm]{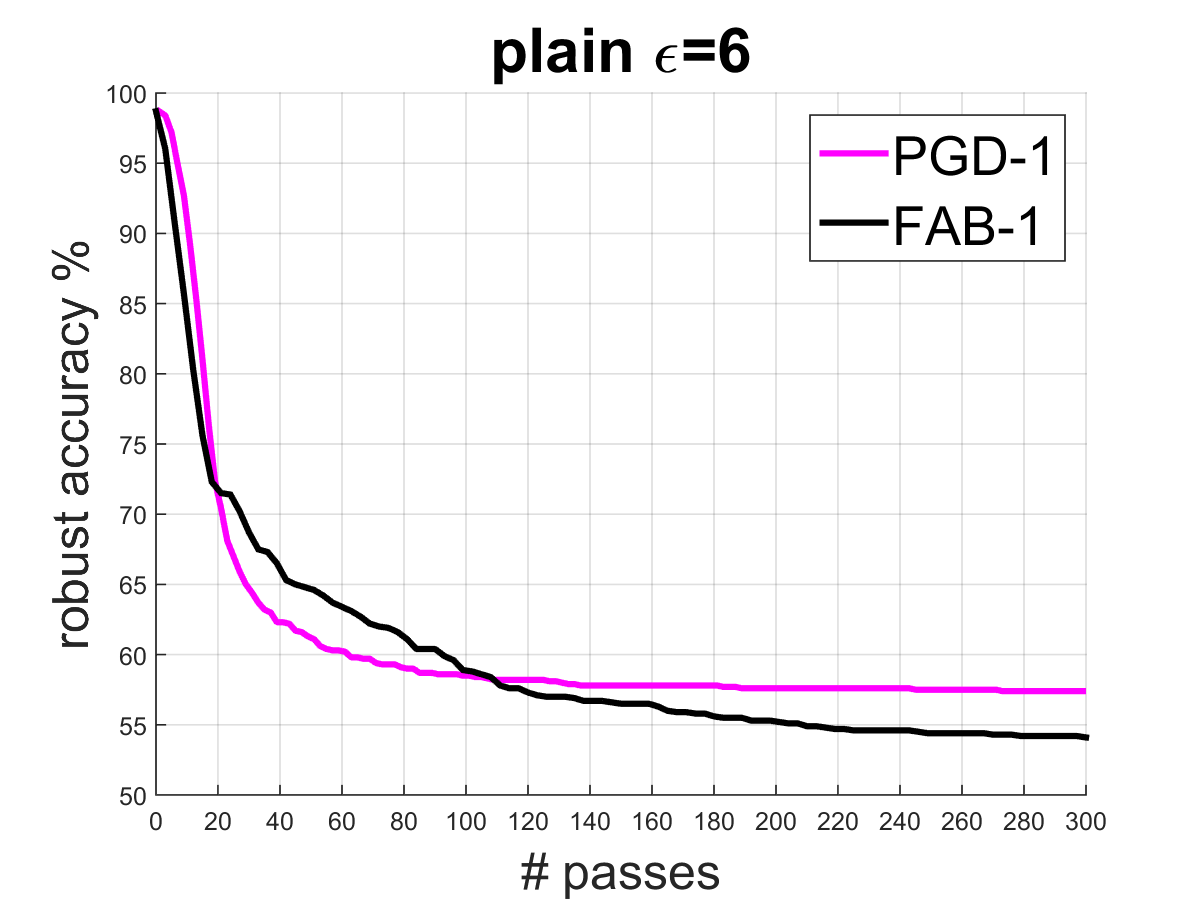}\includegraphics[width=0.5\columnwidth, clip, trim=10mm 0mm 15mm 0mm]{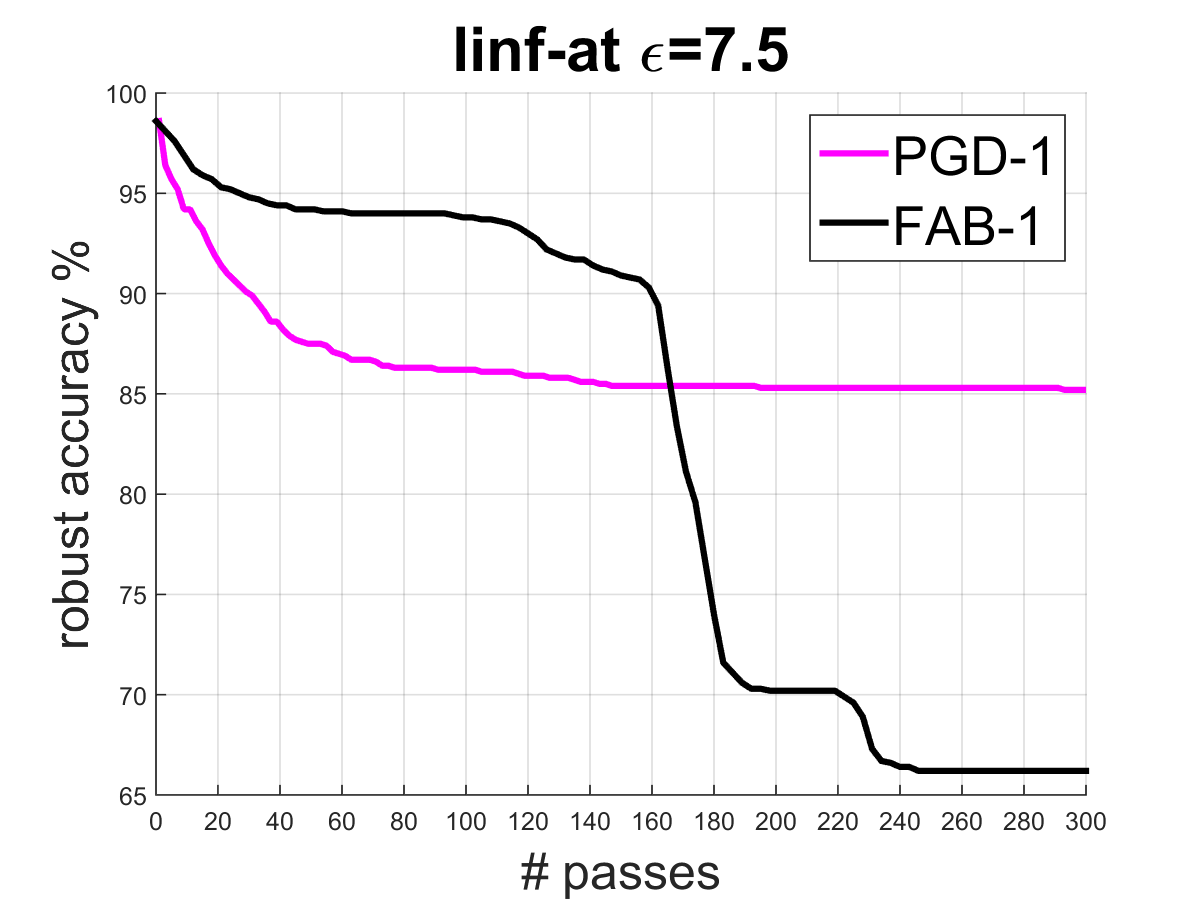} \includegraphics[width=0.5\columnwidth, clip, trim=10mm 0mm 15mm 0mm]{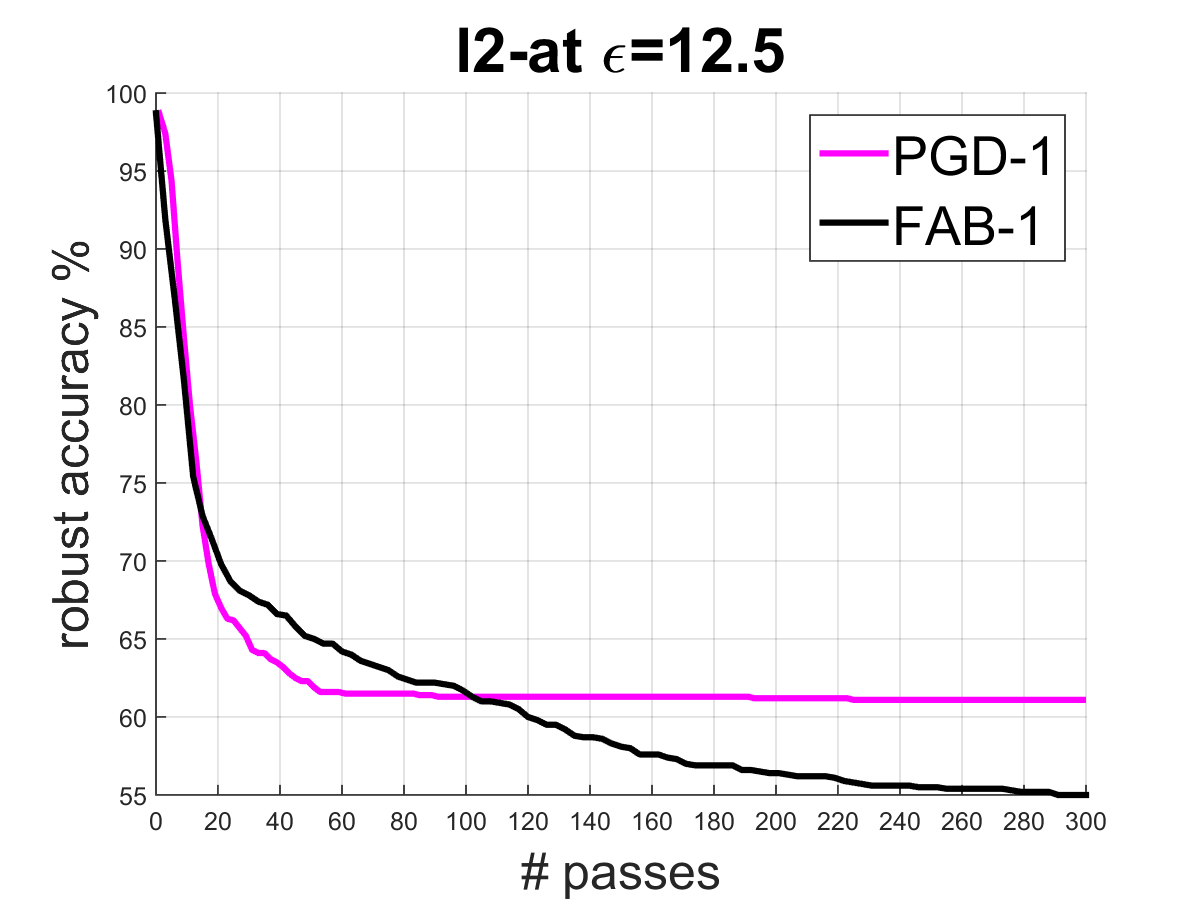}\\
		\hline
	\end{tabular}
	\caption{Evolution of robust accuracy as a function of the employed forward/backward passes on MNIST to ensure a fair comparison of PGD and FAB. We compare PGD-1 (magenta) and FAB-1 (black).
		Models: plain in the first column, $l_\infty$-at in the second and $l_2$-at in the third. Threat models: $l_\infty$ in the first row, $l_2$ in the second and $l_1$ in the third. The thresholds $\epsilon$ used can be read above the plots.
	}\label{fig:evol_iter_mnist}
	\end{figure*}

\begin{figure*}[p]\centering
	\begin{tabular}{c}
\multicolumn{1}{c}{\textbf{CIFAR-10, $l_\infty$}}\\[2mm]
\includegraphics[width=0.5\columnwidth, clip, trim=10mm 0mm 15mm 0mm]{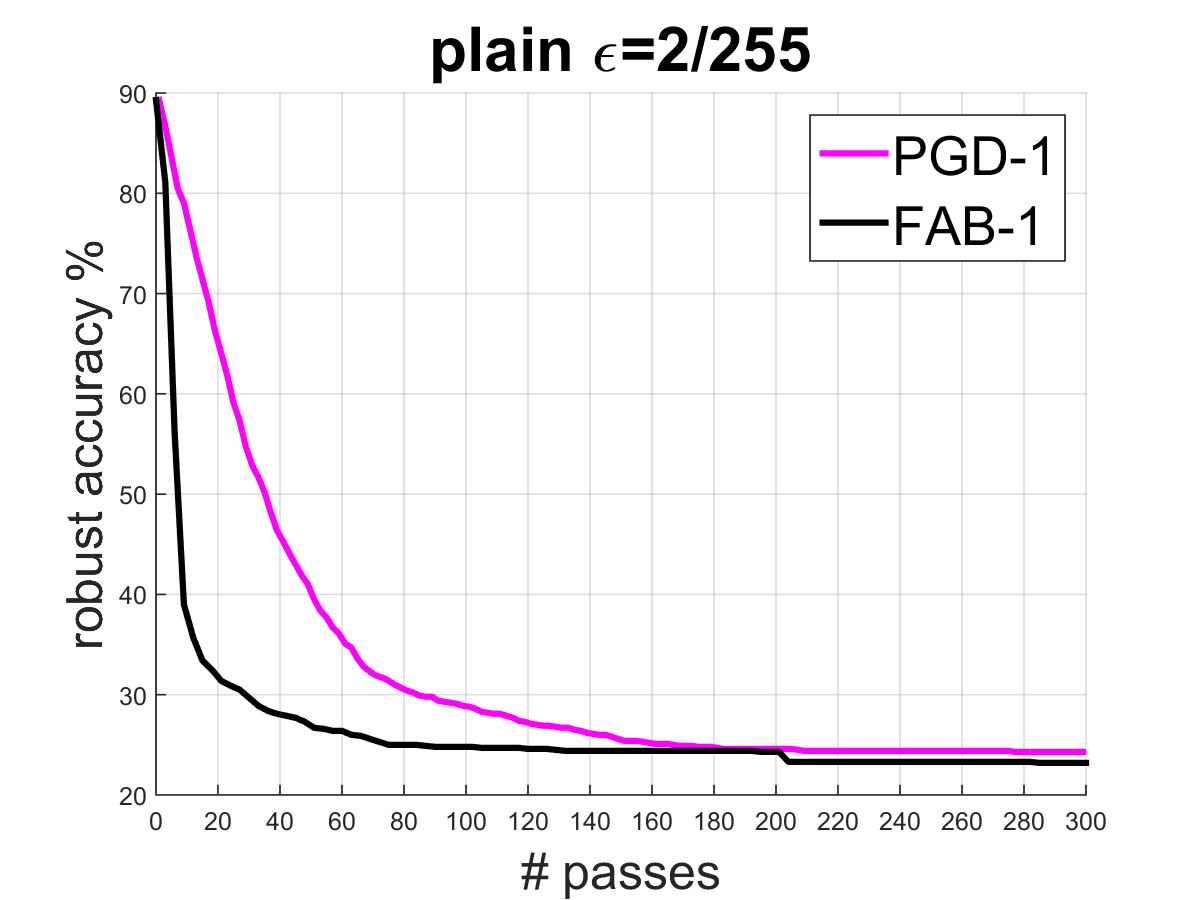}\includegraphics[width=0.5\columnwidth, clip, trim=10mm 0mm 15mm 0mm]{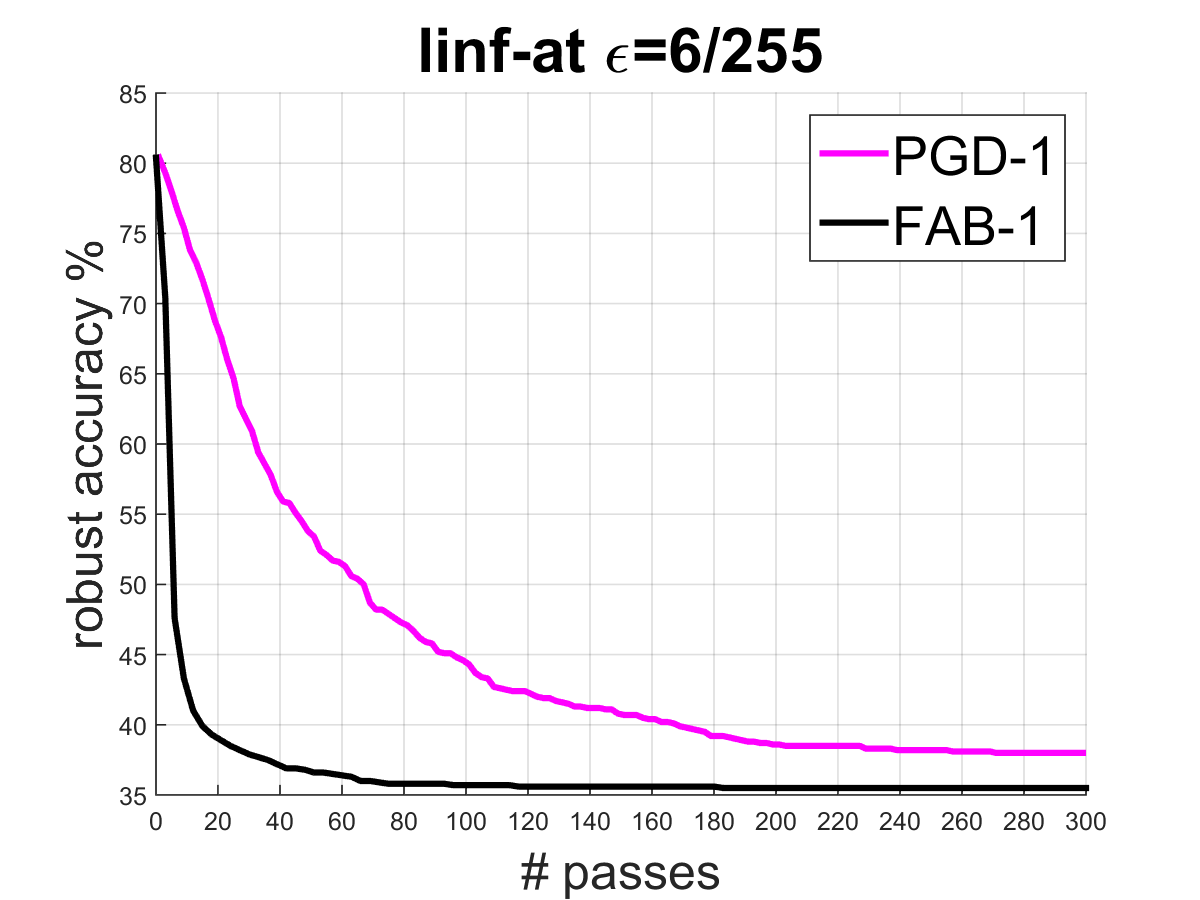} \includegraphics[width=0.5\columnwidth, clip, trim=10mm 0mm 15mm 0mm]{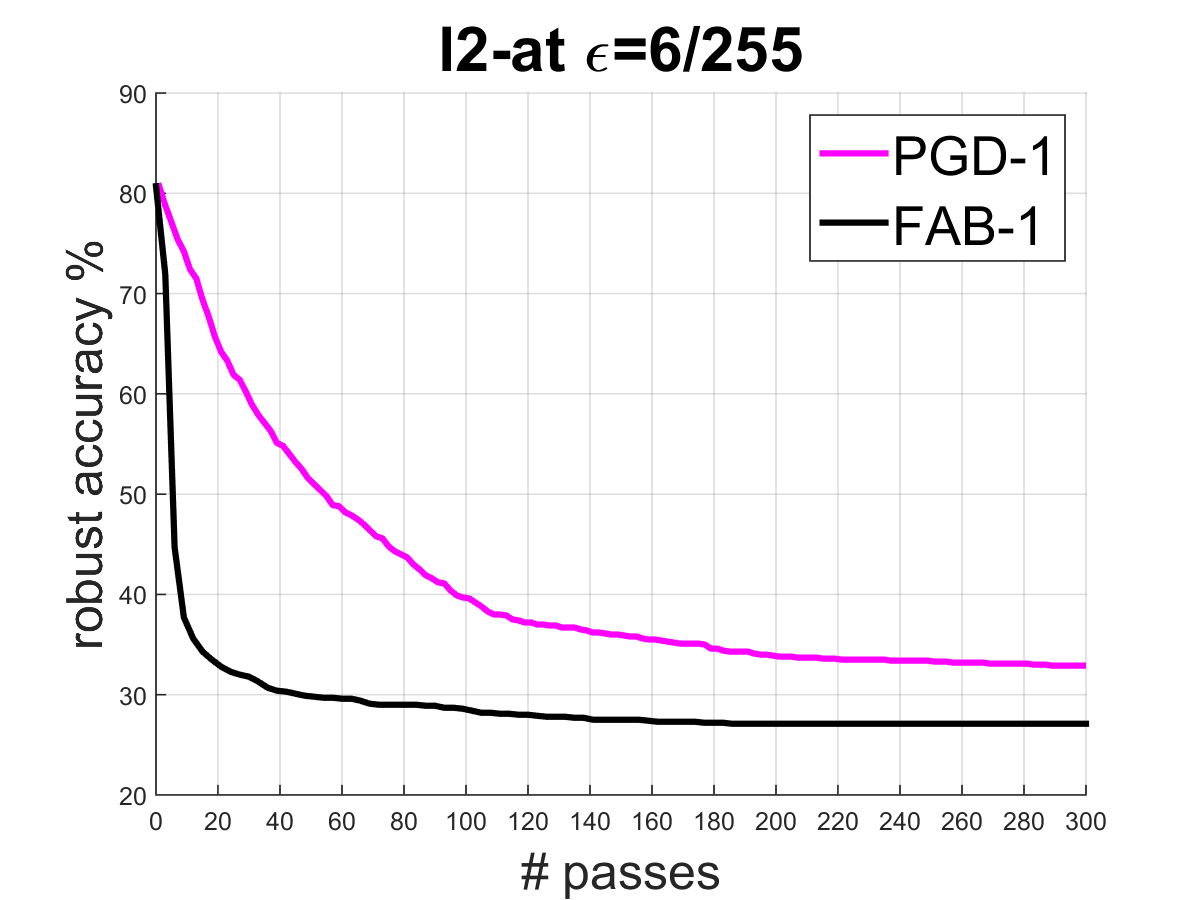}\\
\hline
\\
\multicolumn{1}{c}{\textbf{CIFAR-10, $l_2$}}\\[2mm]
\includegraphics[width=0.5\columnwidth, clip, trim=10mm 0mm 15mm 0mm]{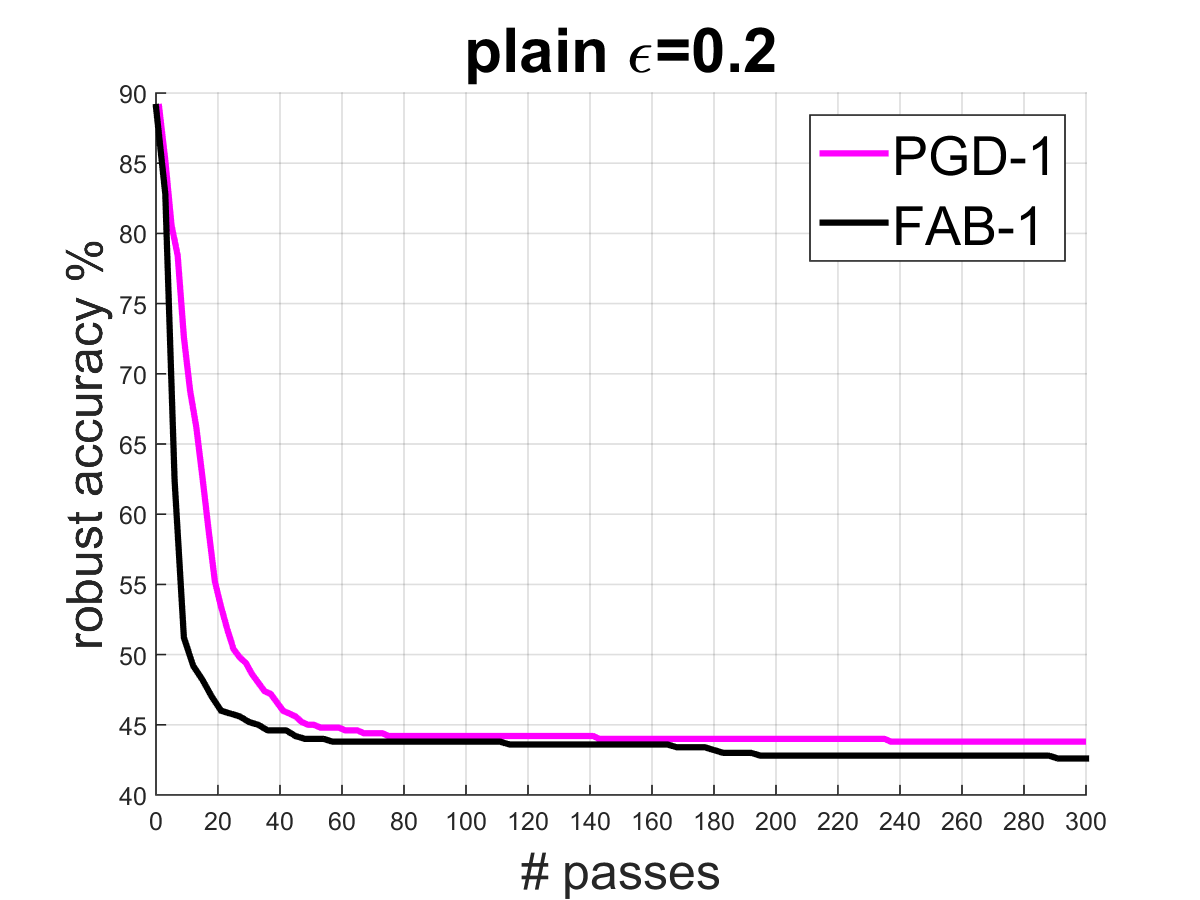}\includegraphics[width=0.5\columnwidth, clip, trim=10mm 0mm 15mm 0mm]{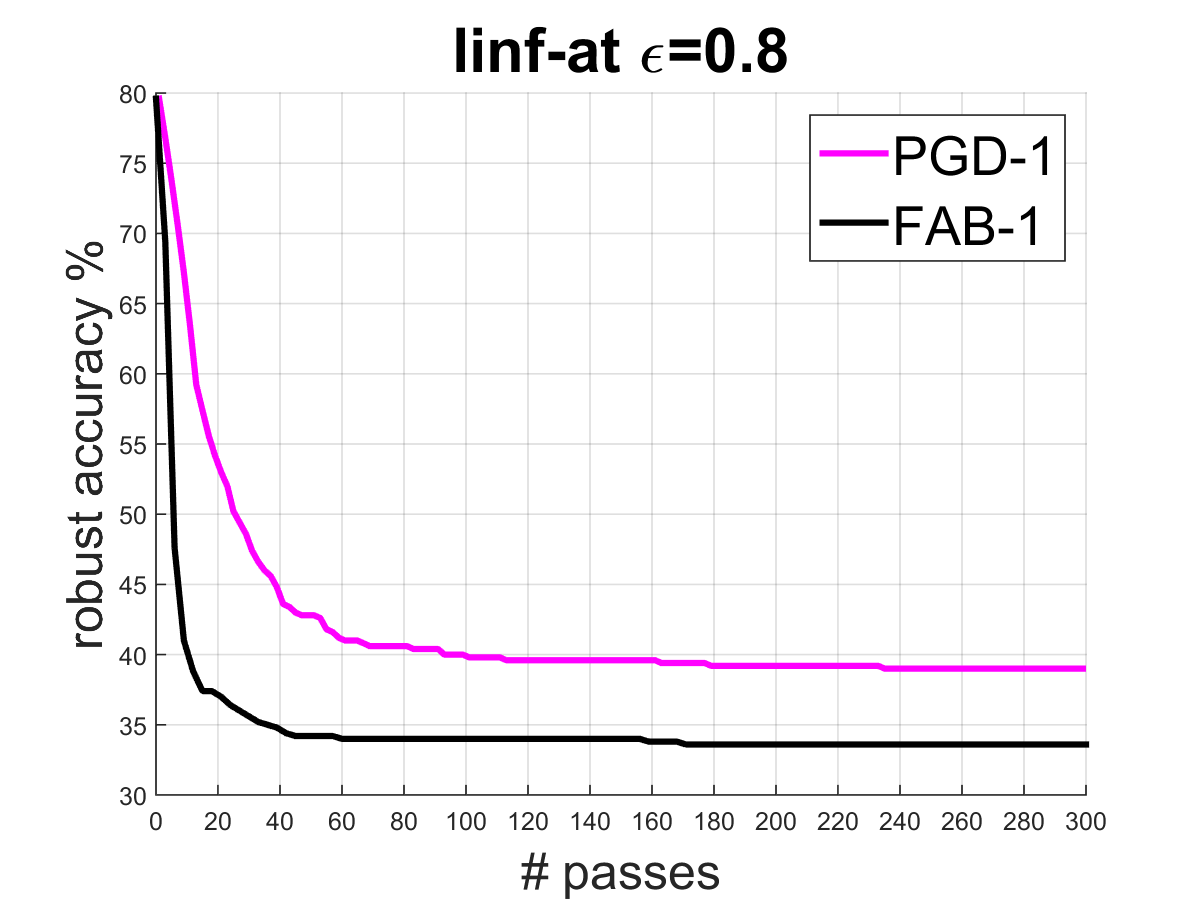} \includegraphics[width=0.5\columnwidth, clip, trim=10mm 0mm 15mm 0mm]{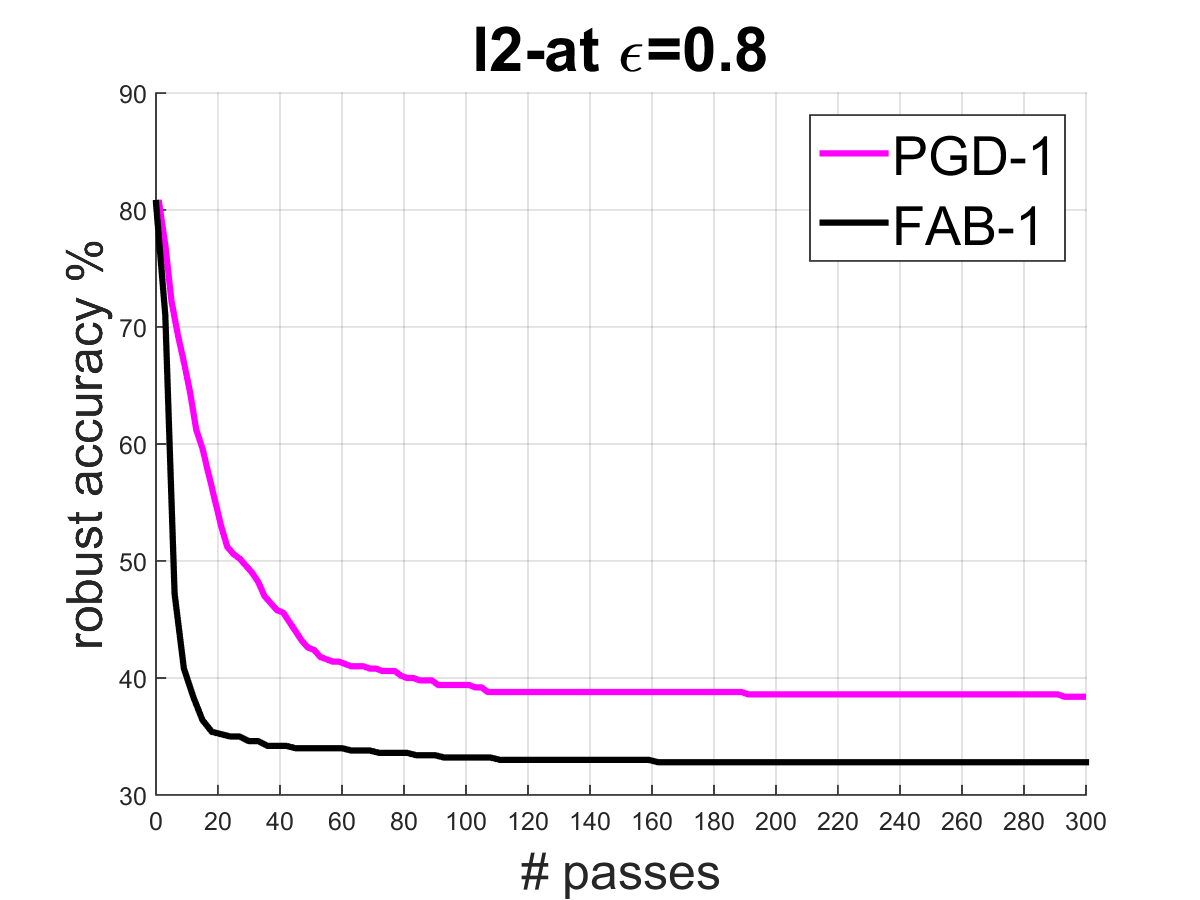}\\
\hline
\\
\multicolumn{1}{c}{\textbf{CIFAR-10, $l_1$}}\\[2mm]
\includegraphics[width=0.5\columnwidth, clip, trim=10mm 0mm 15mm 0mm]{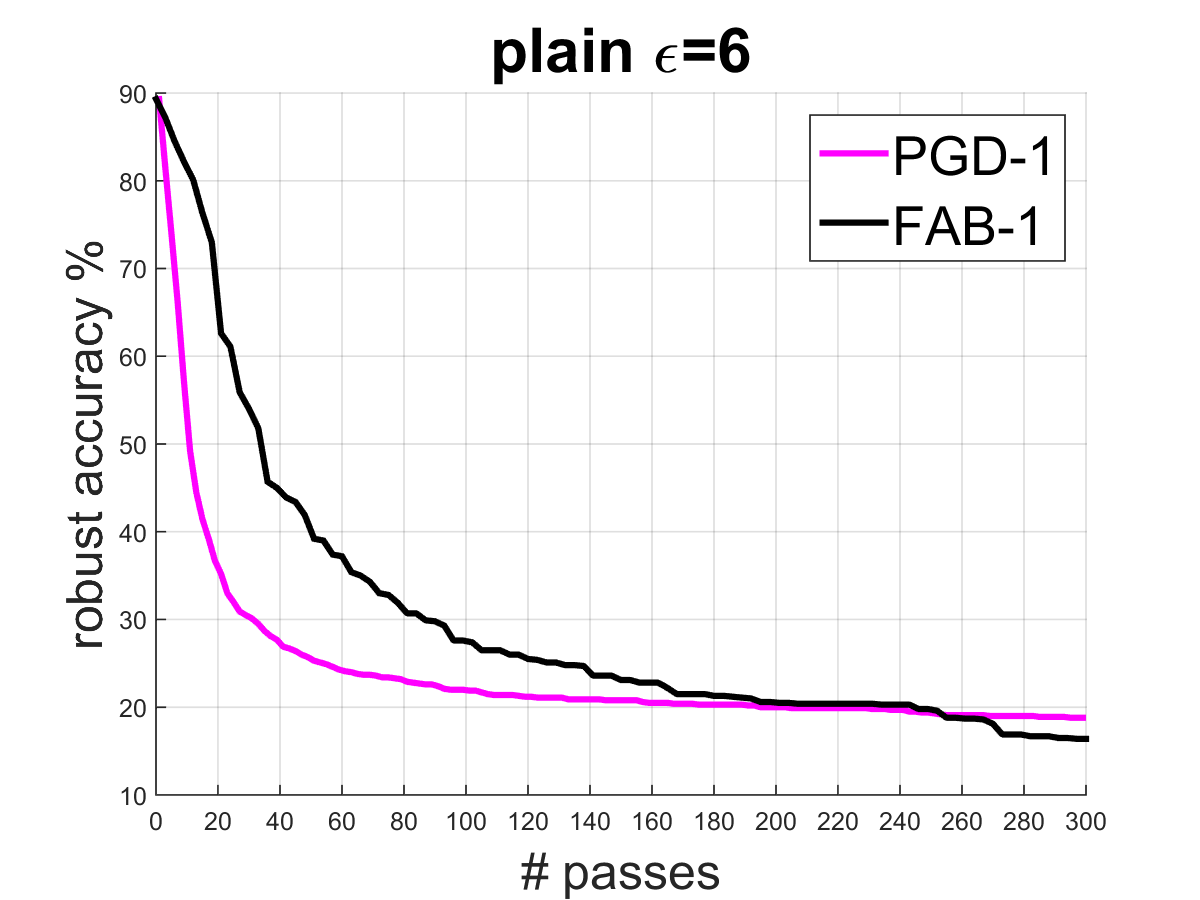}\includegraphics[width=0.5\columnwidth, clip, trim=10mm 0mm 15mm 0mm]{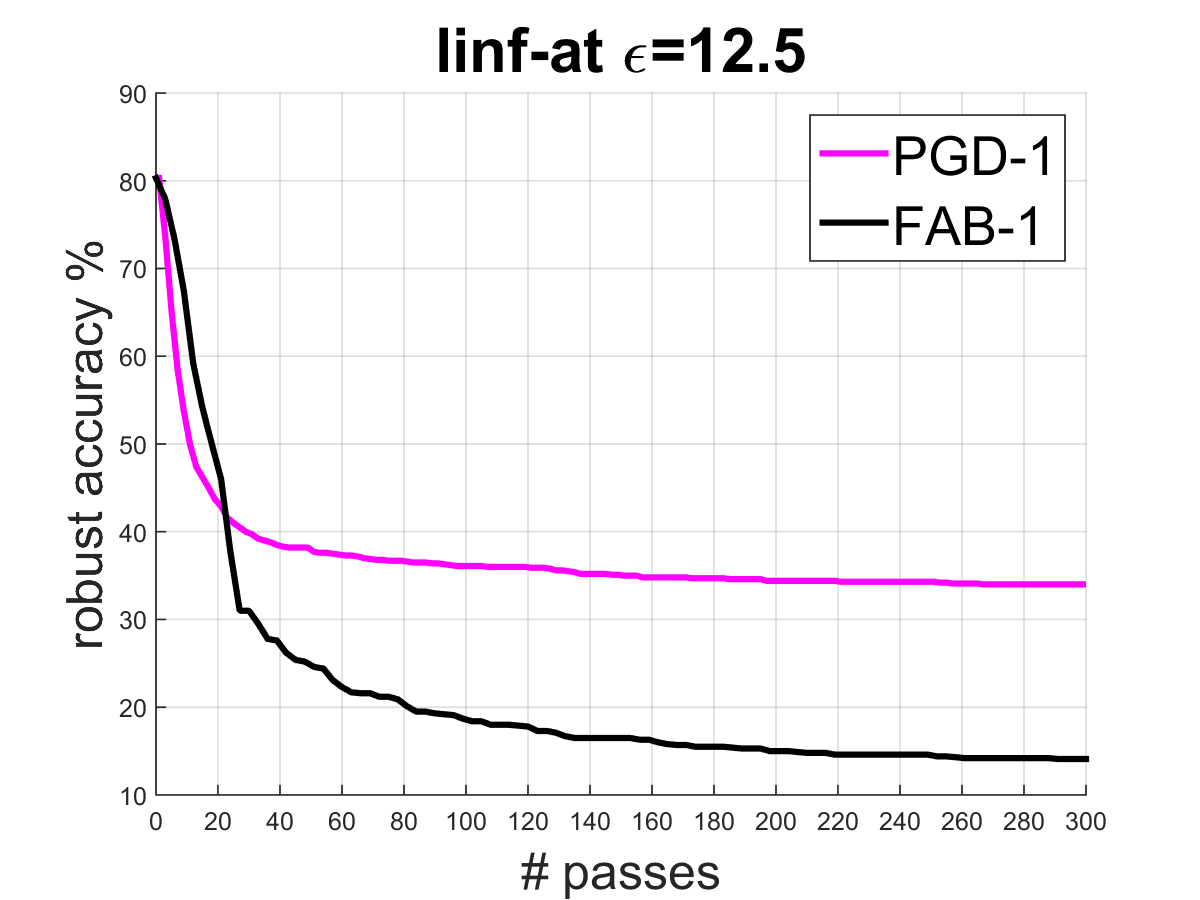} \includegraphics[width=0.5\columnwidth, clip, trim=10mm 0mm 15mm 0mm]{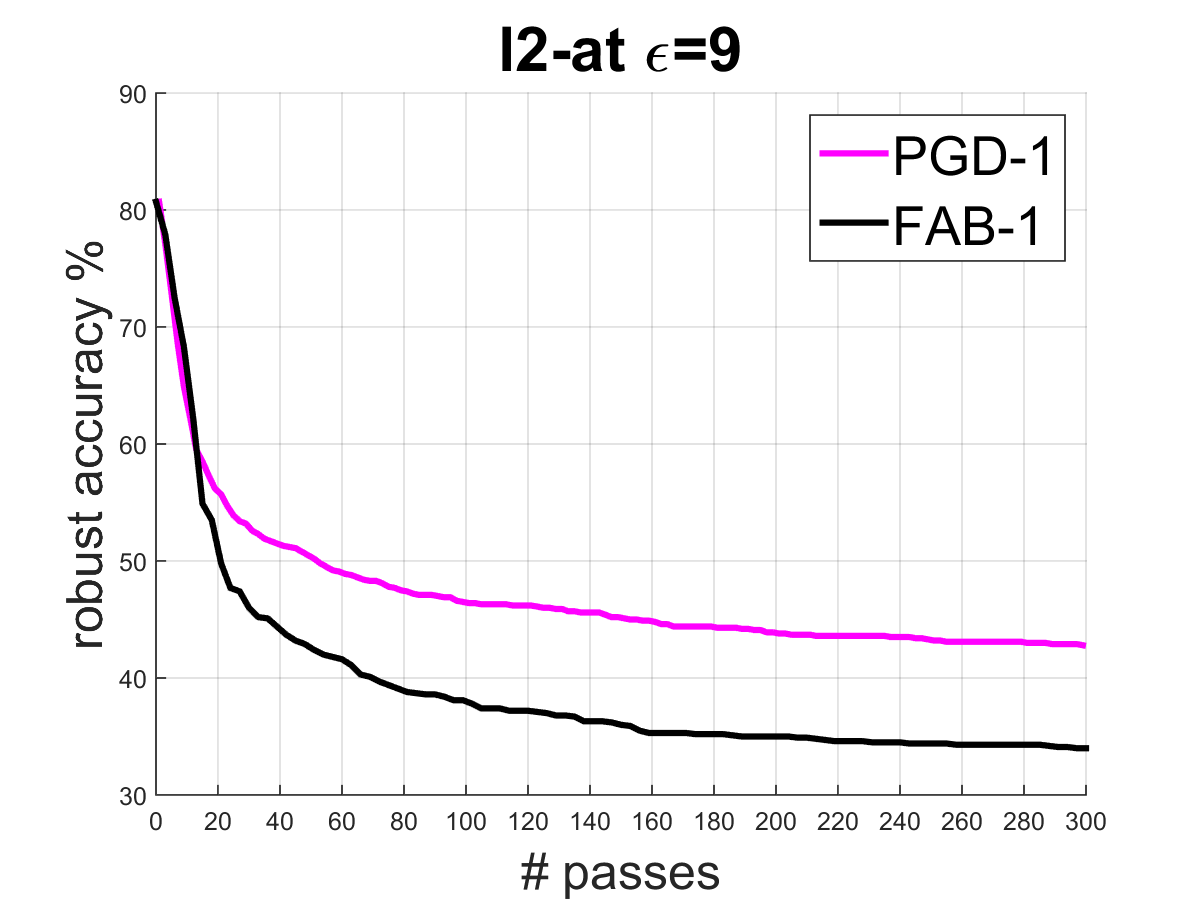}\\
\hline
\end{tabular}
	\caption{Evolution of robust accuracy as a function of the employed forward/backward passes on CIFAR-10 to ensure a fair comparison of PGD and FAB. We compare PGD-1 (magenta) and FAB-1 (black).
		Models: plain in the first column, $l_\infty$-at in the second and $l_2$-at in the third. Threat models: $l_\infty$ in the first row, $l_2$ in the second and $l_1$ in the third. The thresholds $\epsilon$ used can be read above the plots.}\label{fig:evol_iter_cifar10}
\end{figure*}

\begin{figure*}[p]\centering
	\begin{tabular}{c}
		\multicolumn{1}{c}{\textbf{Restricted ImageNet, $l_\infty$}}\\[2mm]
		\includegraphics[width=0.5\columnwidth, clip, trim=10mm 0mm 15mm 0mm]{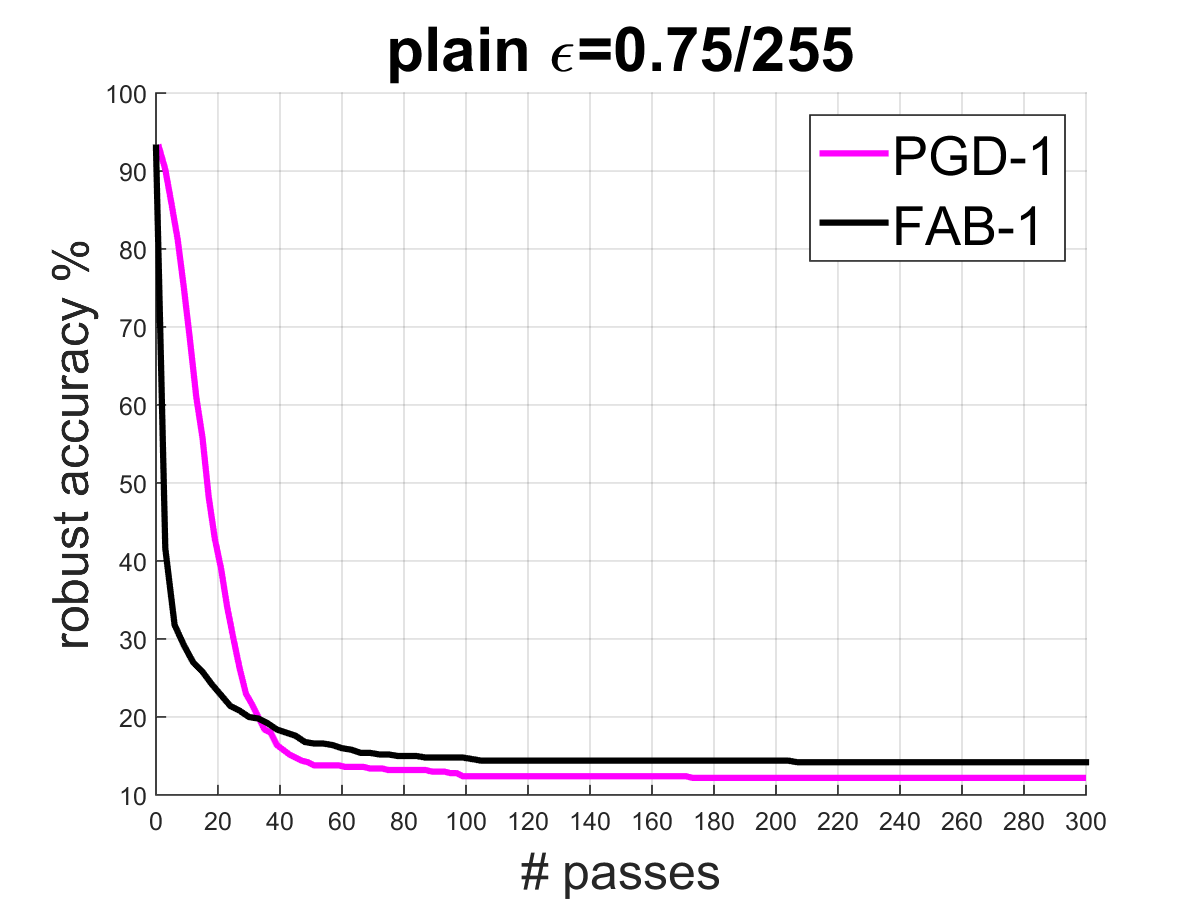}\includegraphics[width=0.5\columnwidth, clip, trim=10mm 0mm 15mm 0mm]{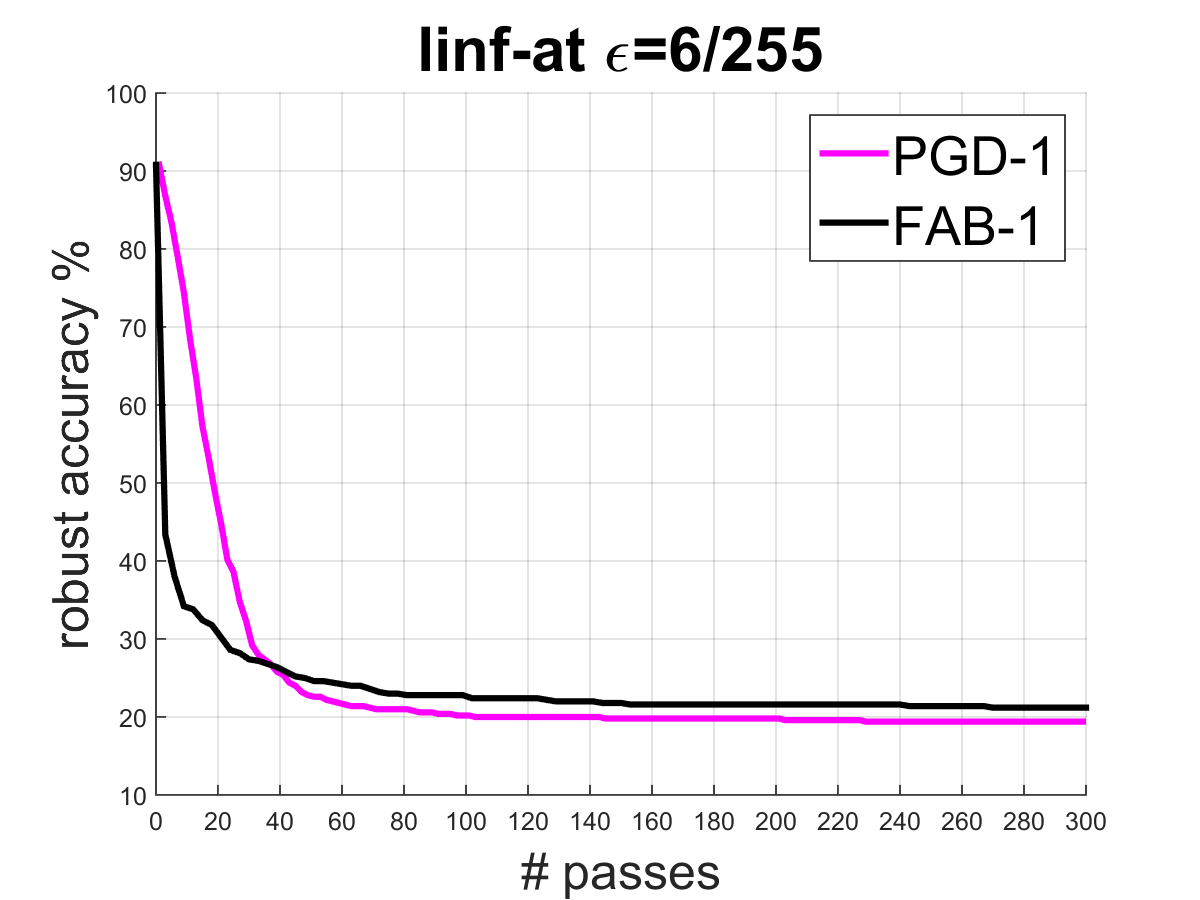} \includegraphics[width=0.5\columnwidth, clip, trim=10mm 0mm 15mm 0mm]{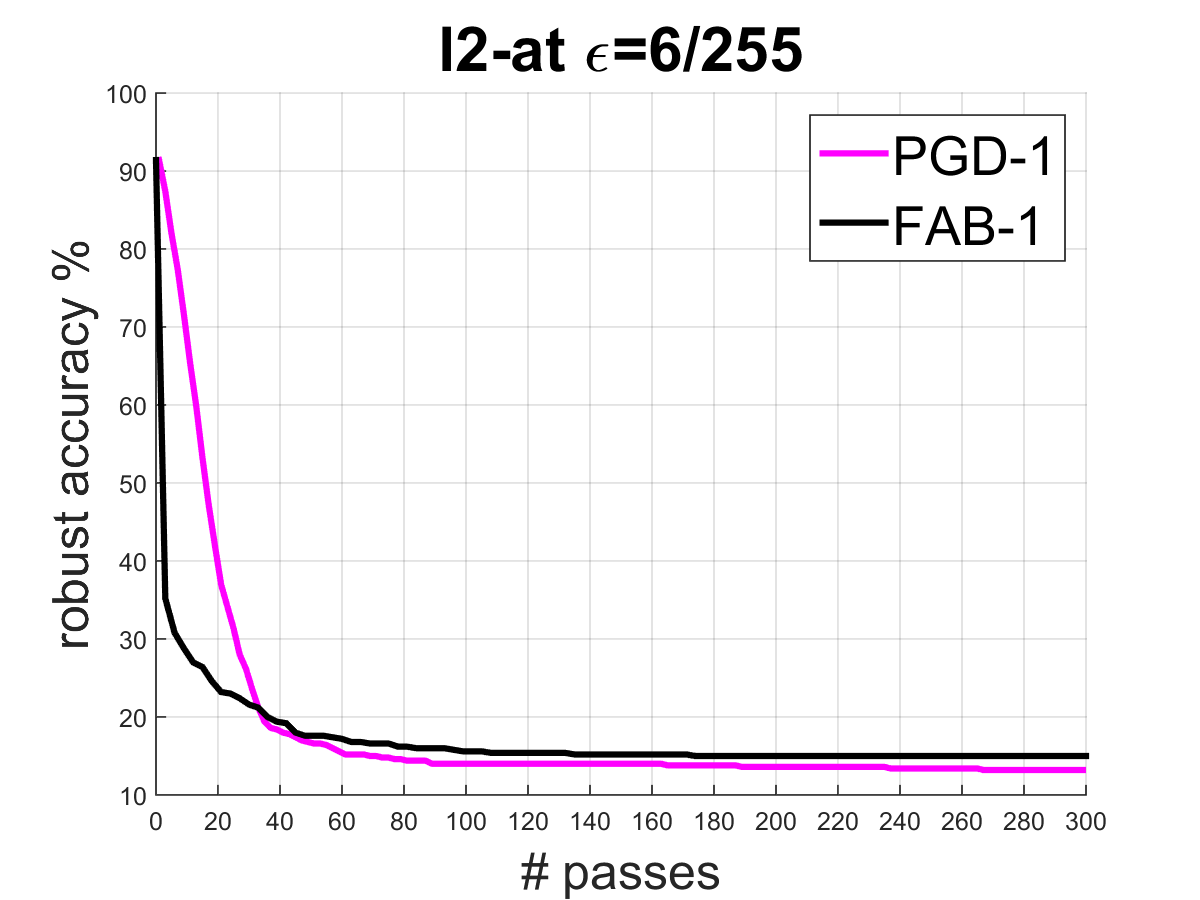}\\
		\hline
		\\
		\multicolumn{1}{c}{\textbf{Restricted ImageNet, $l_2$}}\\[2mm]
		\includegraphics[width=0.5\columnwidth, clip, trim=10mm 0mm 15mm 0mm]{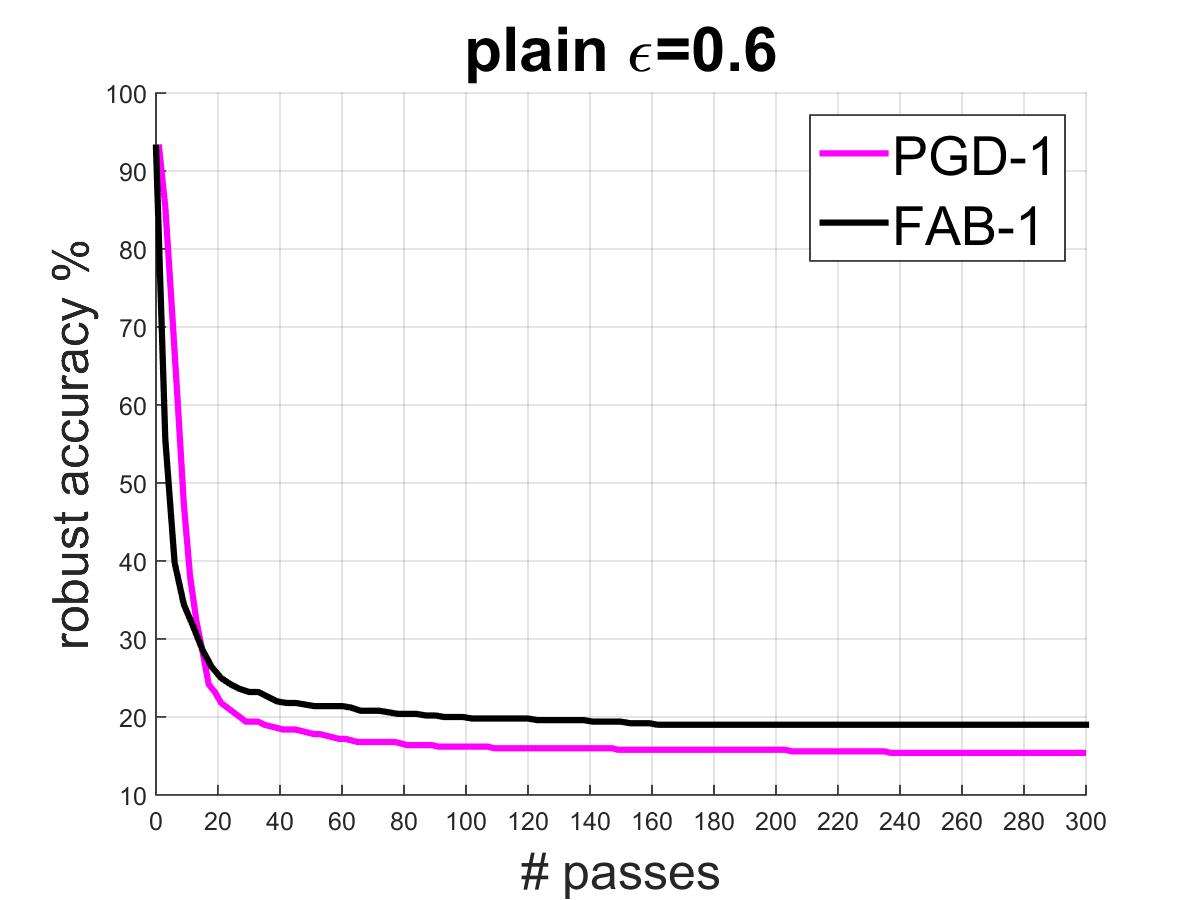}\includegraphics[width=0.5\columnwidth, clip, trim=10mm 0mm 15mm 0mm]{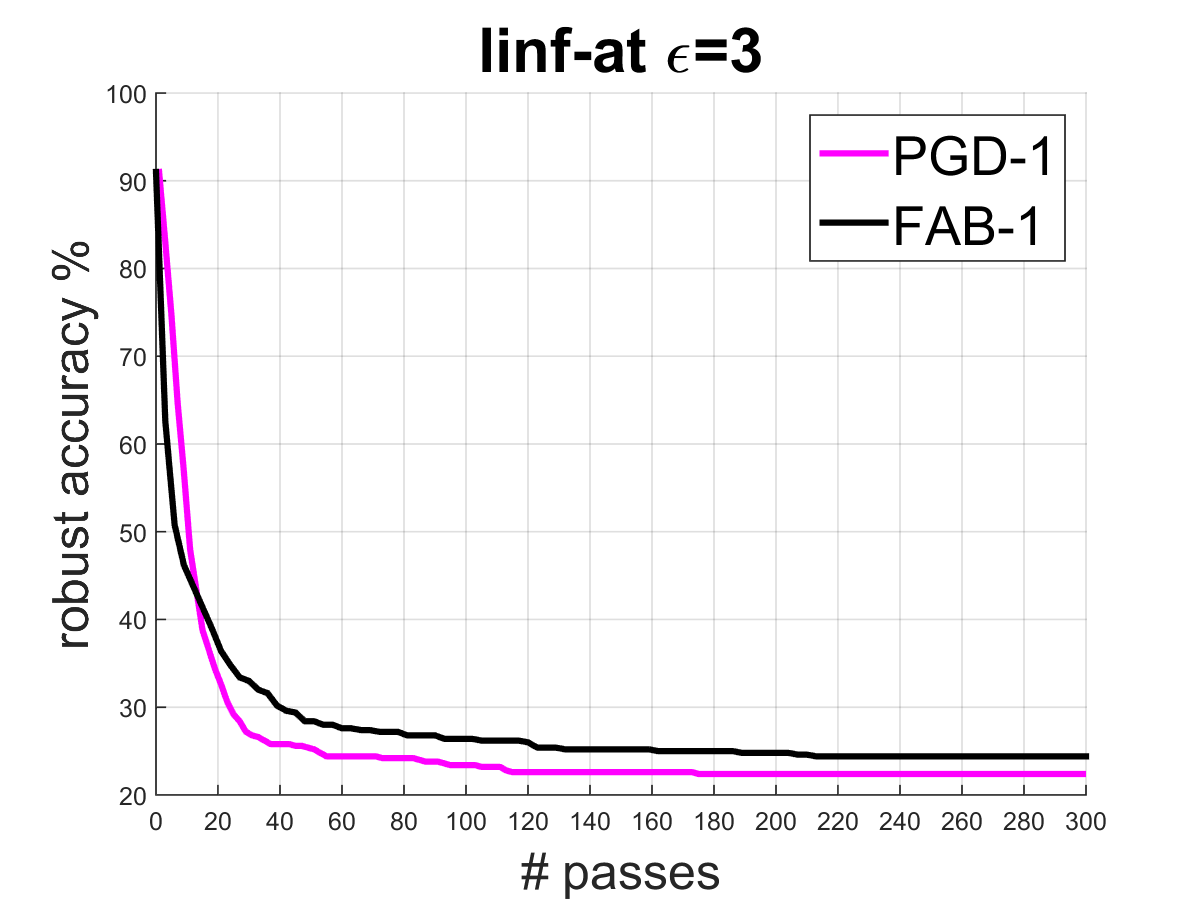} \includegraphics[width=0.5\columnwidth, clip, trim=10mm 0mm 15mm 0mm]{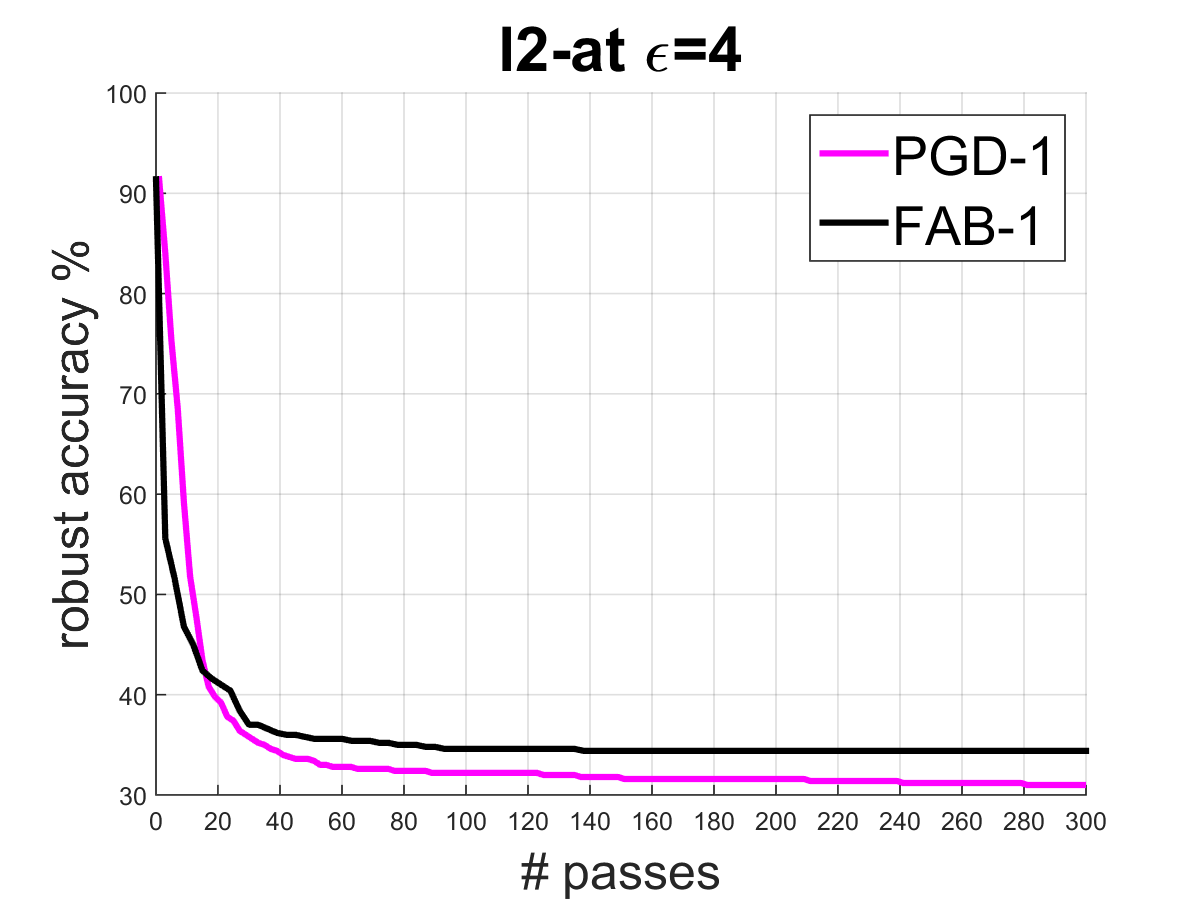}\\
		\hline
		\\
		\multicolumn{1}{c}{\textbf{Restricted ImageNet, $l_1$}}\\[2mm]
		\includegraphics[width=0.5\columnwidth, clip, trim=10mm 0mm 15mm 0mm]{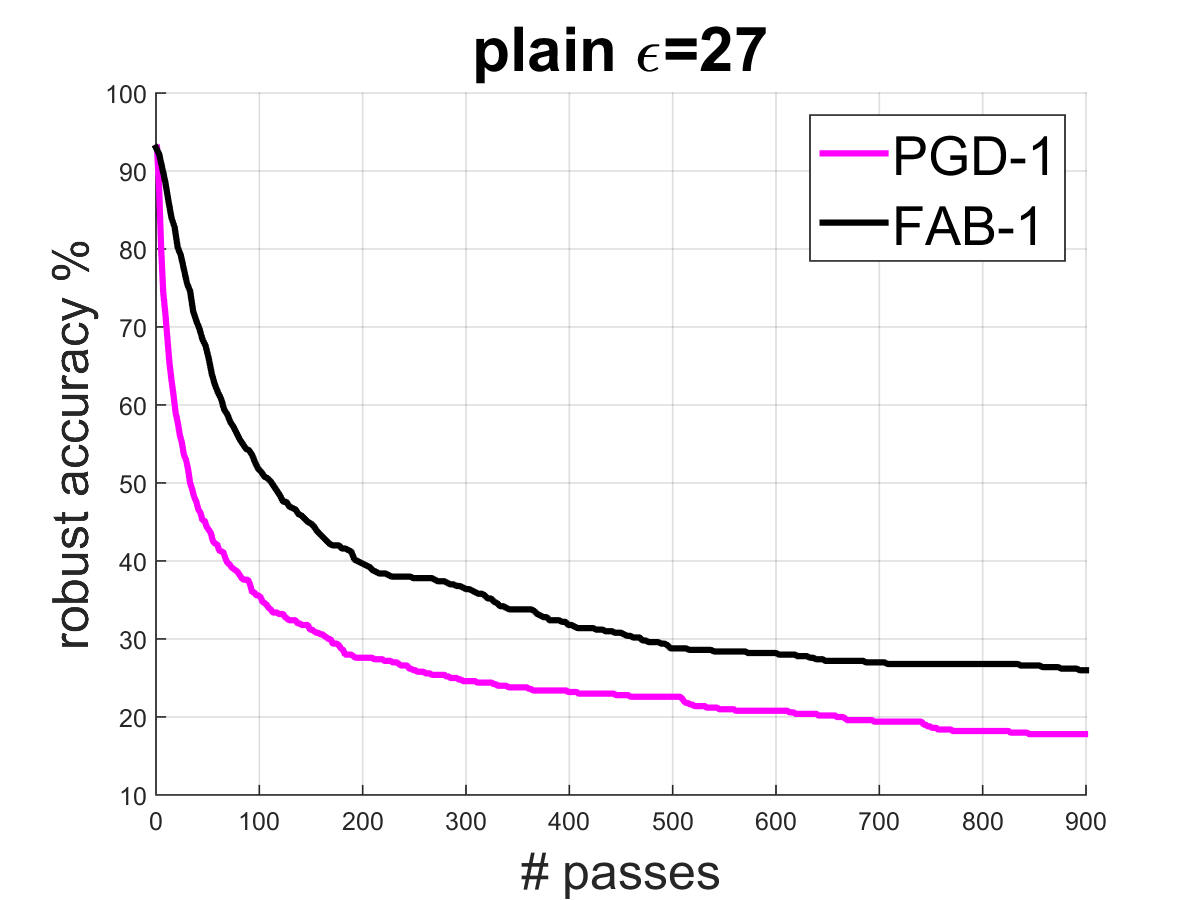}\includegraphics[width=0.5\columnwidth, clip, trim=10mm 0mm 15mm 0mm]{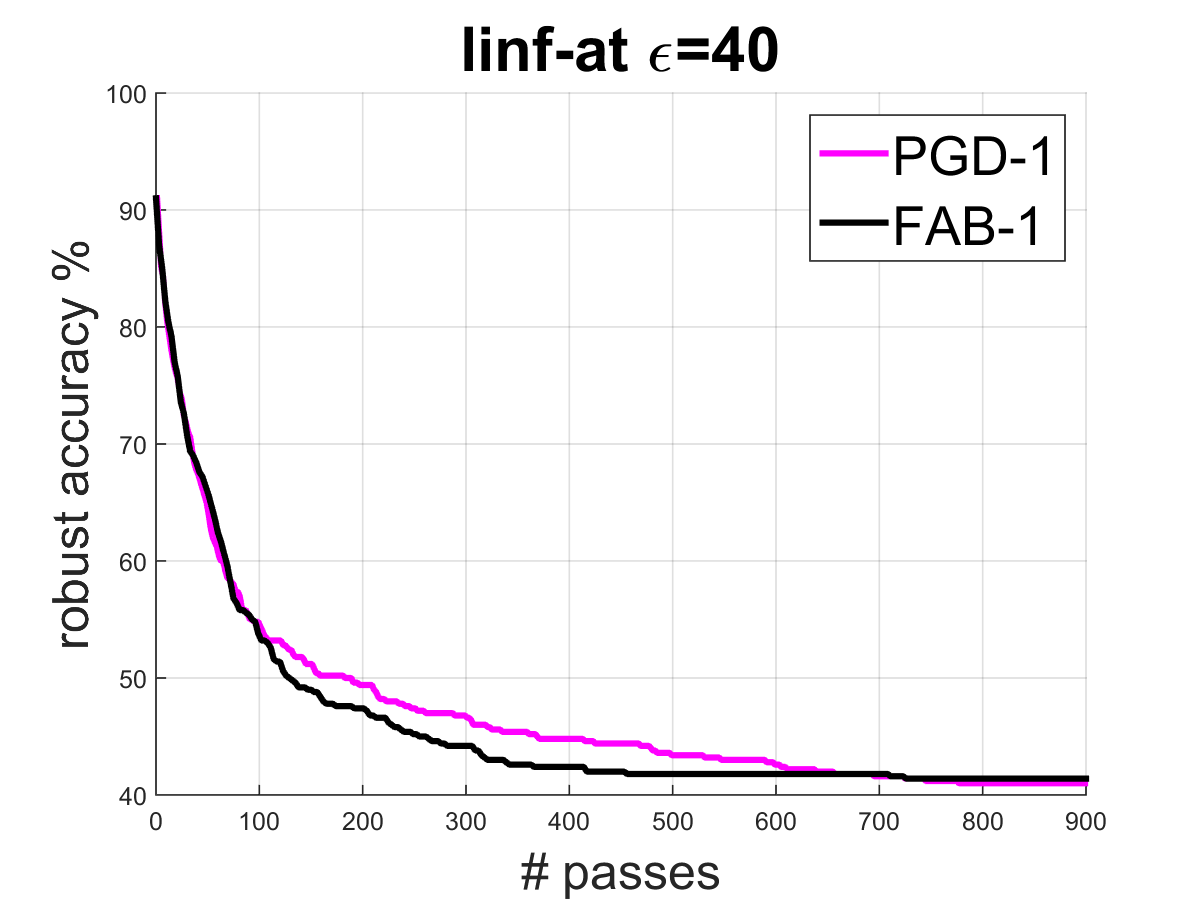} \includegraphics[width=0.5\columnwidth, clip, trim=10mm 0mm 15mm 0mm]{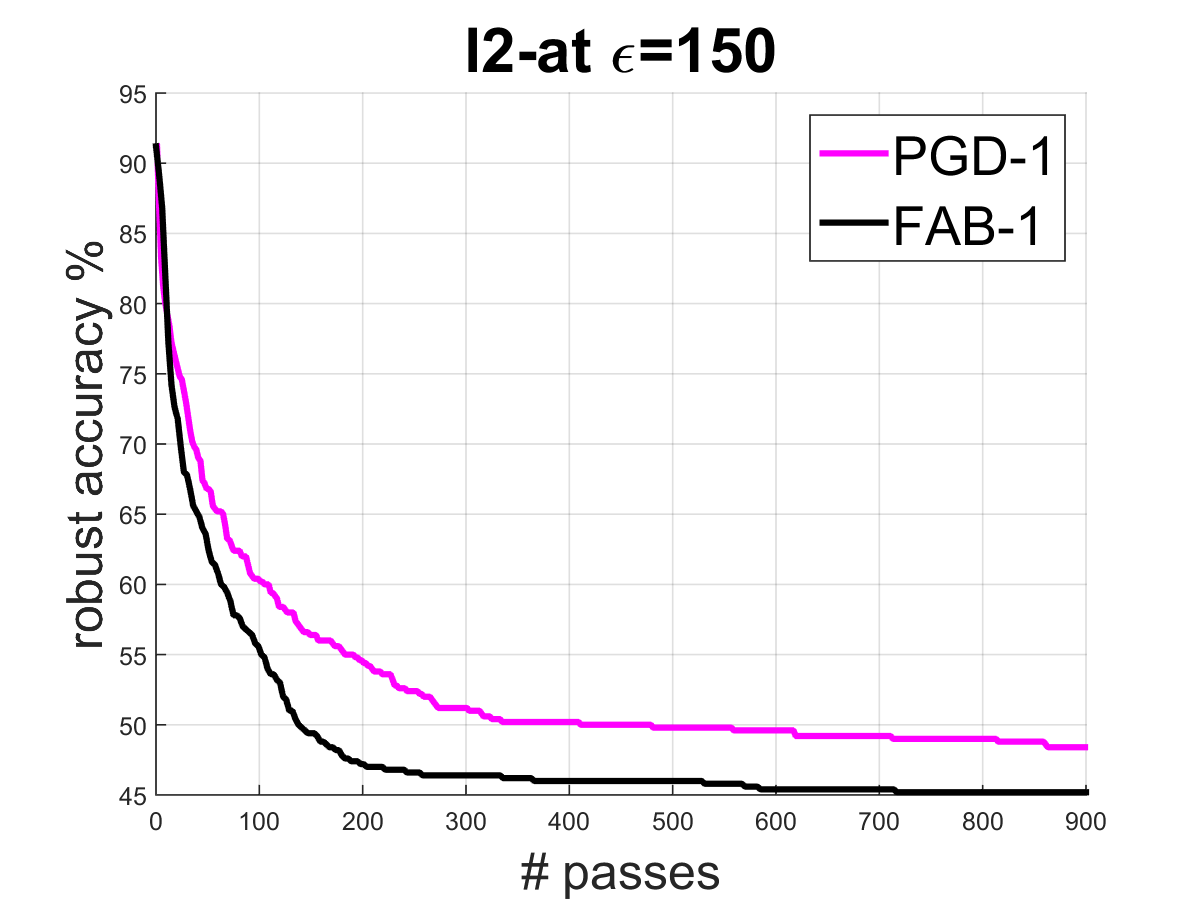}\\
		\hline
	\end{tabular}
	\caption{Evolution of robust accuracy as a function of the employed forward/backward passes on Restricted ImageNet to ensure a fair comparison of PGD and FAB.
		We compare PGD-1 (magenta) and FAB-1 (black).
	Models: plain in the first column, $l_\infty$-at in the second and $l_2$-at in the third. Threat models: $l_\infty$ in the first row, $l_2$ in the second and $l_1$ in the third. The thresholds $\epsilon$ used can be read above the plots.}\label{fig:evol_iter_r-in}
	\end{figure*}

\subsection{Evolution across iteration} \label{sec:evol_iter}
We compare the evolution of the robust accuracy across the iterations of PGD-1 (random starting point) and FAB-1 (starting at the target point $x_\textrm{orig}$).
for the models and datasets from \ref{tab:MNIST_plain} to \ref{tab:ImageNet_l2}. Since PGD performs 1 forward and 1 backward pass for each iteration and FAB 2 forward passes and 1 backward pass, we rescale the robust accuracy so to compare the two methods when they have used exactly the same number of passes of the network.
Then 300 passes correspond to 150 iterations of PGD and to 100 of FAB.
In Figures \ref{fig:evol_iter_mnist}, \ref{fig:evol_iter_cifar10} and \ref{fig:evol_iter_r-in} we show the evolution of robust accuracy for the different datasets, models and threat models ($l_\infty$, $l_2$ and $l_1$), computed at two thresholds $\epsilon$ among the five used in Tables \ref{tab:MNIST_plain} to \ref{tab:ImageNet_l2}. One can see how FAB achieves in most of the cases good results within a few iterations, often faster than PGD.

%%
%}%

\fi

\end{document}